\documentclass{article} % For LaTeX2e
\usepackage{iclr2023_conference,times}

\usepackage{amsmath}
\usepackage{amsfonts}
\usepackage{bm}

\def\eqref#1{equation~\ref{#1}}
\def\twoEqrefs#1#2{Equations~\ref{#1} and \ref{#2}}
\def\Eqref#1{Equation~\ref{#1}}
\def\1{\bm{1}}

\DeclareMathAlphabet{\mathsfit}{\encodingdefault}{\sfdefault}{m}{sl}
\SetMathAlphabet{\mathsfit}{bold}{\encodingdefault}{\sfdefault}{bx}{n}

\DeclareMathOperator*{\argmin}{arg\,min}

\usepackage{natbib}
\usepackage[utf8]{inputenc} % allow utf-8 input
\usepackage[T1]{fontenc}    % use 8-bit T1 fonts
\usepackage{hyperref}       % hyperlinks
\usepackage{url}            % simple URL typesetting
\usepackage{booktabs}       % professional-quality tables
\usepackage{nicefrac}       % compact symbols for 1/2, etc.
\usepackage{microtype}      % microtypography
\usepackage{xcolor}         % colors
\usepackage{multirow}
\usepackage{mathtools}
\usepackage{graphicx, wrapfig}
\usepackage{subfigure}
\usepackage{caption}
\usepackage{xfrac}
\usepackage{diagbox}

\newtheorem{definition}{Definition}

\newtheorem{theorem}{Theorem}

\usepackage{tabularx}
\usepackage{adjustbox}
\usepackage[graphicx]{realboxes}
\usepackage{stfloats}

\usepackage[linesnumbered,ruled,noend]{algorithm2e} 
\usepackage{caption}

\title{\texttt{STaSy}: Score-based Tabular data Synthesis}

\author{Jayoung Kim, Chaejeong Lee, and Noseong Park \\
Yonsei University\\
South Korea \\
\texttt{\{jayoung.kim, chaejeong\_lee, noseong\}@yonsei.ac.kr} 
}
\iclrfinalcopy % Uncomment for camera-ready version, but NOT for submission.
\begin{document}

\maketitle
\begin{abstract}
Tabular data synthesis is a long-standing research topic in machine learning. Many different methods have been proposed over the past decades, ranging from statistical methods to deep generative methods. However, it has not always been successful due to the complicated nature of real-world tabular data. In this paper, we present a new model named \textbf{S}core-based \textbf{Ta}bular data \textbf{Sy}nthesis (\texttt{STaSy}) and its training strategy based on the paradigm of score-based generative modeling. Despite the fact that score-based generative models have resolved many issues in generative models, there still exists room for improvement in tabular data synthesis. Our proposed training strategy includes a self-paced learning technique and a fine-tuning strategy, which further increases the sampling quality and diversity by stabilizing the denoising score matching training. Furthermore, we also conduct rigorous experimental studies in terms of the generative task trilemma: sampling quality, diversity, and time. In our experiments with 15 benchmark tabular datasets and 7 baselines, our method outperforms existing methods in terms of task-dependant evaluations and diversity. Code is available at \url{https://github.com/JayoungKim408/STaSy}.
% Despite the fact that score-based generative models have resolved many issues in generative models in terms of sampling quality and diversity, there still exists room for improvement in tabular data synthesis.
% In our experiments with 15 benchmark tabular datasets and 7 baselines, our method generates fake data whose quality is comparable to the original data in terms of task-dependant evaluations and diversity.
\end{abstract}

\section{Introduction}

\begin{wraptable}[15]{r}{7.5cm}
\vspace{-1em} 
\setlength\tabcolsep{1.8pt}
% \begin{table}[t]
\small
% \caption{\jayoung{Summary of experimental results in Section~\ref{sec:experiments}. We report average sampling quality, runtime, and diversity scores. \texttt{Identity} means ``train on real, and test on real'', while others follow the ``train on synthetic, test on real (TSTR)'' framework~\citep{esteban2017realvalued,Jordon2019PATEGANGS}. }}\texttt{Identity} means the original data.
\caption{Summary of experimental results. We report the average sampling quality and diversity scores, and time.}

\label{tbl:teaser}
\resizebox{0.53\textwidth}{!}{
\begin{tabular}{lcccccccc}
\toprule
% \specialrule{1pt}{1pt}{1pt}
% \multirow{2}{*}{Baselines}  &&  Sampling  &&  Sampling && Sampling & \\
  \multirow{2}{*}{Methods} &&  Quality $\uparrow$  && Diversity $\uparrow$ &&  Runtime $\downarrow$  & \\
    &&  (F1 \& $R^2$)   && (coverage) &&  (second)   & \\
% \specialrule{1pt}{1pt}{1pt}
\midrule
% \texttt{Identity}&&  && \textcolor{red}{0.994}  && N/A& \\
% \midrule
\texttt{VEEGAN} &&0.006 &&  0.038 && 0.109  &\\ 
\texttt{CTGAN}&& 0.569  & & 0.352 &&  0.704& \\ 
% \texttt{TVAE}&& 0.537 & & 0.494 &&  0.100& \\ 
\texttt{TableGAN}&& 0.501 & & 0.434 &&  \textbf{0.046}& \\ 
\texttt{OCT-GAN} &&  0.567&&0.381 & &  26.926 & \\ 
\texttt{RNODE}&& 0.490&  &  0.328 &&  13.392& \\ 
\midrule
\texttt{Na\"ive-STaSy} && 0.717  &&0.637   &&  8.855 &\\ 
\texttt{STaSy}& & \textbf{0.733} & & \textbf{0.658} && 10.663  & \\ 

\bottomrule
% \specialrule{1pt}{1pt}{1pt}
\end{tabular}
}
% \vspace{-0.5em}
\end{wraptable}

Tabular data synthesis is of non-trivial importance in real-world applications for various reasons: protecting the privacy of original tabular data by releasing fake tabular data~\citep{park2018data, lee2021invertible}, augmenting the original tabular data with fake data for better training machine learning models~\citep{smote, borderlinesmote, adasyn, sos}, and so on. However, it is well-known that tabular data frequently has such peculiar characteristics that deep generative models are not able to synthesize all possible details of the original tabular data ~\citep{park2018data, NIPS2019_8953} --- given a set of columns in tabular data, columns typically follow unpredictable (multi-modal) distributions and therefore, it is hard to model their joint probability. 
% Furthermore, real-world data frequently contains noisy records caused by a range of factors, e.g., equipment errors, human mislabeling, programming bugs, and so on, making the modeling task challenging.

A couple of recent methods, however, showed remarkable successes (with some failure cases) in synthesizing fake tabular data, such as \texttt{CTGAN}~\citep{NIPS2019_8953}, \texttt{TVAE}~\citep{NIPS2019_8953}, IT-GAN~\citep{lee2021invertible}, and \texttt{OCT-GAN}~\citep{10.1145/3442381.3449999}. In addition, a recent generative model paradigm, called score-based generative modeling (SGMs), successfully resolves the two problems of \textit{the generative learning trilemma}~\citep{trillemma}, i.e., score-based generative models provide high sampling quality and diversity, although their training/sampling time is relatively longer than other deep generative models. In this paper, we adopt a score-based generative modeling paradigm and design a \textbf{S}core-based \textbf{Ta}bular data \textbf{Sy}nthesis (\texttt{STaSy}) method. 
% In the case of the entire data generation, training each class separately may yield inconsistent between real and learned distribution, since each class may share similar information and approximate them separately can be a loss of information.

% \old{In the tabular data synthesis domain, however, score-based generative modeling has not been applied yet to our knowledge. In this paper, we adopt the score-based generative modeling paradigm, and design a score-based tabular data synthesis (\texttt{STaSy}) method.  To our knowledge, we are the first to propose a tabular data synthesis method based on score-based generative models (although SGMs had been adopted to other domains, e.g., voice synthesis~\citep{bddm}).}

\begin{wrapfigure}[13]{l}{0.35\linewidth}
% \vspace{-1.2em}
\centering
\includegraphics[width=0.33\textwidth]{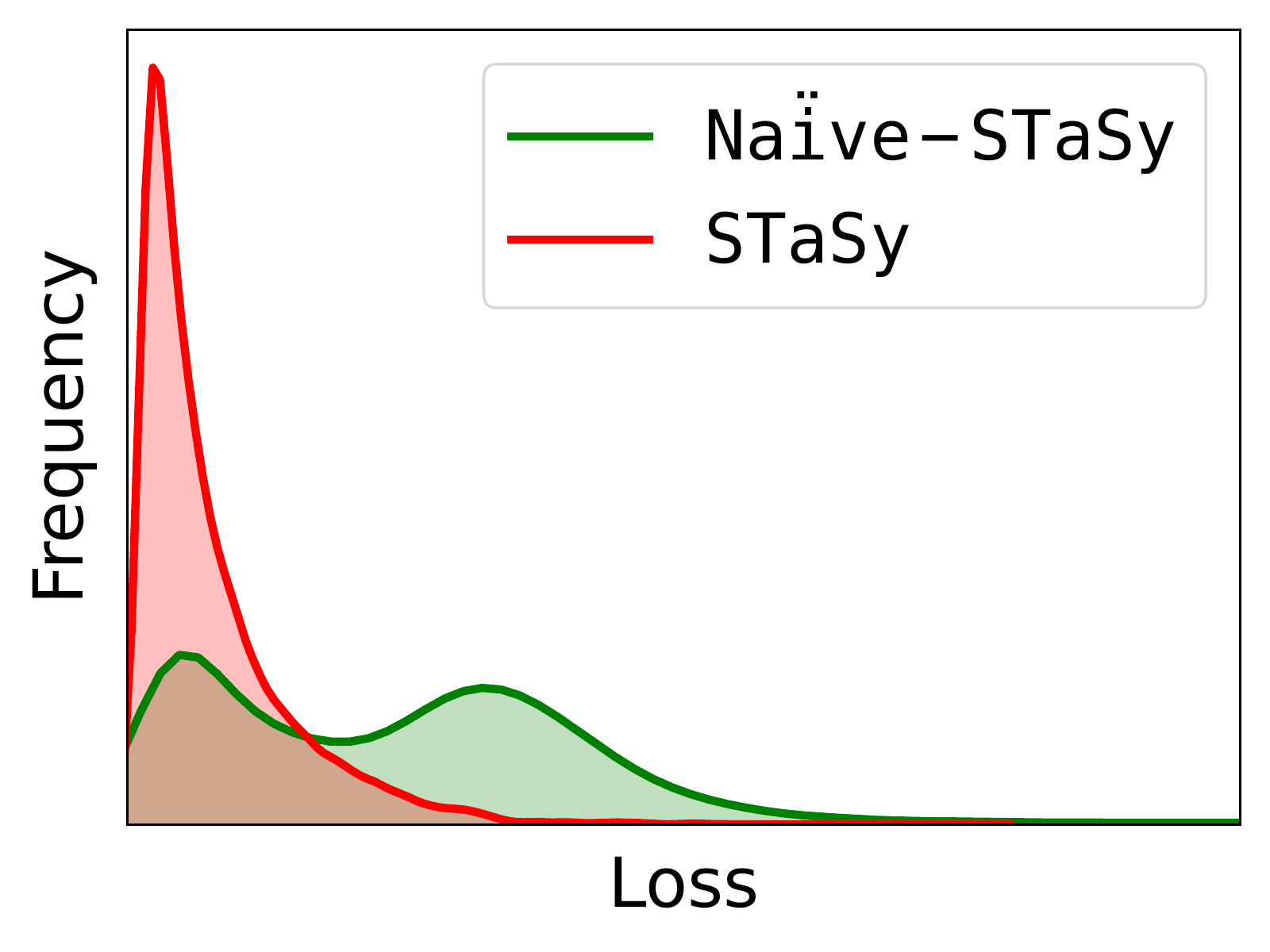}
\caption{Distributions of denoising score matching loss in \texttt{Shoppers}}\label{fig:loss_hist}
% \vspace{-2em}
\end{wrapfigure}

Our model designs significantly outperform all existing baselines in terms of the sampling quality and diversity (cf. \texttt{Na\"ive-STaSy} and \texttt{STaSy} in Table~\ref{tbl:teaser}) --- \texttt{Na\"ive-STaSy} is a naive conversion of SGMs toward tabular data, and \texttt{STaSy} additionally uses our proposed self-paced learning and fine-tuning methods. However, Figure~\ref{fig:loss_hist} shows the difficulty of training \texttt{Na\"ive-STaSy}. The uneven and long-tailed loss distribution of \texttt{Na\"ive-STaSy} at the end of its training process means that the training is not stable for some records in training tabular data. In contrast, \texttt{STaSy} with our two proposed training methods yields many loss values around the left corner (i.e., close to 0).

% In particular, the training loss of tabular data synthesis models varies greatly depending on the mode for each sample. Fig. ~\ref{fig:loss_hist} shows a distribution of training loss of score-based model for tabular data, which is uneven and long tailed, implying that the tabular data contains a considerable number of complex samples to train. 

% In order to alleviate the training difficulty, we design a self-paced learning and a fine-tuning approach to further enhance its sampling quality. 
In order to alleviate the training difficulty of \texttt{Na\"ive-STaSy}, we design i) a self-paced learning method, and ii) a fine-tuning approach. % , further enhancing its sampling quality. 
% Most tabular data consists of multiple types of columns, i.e., continuous, categorical, and ordinal columns. Tabular data synthesis models must learn those multi-modal distributions to synthesize several columns simultaneously, which is a non-trivial task.  As a result, the training loss of tabular data synthesis models varies greatly depending on the mode for each sample. For example, fig. ~\ref{} shows a distribution of training loss of score-based model for tabular data, which has a long tail, implying that the data set contains a considerable number of complex samples to train.
Our proposed self-paced learning technique trains a model from easy to hard records based on their loss values. 
% We utilize self-paced learning in particular; the model chooses and learns simple samples on its own, gradually increasing the learning challenge, by adjusting the training objective depending on the loss values of each training sample. 
In addition, our proposed fine-tuning method, which modestly adjusts the model parameters, can further improve the sampling quality and diversity. 
% SDE satisfies an ODE in some conditions. 

% Leveraging SDE that satisfies an ODE, and the change of variable theorem, one can compute the exact likelihood of each data sample.  Once the model has been trained, we fine-tune the model by retraining data samples whose likelihood is less than a threshold.

% In Table~\ref{tbl:teaser}, we show a snapshot of our experimental results, where we compare testing machine learning models trained with original tabular data vs. trained with fake tabular data --- \texttt{Identity} means the task-oriented evaluation score of the machine learning model trained with original tabular data. As shown, all existing baselines do not show scores comparable to \texttt{Identity}, but our method (even without our proposed self-paced learning or fine-tuning) significantly outperforms all baselines and even \texttt{Identity} in many cases.
In Table~\ref{tbl:teaser}, we summarize our experimental results, where we compare our \texttt{STaSy} with other existing tabular data synthesis methods in terms of the sampling quality, diversity, and time. As shown, our basic model even without our proposed self-paced learning and fine-tuning, denoted \texttt{Na\"ive-STaSy}, significantly outperforms all baselines except for runtime. 
% \jayoung{Especially for the sampling quality, \texttt{STaSy} with our proposed improved training shows quality comparable to \texttt{Identity}. }

% We consider \texttt{STaSy} as a milestone in the field of tabular data synthesis for two reasons: i) our method clearly balances among the key requirements of generative modeling when compared to others, and ii) for the first time, our method shows quality comparable to (or better than) \texttt{Identity}.

In summary, our contributions are as follows: i) We design a score-based generative model for tabular data synthesis. ii) We alleviate the training difficulty of the denoising score matching loss by designing a self-paced learning strategy and further enhance the sampling quality and diversity using a proposed fine-tuning method. \texttt{STaSy}, thus, clearly balances among \textit{the generative learning trilemma}: sampling quality, diversity, and time. iii) Our proposed method outperforms other deep learning methods in all cases by large margins, which we consider a significant advance in the field of tabular data synthesis.
% Moreover, \texttt{STaSy} is faster than many other methods, which we consider a significant advance in the field of tabular data synthesis. 
iv) We evaluate various methods in terms of \textit{the generative learning trilemma} in a rigorous manner.

% \chaejeong{, we measure their diversity for the first time to our knowledge, using coverage and various visualizations.}}
% We measure diversity of tabular data to evaluate them in terms of many aspects for the first time to our knowledge, using coverage and various visualizations.

% \jayoung{Our contributions are as follows: 
% \textbf{Tabular synthesis model based on SGMs:} We adopt the score-based generative model on tabular data synthesis modeling and customize the score-network which is suitable for tabular data synthesis. 
% \textbf{SPL and fine-tuning framework:} We alleviate training difficulties from irregular distribution of the training loss stem from multi-modality of tabular data by adapting curriculum learning strategy and proposed fine-tuning method which can improve the overall likelihood of training samples take advantages of diffusion process.  
% \textbf{Breakthrough performance in synthesizing tabular data:} Our proposed method not only outperforms other deep learning method but \texttt{Identity} in some cases. 
% }

\section{Related work}
\subsection{Score-based generative models}

Score-based generative models (SGMs) use a diffusion process defined by the following It\^o stochastic differential equation (SDE):
% Score-based generative models (SGMs) use the following It\^o stochastic differential equation (SDE) to define the diffusion process:
\begin{align}\label{eq:forward}
d\mathbf{x}=\mathbf{f}(\mathbf{x},t)dt + g(t)d\mathbf{w},
\end{align}where $\mathbf{f}(\mathbf{x},t) = f(t)\mathbf{x}$, $f$ and $g$ are drift and diffusion coefficients of $\mathbf{x}(t)$, and $\mathbf{w}$ is the standard Wiener process. Depending on the types of $f$ and $g$, SGMs can be divided into variance exploding (VE), variance preserving (VP), and sub-variance preserving (sub-VP) models~\citep{songyang}. The definitions of $f$ and $g$ are in Appendix~\ref{fandg}. The reverse of the diffusion process is a denoising process as follows:% where $\mathbf{f}(\mathbf{x},t) = f(t)\mathbf{x}$, and its reverse SDE process is a denoising process as follows:
\begin{align}\label{eq:reverse}
d\mathbf{x}=\big(\mathbf{f}(\mathbf{x},t)-g^2(t)\nabla_\mathbf{x} \log p_t(\mathbf{x})\big)dt + g(t)d\mathbf{w},
\end{align} 
where this reverse SDE is a process of generating samples. % a generative process
The score function $\nabla_\mathbf{x} \log p_t(\mathbf{x})$ is approximated by a time-dependent score-based model $S_{\boldsymbol{\theta}}(\mathbf{x}, t)$, called \emph{score network}.
% According to the types of $f$ and $g$, SGMs can be classified as i) variance exploding (VE), ii) variance preserving (VP), and iii) sub-variance preserving (sub-VP) models~\citep{songyang}. The definitions of $f$ and $g$ are in Appendix~\ref{fandg}.

% We can construct a diffusion process in ~\Eqref{eq:forward} by a continuous time variable $t \in [0,T]$, such that $\mathbf{x}(0)$ means a real sample and $\mathbf{x}(T)$ means a noisy sample.
In general, following the diffusion process in~\Eqref{eq:forward}, we can derive $\mathbf{x}(t)$ at time $t \in [0,T]$, where $\mathbf{x}(0)$ and $\mathbf{x}(T)$ means a real and noisy sample, respectively.
% \jayoung{In general, $\mathbf{x}(0)$ means a real sample, and we can derive $\mathbf{x}(t)$ at time $t \in [0,T]$, following the diffusion process in~\Eqref{eq:forward}. }
The transition probability $p(\mathbf{x}(t)|\mathbf{x}(0))$ at time $t$ is easily approximated by this process, and it always follows a Gaussian distribution.
% This process easily approximates the transition probability $p(\mathbf{x}(t)|\mathbf{x}(0))$ at time $t$ with a Gaussian distribution.% The reverse SDE in Eq.~\eqref{eq:reverse} maps a noisy sample at $t=T$ to a data sample at $t=0$.
It allows us to collect the gradient of the log transition probability, $\nabla_{\mathbf{x}(t)} \log p(\mathbf{x}(t)|\mathbf{x}(0))$, during the diffusion process.
% We can collect the gradient of the log transition probability, $\nabla_{\mathbf{x}(t)} \log p(\mathbf{x}(t)|\mathbf{x}(0))$, during the diffusion process.
Therefore, we can train a score network $S_{\boldsymbol{\theta}}(\mathbf{x}, t)$ as follows:
% Therefore, we can train a score network $S_{\boldsymbol{\theta}}:\mathbb{R}^{\dim(\mathbf{x})} \times [0,T] \rightarrow \mathbb{R}^{\dim(\mathbf{x})}$ as follows, where $\dim(\mathbf{x})$ means the dimensionality of $\mathbf{x}$:
% e.g., the number of pixels in the case of images or the number of columns in the case of tabular data:
\begin{align}\label{eq:sgm}
    \argmin_{\boldsymbol{\theta}} \mathbb{E}_t \mathbb{E}_{\mathbf{x}(t)} \mathbb{E}_{\mathbf{x}(0)} \Big[\lambda(t) \|S_{\boldsymbol{\theta}}(\mathbf{x}(t), t) -\nabla_{\mathbf{x}(t)} \log p(\mathbf{x}(t)|\mathbf{x}(0)) \|_2^2 \Big],
\end{align}where $\lambda(t)$ is to control the trade-off between the sampling quality and likelihood. This is called \emph{the denoising score matching}, and $\boldsymbol{\theta}^*$ solving~\Eqref{eq:sgm} can accurately solve the reverse SDE in~\Eqref{eq:reverse}~\citep{10.1162/NECO_a_00142}.
% In other words, the optimal solution $\theta^*$ of the denoising score matching training is equivalent to the optimal solution of the exact score matching training~\citep{10.1162/NECO_a_00142}.

% There exist several other improvements on SGMs. For instance, the adversarial score matching model~\citep{jolicoeur2020adversarial} is to combine SGMs and GANs. Therefore, the reverse SDE is not solely determined as a reverse of the forward SDE but as a trade off between the denoising score matching and the adversarial training of GANs (cf. Fig.~\ref{fig:asm} (a)). 

% \begin{figure}
%     \centering
%     \subfigure[Adversarial score matching where the score network $S_\theta$ is trained with i) the denoising score matching loss in Eqs.~\eqref{eq:sgm} and ii) the adversarial training]{\includegraphics[width=0.9\columnwidth]{images/asm.pdf}}
%     \subfigure[Langevin corrector highlighted in purple]{\includegraphics[width=0.7\columnwidth]{images/lan.pdf}}
%     \caption{Two ideas further improving the synthesis quality of SGMs. (a) The adversarial score matching model combining SGMs and GANs. (b) The correction algorithm, called \emph{Langevin corrector}, where $\hat{\mathbf{x}}'_0$ has a higher log probability than $\hat{\mathbf{x}}_0$ since we move following the gradient of the log probability.}
%     \label{fig:asm}
% \end{figure}

After the training process, we can synthesize fake data records with i) the \emph{predictor-corrector} framework or ii) the \emph{probability flow} method, a deterministic method based on the ordinary differential equation (ODE) whose marginal distribution is equal to that of ~\Eqref{eq:forward}~\citep{songyang}. In particular, the latter enables fast sampling and exact log-probability computation.

% In the prediction phase, we solve the reverse SDE in Eq.~\eqref{eq:reverse} after sampling a Gaussian noise. Let $\hat{\mathbf{x}}_0$ be a solution of the reverse SDE. In the correction phase, we then enhance it by using the Langevin corrector~\citep{songyang}. 
% % {\color{purple} The key idea of the correction phase is in Fig.~\ref{fig:asm} (b). 
% Since we correct the predicted sample $\hat{\mathbf{x}}_0$, following the direction of the gradient of the log probability, the corrected sample has a higher log probability. 
% %However, only the VP and VE models use this correction mechanism whereas the sub-VP model does not.

% \vspace{-0.2em}
\subsection{Tabular data synthesis}\label{sec:tabular_data_synthesis}
Many distinct methods exist for tabular data synthesis, which creates realistic synthetic tables depending on the data types. 
% by modeling a joint probability distribution of columns in a table. 
For example, a recursive table modeling utilizing a Gaussian copula is used to synthesize continuous variables~\citep{7796926}. Discrete variables can be generated by Bayesian networks~\citep{10.1145/3134428, avino2018generating} and decision trees~\citep{article}. 
% The spatial data is generated using a differentially private decomposition process~\citep{DBLP:journals/corr/abs-1103-5170, DBLP:journals/corr/ZhangXX16}. Due to model limitations such as specific types of distributions or computational issues, high-fidelity data synthesis is challenging.
Several data synthesis methods based on GANs have been presented to generate tabular data in recent years. RGAN~\citep{esteban2017realvalued} creates continuous time-series healthcare records, whereas \texttt{MedGAN}~\citep{DBLP:journals/corr/ChoiBMDSS17} and corrGAN~\citep{DBLP:journals/corr/abs-1804-00925} generate discrete records. EhrGAN~\citep{DBLP:journals/corr/abs-1709-01648} utilizes semi-supervised learning to generate plausible labeled records to supplement limited training data. PATE-GAN~\citep{Jordon2019PATEGANGS} generates synthetic data without jeopardizing the privacy of real data. \texttt{TableGAN}~\citep{park2018data} employs convolutional neural networks to enhance tabular data synthesis and maximize label column prediction accuracy. \texttt{CTGAN} and \texttt{TVAE}~\citep{NIPS2019_8953} adopt column-type-specific preprocessing steps to deal with multi-modality in the original dataset distribution. \texttt{OCT-GAN}~\citep{10.1145/3442381.3449999} is a generative model design based on neural ODEs. \texttt{SOS}~\citep{sos} proposed a style-transfer-based oversampling method for imbalanced tabular data using SGMs, whose main strategy is converting a major sample to a minor sample. Since its task is not compatible to our task to generate from the scratch, direct comparisons are not possible. However, we convert our method to an oversampling method following their design guidance and compare with \texttt{SOS} in Appendix~\ref{sec:oversampling}.

% \vspace{-0.2em}
\subsection{Self-paced learning}\label{sec:spl}
Self-paced learning (SPL) is a training strategy related to curriculum learning to select training records in a meaningful order, inspired by the learning process of humans~\citep{kumar2010self, jiang2014self}. It refers to training a model only with a subset of data that has low training losses and gradually expanding to the entire training data. We denote the training set as $\mathcal{D} = \{\mathbf{x}_i\}^{N}_{i=1}$, where $\mathbf{x}_i$ is the $i$-th record.
% --- for simplicity but without loss of generality, we assume classification tabular data. 
The model $M$ with parameters $\boldsymbol{\theta}$ has a loss $l_{i} = L(M(\mathbf{x}_{i}, \boldsymbol{\theta}))$, where $L$ is the loss function. 
A vector $\mathbf{v} = [v_i]^{N}_{i=1}, v_i \in \{0, 1\} $ indicates whether $\mathbf{x}_{i}$ is easy or not for all $i$. SPL aims to learn the model parameter $\boldsymbol{\theta}$ and the selection importance $\mathbf{v} $ by minimizing:
\begin{align}\label{eq:spl}
    \min_{\boldsymbol{\theta},\mathbf{v}}\mathbb{E}(\boldsymbol{\theta}, \mathbf{v}) = \sum_{i=1}^N v_{i}L(M(\mathbf{x}_{i},\boldsymbol{\theta})) - \frac{1}{K} \sum_{i=1}^N v_i,
    % +g(\mathbf{v};\lambda),
\end{align} where $K$ is a parameter to control the learning pace. In general, the second term in~\Eqref{eq:spl}, called a self-paced regularizer, can be customized for a downstream task. 

% To solve Eq.~\eqref{eq:spl}, the alternative convex search (ACS)~\citep{acs} is generally used~\citep{spl, 10.5555/2999134.2999206}. ACS is an iterative method for biconvex optimization. 
The alternative convex search (ACS)~\citep{acs} is typically used to solve~\Eqref{eq:spl}~\citep{spl, 10.5555/2999134.2999206}. By alternately optimizing variables while fixing others, we can optimize~\Eqref{eq:spl}, i.e., update $\mathbf{v}$ after fixing $\boldsymbol{\theta}$, and vice versa. With fixed $\boldsymbol{\theta}$, the global optimum $\mathbf{v}^* = [v^*_i]^N_{i=1}$ is defined as follows:
\begin{align}
    v^*_{i}&=\begin{cases}1,  & l_{i}<\frac{1}{K},\\
   0, & l_{i}\geq \frac{1}{K}, \\
    \end{cases}
    \label{eq:v}
\end{align}
% It works by iteratively solving a biconvex optimization, which selects easy samples while updating the parameters. 
% It alternates between optimizing $\boldsymbol{\theta}$ and $\mathbf{v}$ while keeping the other set of variables fixed. 

When updating $\mathbf{v}$ with fixed $\boldsymbol{\theta}$, a record $\mathbf{x}_i$ with $l_{i} < \frac{1}{K}$ is regarded as an easy record and will be chosen for training. Only easy records are used to train the model. Otherwise, $\mathbf{x}_{i}$ is regarded as a hard record and will be unselected. To involve more records in the training process, $K$ is gradually decreased.
% To learn new records, $K$ is gradually decreased during the process of the training. More records with higher losses should be gradually added to the training.
% The negative ll-norm regularizer is general and applicable to various tasks with different loss functions.

% In addition, SPL is known to be able to exclude noises from the training process because it allows the re-ordering of training records by training difficulty~\citep{whendocurriculawork, jiang2018mentornet, jiang2020beyond, guo2018curriculumnet}, which is one of the main reasons why the model becomes more stable with SPL.

\section{Proposed method}
\texttt{STaSy} is an SGM-based method for tabular data synthesis. \texttt{STaSy} uses SPL to ensure its training stability. The suggested fine-tuning method takes advantage of a favorable property of SGMs, which is that we can measure the log-probabilities of records.

% In this section, we depict the proposed score network and training framework for our \texttt{STaSy}.

\subsection{Score network architecture \& miscellaneous designs}
% \subsection{Miscellaneous Designs}
\label{archi}

It is known that each column in tabular data typically has complicated distributions, whereas pixel values in image datasets typically follow Gaussian distributions~\citep{NIPS2019_8953}. Moreover, tabular synthesis models should learn the joint probability of multiple columns to generate a record, which is one main reason why tabular data synthesis is difficult. However, one good design point is that the dimensionality of tabular data is typically far less than that of image data, e.g., 784 pixels even in MNIST, one of the simplest image datasets, vs. 30 columns in \texttt{Credit}.

We found through our preliminary experiments that the SDE in~\twoEqrefs{eq:forward}{eq:reverse} can well model the joint probability \emph{iff} its score network, which approximates $\nabla_\mathbf{x} \log p_t(\mathbf{x})$, is well trained. We carefully design our score network for tabular data synthesis considering these points. Our proposed score network architecture is in Appendix~\ref{networkarchi}.
Since SGMs were theoretically designed from the idea of perturbing data with an infinite number of noise scales, SGMs typically require large-scale computation, e.g., $T=1,000$ for images in~\citep{songyang}, as an approximation to the infinite number. With a large number of steps, the denoising process requires a long time to complete, which is one part of \textit{the generative learning trilemma}.
% \textcolor{blue}{One drawback of SGMs is that their long SDE steps require large-scale computation, e.g., $T=1,000$ for the images in~\citep{songyang}. With a large number of steps, the denoising process requires a long time to complete, which is one part of \textit{the generative learning trilemma}.} 
However, we found that $T=50$ steps in~\Eqref{eq:forward} are enough to train a network to approximate the gradient of the log-likelihood, which means that our \texttt{STaSy} naturally has less sampling time than SGMs for images with $T=1,000$ steps. 
% \textcolor{blue}{At the same time, it is also known that small $T$ values work well for time series forecasting~\citep{rasul2021autoregressive}.}

% \old{However, we found that $T=50$ steps in~\Eqref{eq:forward} are enough to map $\mathbf{x}(0)$ to a Gaussian noise, which means that our \texttt{STaSy} naturally has a less sampling time than SGMs for images with $T=1,000$ steps. At the same time, it is also known that small $T$ values work well for time series forecasting~\citep{rasul2021autoregressive}.}

% \chaejeong{To handle mixed types of data, we pre/post-process columns. Following the widely used preprocessing method in \texttt{CTGAN}~\citep{NIPS2019_8953}, we use the min-max scaler to pre-process numerical columns, and its reverse scaler is used for post-processing after generation. We also apply one-hot encoding to pre-process categorical columns~\citep{hancock2020survey,borisov2021deep}, and use the softmax function, followed by the rounding function, when generating.}

\noindent\textbf{\textbf{Pre/post-processing of tabular data}} To handle mixed types of data, which is a challenge in tabular data generation, we pre/post-process columns. We use the min-max scaler to pre-process numerical columns, and its reverse scaler is used for post-processing after generation. We also apply one-hot encoding to pre-process categorical columns, and use the softmax function, followed by the rounding function, when generating.

\noindent\textbf{\textbf{How to generate}} 
% \chaejeong{We use the min-max scaler to pre-process numerical columns, and its reverse scaler is used for post-processing after generation. We also apply one-hot encoding to pre-process categorical column, and use the softmax function.} 
After sampling a noisy vector $\mathbf{z} \sim \mathcal{N}(\boldsymbol{\mu},\sigma^2\mathbf{I})$, the reverse SDE can convert $\mathbf{z}$ into a fake record.  The prior distribution $\mathbf{z} \sim \mathcal{N}(\boldsymbol{\mu},\sigma^2\mathbf{I})$ varies depending on the type of SDEs: $\mathcal{N}(\mathbf{0}, \sigma^2_{max}\mathbf{I})$ for VE, and $\mathcal{N}(\mathbf{0}, \mathbf{I})$ for VP and sub-VP. $\sigma_{max}$ is a hyperparameter. In particular, we adopt the \emph{probability flow} method to solve the reverse SDE, which will be shortly described in~\Eqref{eq:pfode}.

\subsection{Self-paced learning approach}\label{sec:selfpacedlearning}

% Properly modeling tabular data is a non-trivial task because of multi-modality of tabular data. For example, Fig.~\ref{fig:loss_hist} depicts training data with a wide variety of training difficulty. Moreover, tabular data includes a number of samples with large training loss values, which are the most challenging samples. 
% i.e., whose training loss is greater than 0.4 in Fig.~\ref{fig:loss_hist}. 
% Previous models [ctgan, octgan] use mode-specific normalization to catch modes that are likely to exist.

In order to alleviate the training difficulty, we apply a curriculum learning technique for \texttt{STaSy}, more specifically, self-paced learning. Instead of letting $v_i \in \{0,1\}$, we use a ``soft'' record sampling method, i.e., $v_i \in [0,1]$.
% To select records to train the model with the denoising score matching loss, we introduce \textcolor{red}{the importance vector} $\mathbf{v}=[v_1, v_2, \dots, v_N]$, which indicates whether $i$-th record is easy or not. 
If $l_i$, which is the denoising score matching objective on $i$-th record, is less than a threshold, we set $v_i$ to 1 to ensure that the record is fully involved in training. At the end of the training, $v_i$ must be set to 1 for all $i$ to train the model with the entire data. The denoising score matching loss for $i$-th training record $\mathbf{x}_i$ is defined as follows:
\begin{align}\label{eq:L_i}
l_i = \mathbb{E}_t \mathbb{E}_{\mathbf{x}_i(t)} \Big[\lambda(t) \|S_{\boldsymbol{\theta}}(\mathbf{x}_i(t), t) -\nabla_{\mathbf{x}_i(t)} \log p(\mathbf{x}_i(t)|\mathbf{x}_i(0)) \|_2^2 \Big].
\end{align}
Then, we have the following \texttt{STaSy} objective:
\begin{align}\label{eq:optimize}
    \underset{\boldsymbol{\theta}, \mathbf{v}}{min} \sum_{i=1}^{N} v_i l_i + r(\mathbf{v}; \alpha, \beta),
\end{align}where $0 \le v_i \le 1$ for all $i$, $r(\cdot)$ is a self-paced regularizer. $\alpha \in [0, 1]$ and $\beta \in [0, 1]$ are variables to define thresholds, which are monotonically increasing as training goes on (see Appendix~\ref{sec:threshold} for their exact controlling mechanism). 

\begin{definition}[]
    Let $Q(p)$ be a quantile function defined as $\inf\{l \in \mathbb{R} : p \le F(l)\}$, where $F$ is a cumulative distribution function of the denoising score matching objective. 
    % \jayoung{That is, $p$ is the probability that $l$ will take a value less than or equal to $Q(p)$.}
    That is, $Q(p)$ is the minimum value for which the CDF is greater than or equal to the given probability $p$.
\end{definition}

\begin{theorem}\label{theorem:v}
    Let the self-paced regularizer $r(\mathbf{v}; \alpha, \beta)$ is defined as follows:
    \begin{align}
        r(\mathbf{v}; \alpha, \beta) = -\frac{Q(\alpha) - Q(\beta)}{2} \sum_{i=1}^N v_i^2 - Q(\beta) \sum_{i=1}^N v_i,
    \end{align}where the closed-form optimal solution for $\mathbf{v}^* = [v_1^*, v_2^*, \dots, v_N^*]$, given fixed $\boldsymbol{\theta}$, is defined as follows --- its proof is in Appendix~\ref{sec:derivation}:
    \begin{align}\label{eq:vstar}
    v_{i}^* = \begin{dcases*}
        1,                                                            & if $l_i \le Q(\alpha)$,\\
        0,                                                            & if $l_i \ge Q(\beta)$,\\
        \frac{l_i - Q(\beta)}{ Q(\alpha)- Q(\beta)} ,  & otherwise.
        \end{dcases*}
\end{align} 
\end{theorem}

% The goal is to jointly learn the model parameter $\theta$ and the weight variable $\mathbf{v}$ by minimizing ~\eqref{eq:optimize}. Utilizing ACS, we optimize two terms in ~\eqref{eq:optimize} iteratively. 
% First, we fix $\mathbf{v}$ and optimize $\theta$ with score matching [cite], and when $\theta$ is fixed, we optimize a self-paced regularizer $g(\cdot)$. Considering $v_i \in [0, 1]$, we get the closed-formed optimal solution for $\mathbf{v}^* = [v_1, v_2, \dots, v_N]$ as follows:

% \begin{align}
%     v_{i} = \begin{dcases*}
%         1,                                                            & if $l_i \le Q(\alpha)$,\\
%         0,                                                            & if $l_i \ge Q(\beta)$,\\
%         \frac{l_i - Q(\beta)}{ Q(\alpha)- Q(\beta)} ,  & otherwise.
%         \end{dcases*}
% \end{align} 

% \begin{wrapfigure}[10]{r}{0.35\linewidth}
%   \begin{center}
%       \raisebox{0pt}[\dimexpr\height-2\baselineskip\relax]{%
%         \includegraphics[trim={0.5cm 0.5cm 0.5cm 0},clip, width=0.33\textwidth]{images/linear.pdf}
%     }
%   \end{center}
%   \caption{Various types of SPL regularize function $g(\cdot)$}
%   \label{fig:v_function}
% \end{wrapfigure}
Specifically, records with $l_i \le Q(\alpha)$ are considered easy records and will be selected for training, whereas records with $l_i \ge Q(\beta)$ are considered complicated (or potentially noisy) and will not be selected. If not both cases, records will be partially selected during training, i.e., $v_i \in [0,1]$. $\alpha$ and $\beta$ are gradually increased to 1 from the initial values $\alpha_0$ and $\beta_0$, proportionally to training progress to ensure that all data records are involved in training. As $\alpha$ and $\beta$ increase, the difficult records are gradually involved in training, and the model also becomes more robust to those difficult cases. 
% To ensure that all data records in training are selected at the end of the training, the thresholds, $\alpha$ and $\beta$, gradually increase to 1. 
% To ensure that all data records are involved in training, $\alpha$ and $\beta$ gradually increase to 1, proportionally to training progress. 
% Starting from the initial values $\alpha_0$ and $\beta_0$, $\alpha$ and $\beta$ increase proportionally to training progress. 
We set $\alpha_0$ and $\beta_0$ in such a way that more than 80\% of the training records are included in the learning process from the beginning.
% ~\ref{alg1} shows a detailed training process for optimizing Eq.~\eqref{eq:spl}.

% The function $g(\cdot)$ can be defined as logarithmic, exponential, or linear, as shown in Fig.~\ref{fig:v_function}, but there is no significant difference between them in our preliminary experiment. Thus our main experiments are limited to linear functions.
% \vspace{-3pt}
% \vspace{-5pt}
% \begin{wrapfigure}[8]{L}{0.6\textwidth}
    % \raisebox{0pt}[\dimexpr\height-1\baselineskip\relax]{
    % \begin{minipage}{0.6\textwidth}
    \begin{algorithm}[t]
    % \footnotesize
    
    \DontPrintSemicolon
    \caption{How to train \texttt{STaSy}}\label{alg1}

      Initialize $\boldsymbol{\theta},\mathbf{v}$ \;
      \tcc{Train SGM based on our SPL training strategy}
      \For{each mini-batch of records}{

            Update $\boldsymbol{\theta}$ after fixing $\mathbf{v}$ with~\Eqref{eq:optimize}\;

            Update $\mathbf{v}$ with~\Eqref{eq:vstar}\;
            Update $\alpha$ and $\beta$ with the control method in Appendix~\ref{sec:threshold}\;
      }
      
      \tcc{Fine-tune the trained model using log-probability}
      
    %   \FOR {$i = 1$ to $N$}{
    %     \lambda_i = \log p(x_i)
    %   }
    % $\lambda = \log p(\mathbf{x})$ \\
     $\tau_i \gets \log p(\mathbf{x}_i)$ \label{a:init} \;
     $\mathcal{F} \gets \{ \mathbf{x}_i | \log p(\mathbf{x}_i)\textrm{, where $\mathbf{x}_i \in \mathcal{D}$, is smaller than the average (or median) log-probability}.\}$ \label{a:set} \;
      \For {each fine-tune epoch}{
      \For {each $\mathbf{x}_i \in \mathcal{F}$}{ 
        Update $\boldsymbol{\theta}$ with~\Eqref{eq:L_i} \label{a:fine}  \;}
    % Compute $\log p(\mathbf{x}_i)$ \;
    $\mathcal{F} \gets \{ \mathbf{x}_i | \log p({\mathbf{x}}_i) < {\tau}_i\}$ \label{a:set2} 
    }
    
    \Return $\boldsymbol{\theta}$
     
    \end{algorithm} 

    % \vspace{-0.5em}
    % \end{minipage}
    % }
% \end{wrapfigure}
% \vspace{-0.8em}
\subsection{Fine-tuning approach}\label{sec:finetune}
% \vspace{-2pt}
For solving the reverse SDE process, score-based generative models rely on various numerical approaches. One of the techniques is the \emph{probability flow} method in~\Eqref{eq:pfode}~\citep{songyang}, which uses a deterministic process whose marginal probability is the same as the SDE. With the approximated score function $S_{\boldsymbol{\theta}}(\cdot)$, the \emph{probability flow} method uses the following neural ordinary differential equation (NODE) based model~\citep{node}:
% \vspace{-10pt}
\begin{align}\label{eq:pfode}
    d\mathbf{x}=\big(\mathbf{f}(\mathbf{x},t)-{1 \over 2}g(t)^2\nabla_\mathbf{x} \log p_t(\mathbf{x})\big)dt.
\end{align}

In our experiments, the \emph{probability flow} shows better quality than other methods to solve the original reverse SDE, and our default solver is the \emph{probability flow} (see Section~\ref{sec:sen}). In addition, NODEs facilitate computing the log-probability defined in~\Eqref{eq:pfode} through the instantaneous change of variables theorem. Consequently, we can calculate the exact log-probability efficiently with the unbiased Hutchinson's estimator~\citep{hutchinson, grathwohl2018ffjord}. Thus, we propose to fine-tune based on the exact log-probability.

% Leveraging this property, we improve the model even further with our proposed fine-tuning method.
% the estimated log-probability.
After learning the model parameter $\boldsymbol{\theta}$ as described in Section~\ref{sec:selfpacedlearning}, we set the sample-wise threshold $\tau_i$ to $\log p(x_i)$ (cf. Line~\ref{a:init} of Algorithm~\ref{alg1}). We then prepare the fine-tuning candidate set $\mathcal{F}$ (cf. Line~\ref{a:set} of Algorithm~\ref{alg1}). After fine-tuning for the samples in $\mathcal{F}$, we update the candidate set (cf. Line~\ref{a:set2} of Algorithm~\ref{alg1}). Our goal is for achieving a better log-probability than the initial one $\tau_i$ before the fine-tuning process.

% calculate the log-probability of $x_i$ for the entire training data. For the first epoch of the fine-tuning, we set the sample-wise fine-tuning threshold $\tau_i$ to the average (or median) of the log-probabilities (cf. Line~\ref{a:init} of Algorithm~\ref{alg1}), and we retrain with every record $x_i$ whose $\log p(x_i)$ is less than $\tau_i$ (cf. Line~\ref{a:fine} of Algorithm~\ref{alg1}). We update $\tau_i$ to the most recent log-probability (cf. Line~\ref{a:update} of Algorithm~\ref{alg1}).

% After learning the model parameter $\boldsymbol{\theta}$ as described in Section~\ref{sec:selfpacedlearning}, we calculate the log-probability of the entire training data. We retrain with the records whose log-probabilities are less than a threshold $\tau$, e.g., a mean or median of the log-probability values.

% \chaejeong{As shown in Figure~\ref{fig:likelihood} (Right), the majority of the testing records' log-probabilities are increased after the fine-tuning step. }
\vspace{-0.8em}
\subsection{Training algorithm}
\vspace{-0.5em}
Algorithm~\ref{alg1} shows the overall training process for our \texttt{STaSy}. Firstly, we initialize the parameters of the score network $\boldsymbol{\theta}$ and the selection importance $\mathbf{v}$. Utilizing ACS, we then train \texttt{STaSy} with the SPL training strategy. At this step, we iteratively optimize $\boldsymbol{\theta}$ and $\mathbf{v}$. We can obtain an optimal $\boldsymbol{\theta}$ by optimizing~\Eqref{eq:optimize} with fixed $\mathbf{v}$, and the global optimum $\mathbf{v}$ is calculated by~\Eqref{eq:vstar}. 
% When $\mathbf{v}$ is fixed, we can obtain an optimal $\boldsymbol{\theta}$ by optimizing Eq.~\eqref{eq:optimize} \jayoung{with fixed $\textbf{v}$}, and with fixed $\boldsymbol{\theta}$, the global optimum $\mathbf{v}$ is calculated by Eq.~\eqref{eq:vstar}. \jayoung{We note that optimizing Eq.~\eqref{eq:optimize} with fixed $\boldsymbol{\theta}$ is equivalent to calculating $\mathbf{v} $ with Eq.~\eqref{eq:vstar}.} 
We also update $\alpha$ and $\beta$ in proportion to training progress. After finishing the main SPL training step, we can generate fake records from the model. However, we further improve the score network in the fine-tuning step. We retrain our model for every record $\mathbf{x}_i$ whose log-probabilities are less than its threshold $\tau_i$. At Line \ref{a:fine} of Algorithm~\ref{alg1}, one can use the log-probability instead of~\Eqref{eq:L_i} as a fine-tuning objective. However, we found that \Eqref{eq:L_i} is more effective (see Appendix~\ref{sec:exactlikelihood}).
% the generalizability in enhancing the sampling quality (see Appendix~\ref{sec:exactlikelihood}).}
\vspace{-0.8em}
% it provides much faster training 
\section{Experiments}\label{sec:experiments}
\vspace{-0.5em}
We analyze various methods in terms of \textit{the generative learning trilemma}. We list only aggregated results over all the tested datasets in the main paper, but in Appendix~\ref{sec:additional_scores} detailed results with their mean and std. dev. scores from 5 different executions are reported.
\vspace{-0.5em}
% We used DecisionTree, AdaBoost, Logistic Regression, and MLP classifiers for the binary classification task; DecisionTree and MLP classifiers for the multiclass classification task; and Linear regression and MLP regressor for the regression task.

\subsection{Experimental environments}\label{sec:environments}
A brief description of our experimental environments is as follows: i) We use 15 real-world tabular datasets for classification and regression and 7 baseline methods. To be specific about our models, \texttt{Na\"ive-STaSy} is a naive conversion of SGMs without our proposed training strategies, and \texttt{STaSy} is trained with self-paced learning and the fine-tuning method. ii) In general, we follow the ``train on synthetic, test on real (TSTR)'' framework~\citep{esteban2017realvalued, Jordon2019PATEGANGS}, which is a widely used evaluation method for tabular data~\citep{NIPS2019_8953,  10.1145/3442381.3449999, lee2021invertible}, to evaluate the quality of sampling --- in other words, we train various models, including DecisionTree, AdaBoost, Logistic/Linear Regression, MLP classifier/regressor, RandomForest, and XGBoost, with fake data and test them with real data. For \texttt{Identity}, we train with real training data and test with real testing data, whose score can be a criterion to evaluate the sampling quality of various generative methods in a dataset. % To be more specific, we use DecisionTree, AdaBoost, Logistic Regression, and MLP classifiers for the binary classification task; DecisionTree and MLP classifiers for the multiclass classification task; and Linear regression and MLP regressor for the regression task.} 
 iii) We use various metrics to evaluate in various aspects. For the sampling quality, we mainly use average F1 for classification, and also report AUROC and Weighted-F1. We use $R^2$ and RMSE for regression. % Moreover, since SGMs and flow-based generative models can compute the exact log-probability of records, we also use them to compare the sampling quality.
 For the sampling diversity, we use coverage~\citep{naeem2020reliable}, which was proposed to measure the diversity of generated records. Full results are in Appendix~\ref{sec:additional_scores}. Detailed environments and hyperparameter settings are in Appendix~\ref{sec:apd_environments} and~\ref{sec:hyperparameters}, respectively.

\subsection{Experimental results}\label{sec:experimental_results}

\begin{table}[t]
\caption{Classification/regression with real data. We report average F1 (resp. macro F1), AUROC, and Weighted-F1 for binary (resp. multi-class) classification, and $R^2$ and RMSE for regression. The best (resp. the second best) results are highlighted in bold face (resp. with underline).}
\centering
\label{tbl:main}
\begin{tabular}{lcccccccccc}
\toprule
\multirow{2}{*}{Methods} && \multicolumn{3}{c}{Classification} && \multicolumn{2}{c}{Regression} & \\ \cmidrule{3-5} \cmidrule{7-8}
 &  & F1   &    AUROC   & Weighted-F1 && $R^2$ & RMSE &  \\ 
 \midrule
\texttt{Identity} && 0.806  & 0.934  & 0.818  && 0.366  & 4109.424  &\\  \midrule
\texttt{MedGAN} && 0.252  & 0.628  & 0.243  && -inf & inf &\\
\texttt{VEEGAN} && 0.371  & 0.704  & 0.356  && -2.369  & 8540.780 & \\
\texttt{CTGAN} && 0.639  & 0.849  & 0.649  && 0.113  & 4096.279 & \\ 
\texttt{TVAE}& & 0.603  & 0.839  & 0.608  && 0.107  & 4218.457 & \\ 
\texttt{TableGAN}& & 0.582  & 0.836  & 0.586 & & -0.026  & 4535.514 & \\ 
\texttt{OCT-GAN}& & 0.630  & 0.865  & 0.637  && 0.160  & 4105.729 & \\ 
\texttt{RNODE} && 0.545  & 0.839  & 0.550  && 0.130  & 4112.928 & \\  \midrule
% \texttt{CLBN} && 0.719  & 0.854  & 0.614  && -inf & inf &\\ 
% \texttt{CART} && 0.722  & 0.904  & 0.750  && 0.299  & 4156.787  &\\ 
\texttt{Na\"ive-STaSy} && \underline{0.777}  & \underline{0.922}  & \underline{0.786}  && \underline{0.324} & \underline{4076.596} & \\ 
\texttt{STaSy}&& \textbf{0.792}  & \textbf{0.925}  & \textbf{0.801}  && \textbf{0.350}  & \textbf{4071.643} & \\ 
\bottomrule
\end{tabular}
\vspace{-1em}
\end{table}

\subsubsection{Sampling quality}
Table~\ref{tbl:main} summarizes the key results on the sampling quality. We use task-oriented metrics, such as F1, $R^2$, and so on, under the TSTR evaluation framework. \texttt{MedGAN} and \texttt{VEEGAN}, two early GAN-based methods, show relatively lower test scores than other GAN-based methods, i.e., \texttt{CTGAN}, \texttt{TableGAN}, and \texttt{OCT-GAN}. In general, \texttt{CTGAN} and \texttt{OCT-GAN} show reliable quality among the baselines. However, our two score-based models always mark the best and the second best quality. 

% In particular, for \texttt{Credit}, which has a severe class imbalance problem, i.e., 99.8\% for class 0 and 0.2\% for class 1, almost all baselines fail to sample high-quality fake records, except \texttt{CTGAN} and \texttt{TableGAN}. 
Our methods, \texttt{Na\"ive-STaSy} and \texttt{STaSy}, significantly outperform all the baselines by large margins. In particular, our methods perform well in small datasets, e.g., \texttt{Crowdsource}, \texttt{Obesity}, and \texttt{Robot}, while other methods show poor quality, as shown in Table~\ref{tab:macro_f1} of Appendix~\ref{sec:additional_quality}. These multi-class classification datasets have a small number of per class record, e.g., the smallest class in \texttt{Crowdsource} has 79 records in the training set, which means that our methods are able to capture fine-grained modes from the original data. Moreover, in \texttt{Credit}, which has a severe class imbalance ratio of 99.7\% for class 0 and 0.3\% for class 1, more than half of the baselines failed to generate the minority class, showing an F1 score close to 0 in Table~\ref{tab:binary_f1} of Appendix~\ref{sec:additional_quality}. \texttt{CTGAN} and \texttt{TableGAN} also achieve a good F1 score close to that of \texttt{Identity}, but \texttt{Na\"ive-STaSy} and \texttt{STaSy} take the first and the second places again. 
% As shown in Appendix~\ref{sec:additional_scores}, \texttt{Na\"ive-STaSy} and \texttt{STaSy} even perform better than \texttt{Identity} on 6 of 15 datasets. 

% Regression is the most challenging task in our experiments, showing none of the methods, including \texttt{Na\"ive-STaSy} and \texttt{STaSy}, achieve better scores than those of \texttt{Identity}. However, our methods show the best quality again.
% show an outstanding generation quality in \texttt{Beijing} compared with others. % In \texttt{News}, \texttt{STaSy} is the only method that achieves a positive $R^2$ score.

% \texttt{CTGAN} and \texttt{TableGAN} show their effectiveness to some degree except for \texttt{Obesity}. In general, \texttt{OCT-GAN} and \texttt{CTGAN} show reliable quality among the baselines.

% Interestingly, \texttt{CART}, which is a traditional non-deep learning method performs the best score in some cases among baselines, i.e., \texttt{Obesity} and \texttt{News}, especially in \texttt{News}, \texttt{CART} outperforms all other methods.
% Among the baselines, \texttt{SGM} shows the best outcome in almost all cases, and in general, it shows robustness on its generation quality. 

% Putting it all together, our proposed score-based generative models, i.e., \texttt{Na\"ive-STaSy} and \texttt{STaSy}, show reasonable performance in all cases. Especially \texttt{STaSy} significantly outperforms all other methods in all datasets and shows even better test scores than \texttt{Identity} in most cases.

% \noindent\textbf{Log-probability.} 

\begin{wraptable}[8]{r}{0.43\textwidth}
\vspace{-1.5em}
\setlength\tabcolsep{1.8pt}
\small
% \captionsetup{justification=centering}
\caption{The median of the log-probabilities of testing records, averaged over all datasets}
\label{tab:likelihood}
\resizebox{0.4\textwidth}{!}{
\begin{tabular}{lccccccc}
        \toprule
        Methods & & Log-probability & \\
        \midrule
        \texttt{RNODE}       & & 59.327   &      \\
        \texttt{STaSy} w/o fine-tuning   & & 129.293   &    \\
        \texttt{STaSy}    & & \textbf{131.734}   &    \\ 
        \bottomrule
        \end{tabular}}
% \vspace{-1.2em}
\end{wraptable}

As flow-based generative models and SGMs with the \emph{probability flow} method can calculate the exact log-probability of records, we present the log-probability as another metric for the sampling quality. Table~\ref{tab:likelihood} shows the median of the log-probabilities of testing records, averaged over all datasets. Since the log-probability is not bounded, we take a median of them to handle the case of outliers. Our methods, even without the fine-tuning, show a much better log-probability than \texttt{RNODE}, which optimizes the log-probability as its objective. Moreover, the median log-probability of testing records even improves after the proposed fine-tuning method.

Putting it all together, our proposed score-based generative models, i.e., \texttt{Na\"ive-STaSy} and \texttt{STaSy}, show reasonable performance in all cases regarding the machine learning efficacy and log-probability. Furthermore, as shown in Tables~\ref{tbl:main} and \ref{tab:likelihood}, the sampling quality always improves with the proposed training strategies, i.e., the self-paced learning and the fine-tuning, which justifies their efficacy.

\subsubsection{Sampling diversity}

\begin{wraptable}[15]{r}{0.35\textwidth}
\vspace{-1.7em}
% \raisebox{0pt}[\dimexpr\height-1\baselineskip\relax]{
\setlength\tabcolsep{1pt}
\small
% \captionsetup{justification=centering}
\caption{Sampling diversity in terms of coverage averaged over all datasets}
\label{tbl:diversity}
\resizebox{0.32\textwidth}{!}{
\begin{tabular}{lcccc}
\toprule
        Methods  &  & Coverage &\\ 
        \midrule
        % \texttt{Identity} &   & 0.994&\\\midrule
        \texttt{MedGAN}    &   & 0.037 &\\
        \texttt{VEEGAN}       &    & 0.038  &\\
        \texttt{CTGAN}      &    & 0.352 &\\
        \texttt{TVAE}        &    & 0.494  & \\
        \texttt{TableGAN} &   & 0.434&\\
        \texttt{OCT-GAN}     &    & 0.381 &\\
        \texttt{RNODE}    &    & 0.328 & \\
        \midrule
        \texttt{Na\"ive-STaSy}   &    & 0.637  &\\
        \texttt{STaSy} w/o fine-tuning    &   & \underline{0.655}  &\\
        \texttt{STaSy} &   & \textbf{0.658}&\\
        \bottomrule
\end{tabular}
% }
}

\end{wraptable}

% \begin{wraptable}{r}{0.38\textwidth}
% \vspace{-1.7em}
% \setlength\tabcolsep{1.8pt}
% \small
% \caption{\new{Sampling diversity}}
% \label{tbl:diversity}
% \resizebox{0.38\textwidth}{!}{
% \begin{tabular}{lcccc}
% \toprule
%         Methods  & & Density & Coverage &\\ 
%         \midrule
%         \texttt{Identity} &  & 0.941  & 0.994&\\\midrule
%         \texttt{MedGAN}    &  & 0.279  & 0.036 &\\
%         \texttt{VEEGAN}       &  & 0.169  & 0.038  &\\
%         \texttt{CTGAN}      &  & 0.234  & 0.356 &\\
%         \texttt{TVAE}        &  & 0.553  & 0.494  & \\
%         \texttt{TableGAN} &  & 0.364  & 0.438&\\
%         \texttt{OCT-GAN}     &  & 0.284  & 0.379 &\\
%         \texttt{RNODE}    &  & 0.379  & 0.329 & \\
%         \midrule
%         \texttt{Na\"ive-STaSy}   &  & 0.603  & 0.637  &\\
%         \texttt{STaSy} w/o fine-tuning    &  & \underline{0.638} & \underline{0.654}  &\\
%         \texttt{STaSy} w/ fine-tuning    & & \textbf{0.648}  & \textbf{0.658}&\\
%         \bottomrule
% \end{tabular}
% }
% \vspace{-1.2em}
% \end{wraptable}

\begin{figure}[h]
\vspace{-0.8em}
        \centering
        \begin{subfigure}{\includegraphics[width=0.24\textwidth]{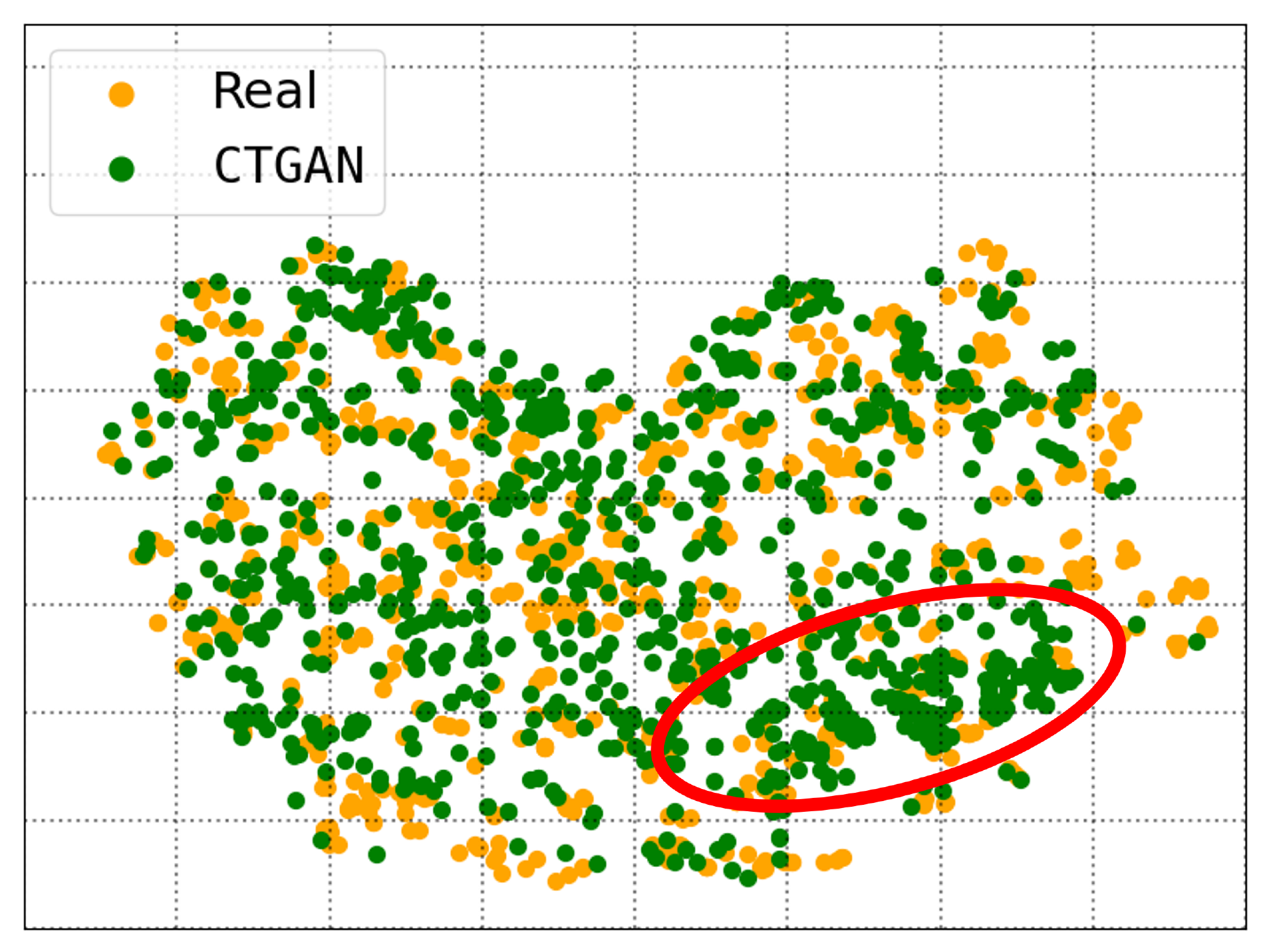}}
        \end{subfigure} \hfill
        \begin{subfigure}{\includegraphics[width=0.24\textwidth]{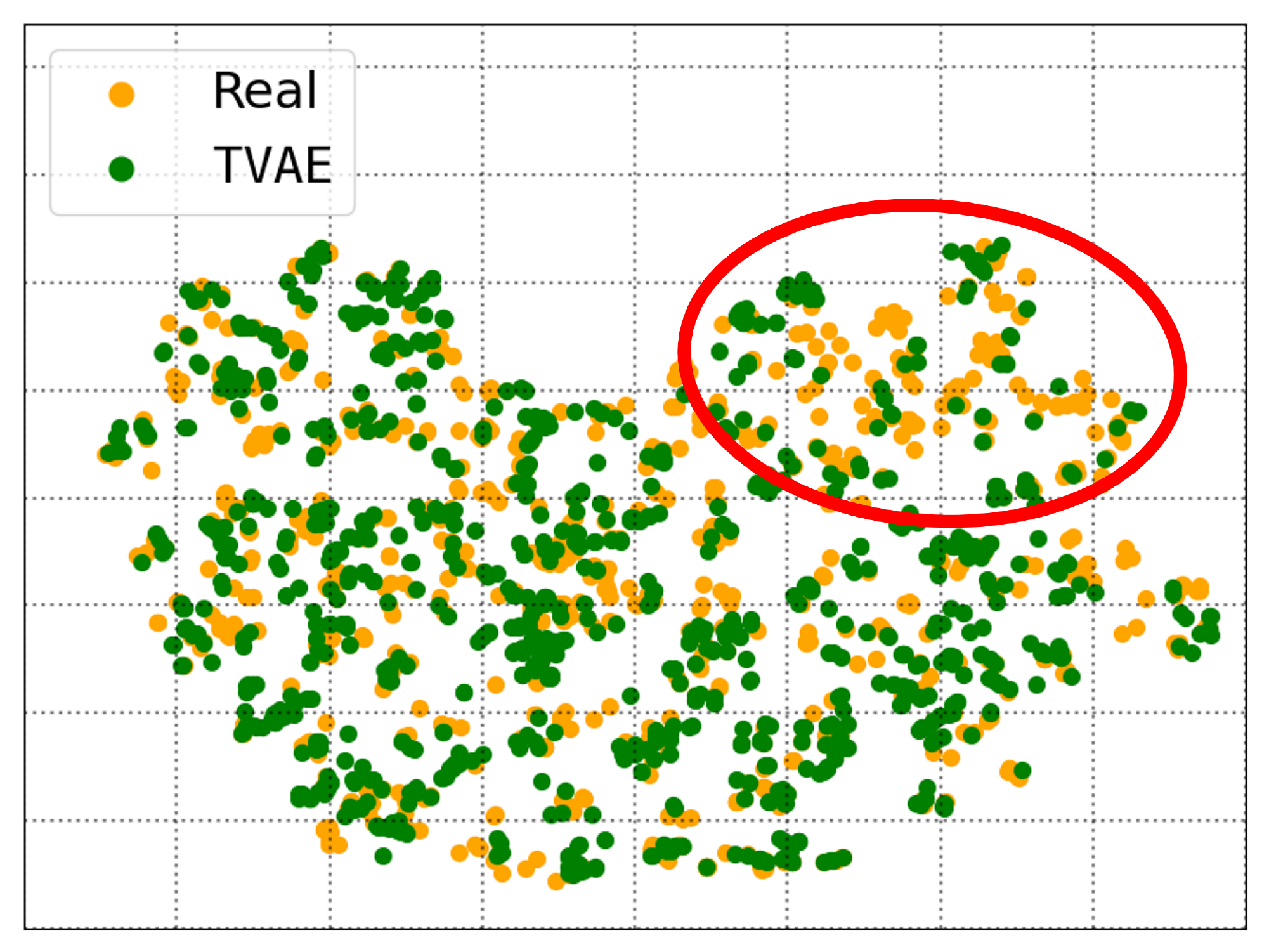}}
        \end{subfigure} \hfill
        \begin{subfigure}{\includegraphics[width=0.24\textwidth]{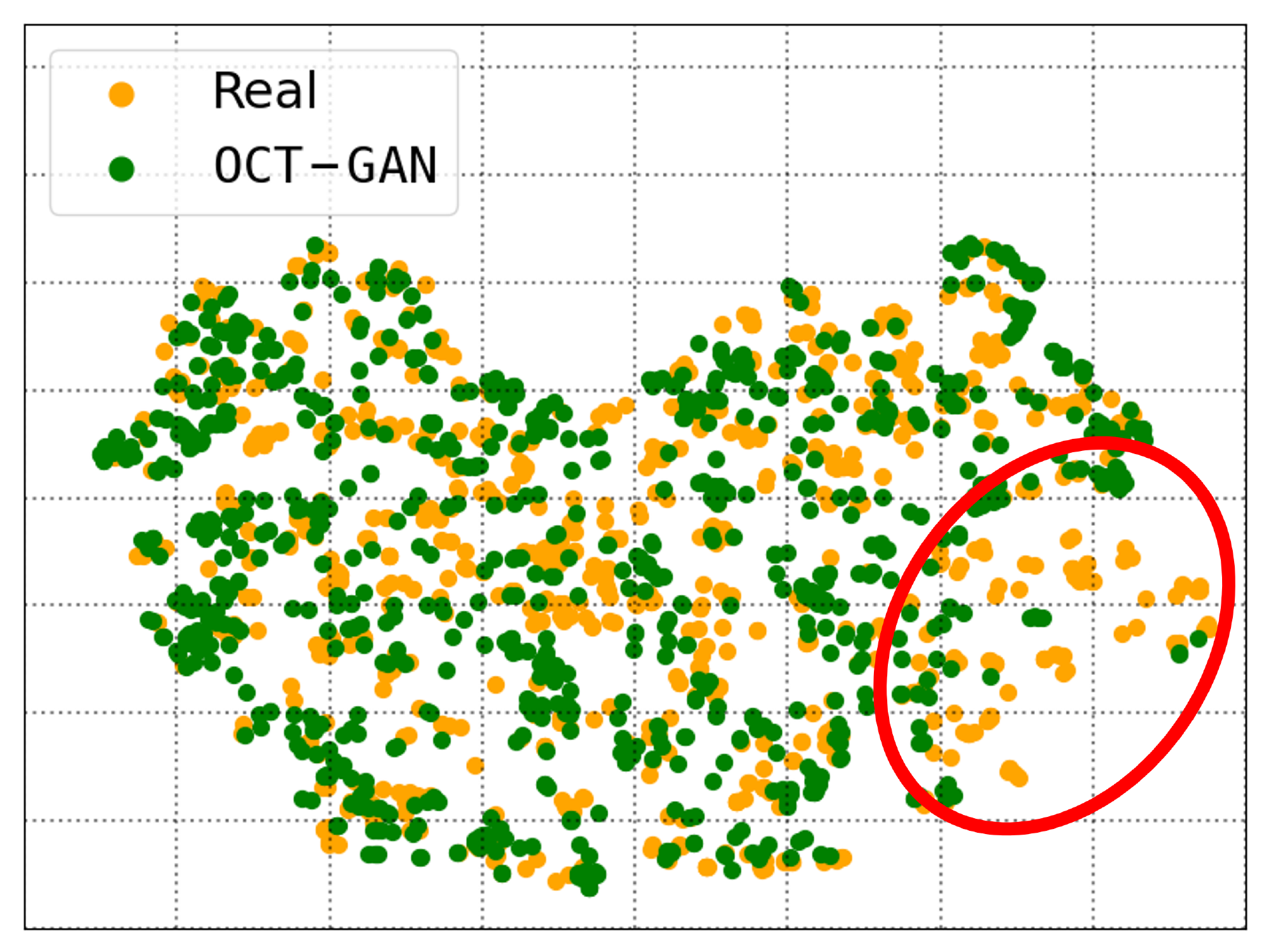}}
        \end{subfigure} \hfill 
        \begin{subfigure}{\includegraphics[width=0.24\textwidth]{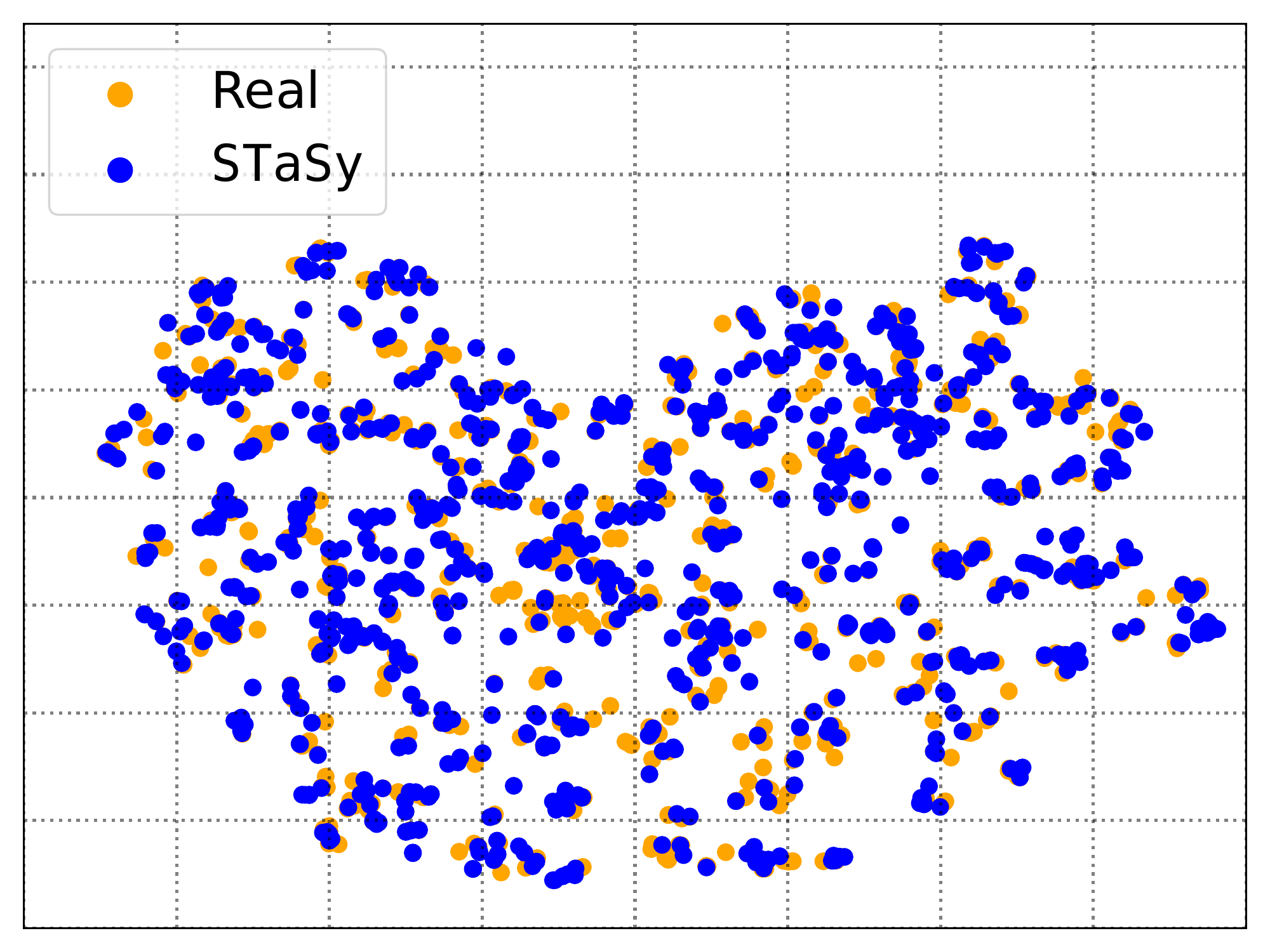}}
        \end{subfigure}
        \caption{t-SNE visualizations of fake and the original records in \texttt{Robot}.}
        \label{fig:robot_tsne}
\vspace{-1em}
\end{figure}

For the quantitative evaluation of the sampling diversity between existing methods and our proposed method, we use the coverage score~\citep{naeem2020reliable}, which is bounded between 0 and 1. Coverage is the ratio of real records that have at least one fake record in its manifold. A manifold is a sphere around the sample with radius $r$, where $r$ is the distance between the sample and the $k$-th nearest neighborhood. Table~\ref{tbl:diversity} summarizes the averaged coverage of each method. \texttt{MedGAN} and \texttt{VEEGAN} show poor coverage scores, close to 0. This trend is also shown in the t-SNE visualizations in Appendix~\ref{sec:visualization_robot}. Among the baseline methods, \texttt{CTGAN} performs the best in terms of the sampling quality, whereas it shows relatively inferior coverage performance than others. % In specific, as shown in Table~\ref{tab:coverage_full}, the top 3 baseline methods on \texttt{Robot} in terms of sampling quality, i.e., \texttt{CTGAN}, \texttt{TVAE}, and \texttt{OCT-GAN}, fail to show as good performance as they show in sampling quality in terms of diversity, showing less than 0.45 of coverage for all of them, while \texttt{STaSy} rates 0.95112.
In specific, in \texttt{Robot}, \texttt{STaSy} shows a coverage of 0.94, while other three top-performing baselines, \texttt{CTGAN}, \texttt{TableGAN}, and \texttt{OCT-GAN}, show coverage scores less than 0.26 in Table~\ref{tab:coverage_multi} of Appendix~\ref{sec:additional_diversity}. Figure~\ref{fig:robot_tsne} also presents the diversity of each fake data by each method qualitatively, which reflects the results of coverage. In general, \texttt{STaSy} shows stable performance across the sampling quality and the sampling diversity, outperforming others by large margins.

\begin{figure}[t]    
        \centering
        \begin{subfigure}{\includegraphics[width=0.32\textwidth]{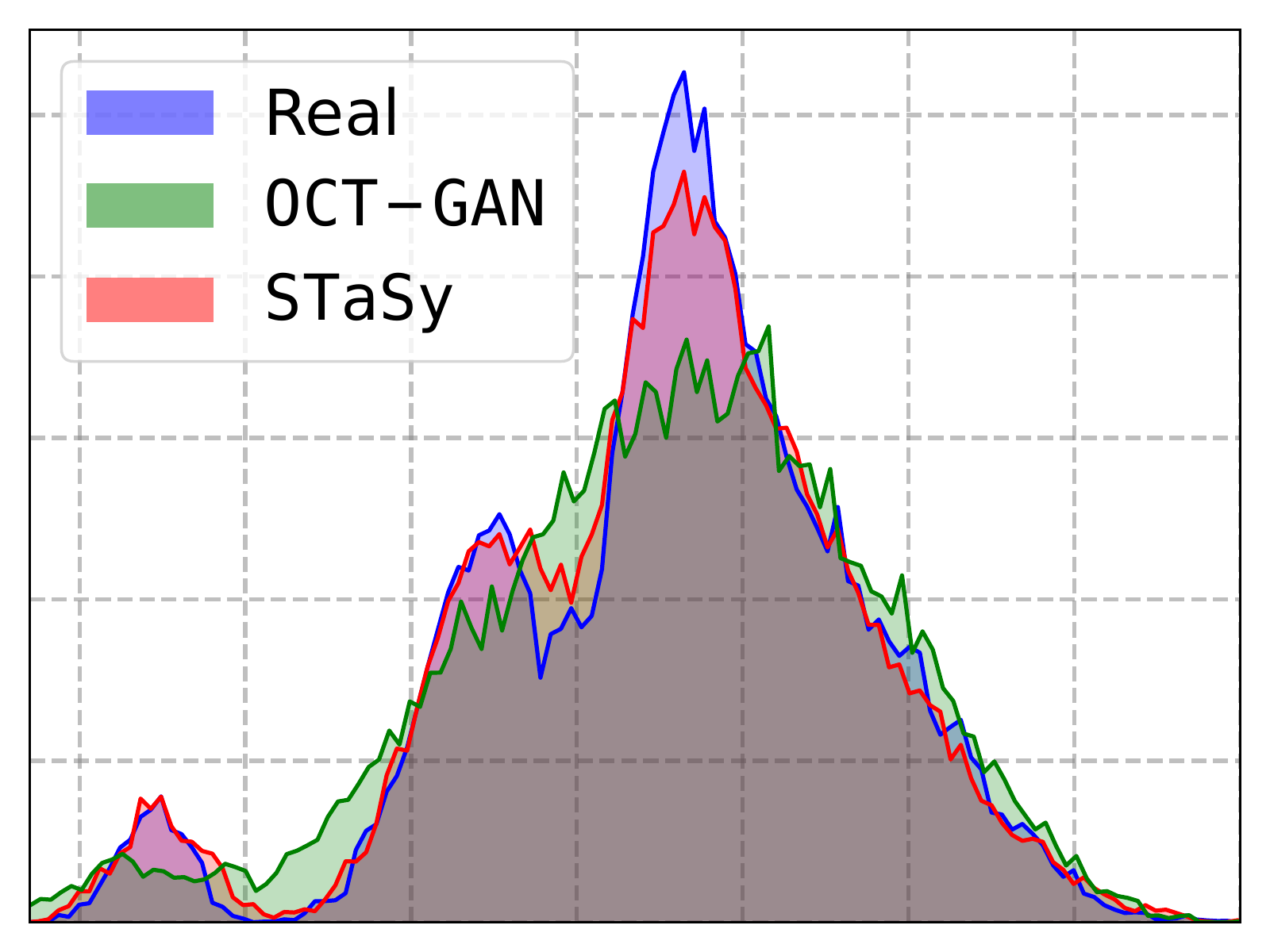}\label{fig:histogram_1}}
        \end{subfigure}
        \begin{subfigure}{\includegraphics[width=0.32\textwidth]{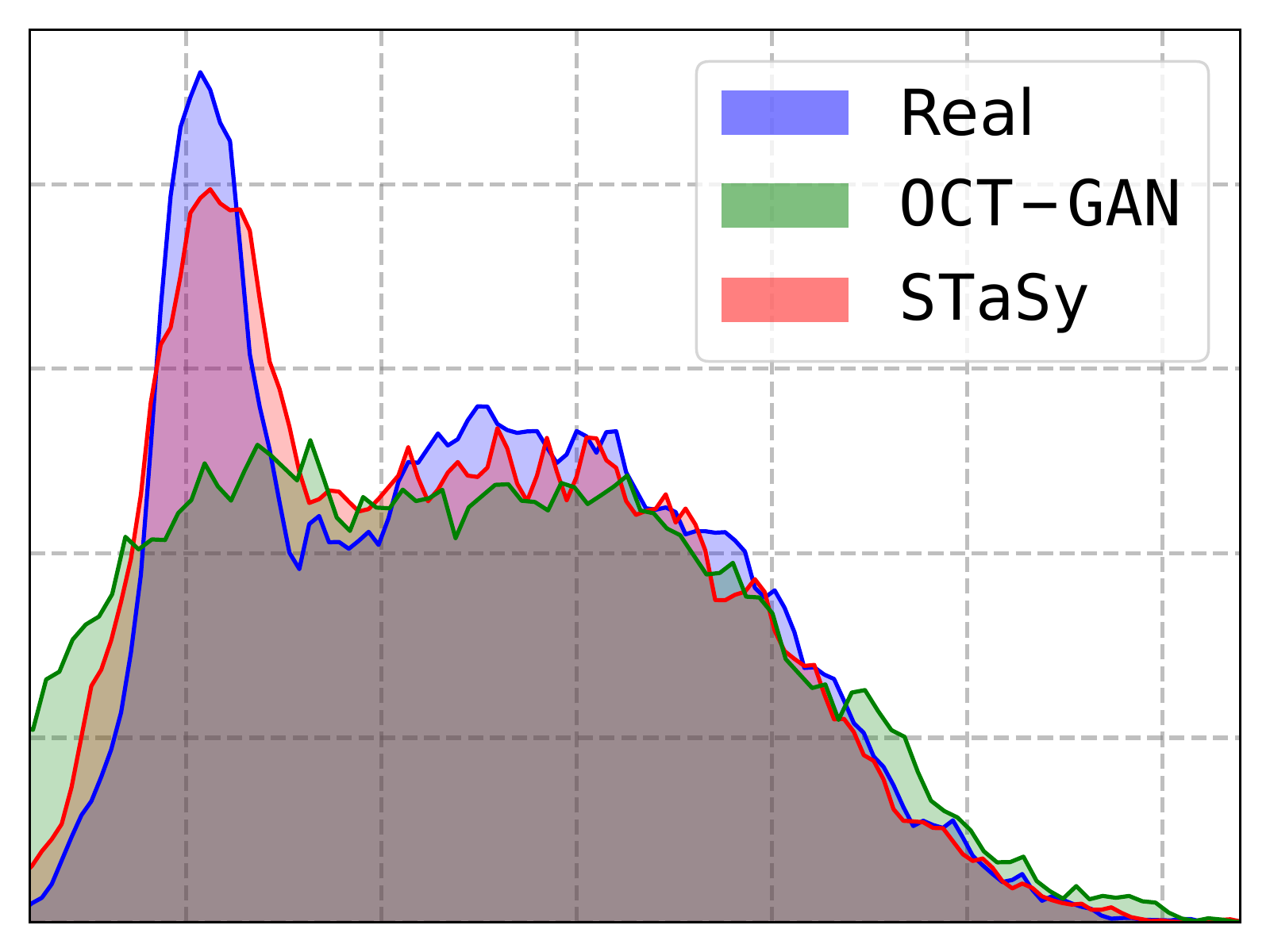}\label{fig:histogram_2}}
        \end{subfigure}
        \begin{subfigure}{\includegraphics[width=0.32\textwidth]{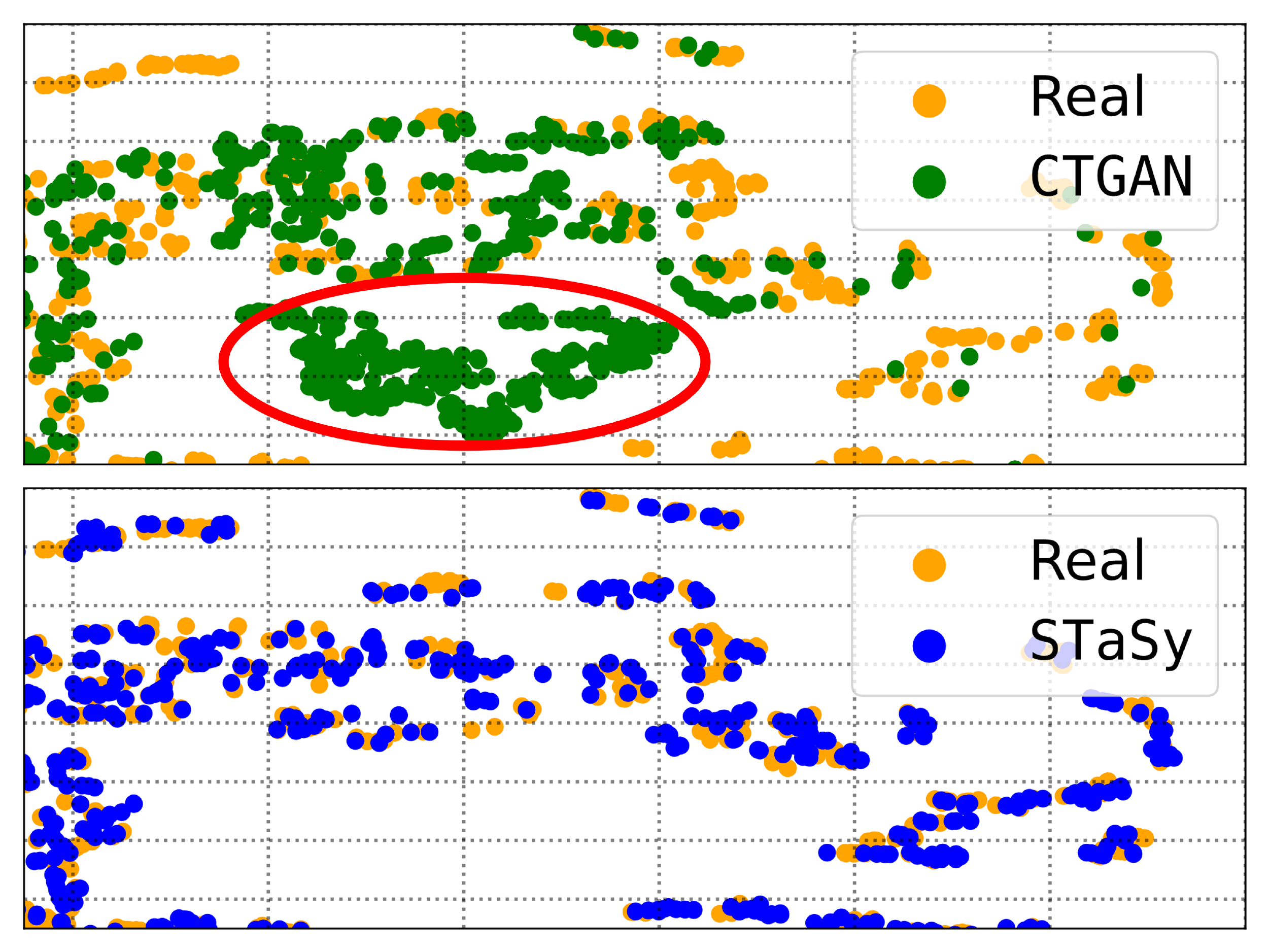}\label{fig:tsne}}
        \end{subfigure}
        % \vspace{-0.7em}
        \caption{(Left and Middle) Histograms of values in two columns of \texttt{Bean}. (Right) t-SNE~\citep{JMLR:v9:vandermaaten08a} visualizations of the fake and original records in \texttt{Obesity}. More visualizations are in Appendix~\ref{sec:additional_visualization}.}
        \label{fig:histogram}
        % \vspace{-0.5em}
        \vspace{-1em}
\end{figure}

In Figure~\ref{fig:histogram} (Left and Middle), the fake data by \texttt{STaSy} shows an almost identical distribution to that of real data. In contrast, \texttt{OCT-GAN}, which was proposed to address the multi-modality issue of tabular data, fails to do it. This means that \texttt{STaSy} is able to capture every mode in the columns, while \texttt{OCT-GAN} is not.
In Figure~\ref{fig:histogram} (Right), \texttt{CTGAN} generates some out-of-distribution records, highlighted in red.

% \jayoung{Our proposed strategies enhance the model in terms of the diversity as well.}

% \begin{wraptable}[10]{l}{0.24\textwidth}
% \vspace{-1.7em}
% \setlength\tabcolsep{1.8pt}
% \small
% \caption{\new{Runtime evaluation results}}
% \label{tab:runtime}
% \resizebox{0.24\textwidth}{!}{
%         \begin{tabular}{lcccc}
%         \toprule
%         Methods  & Runtime & \\ 
%         \midrule
%         \texttt{MedGAN} & 0.246 &  \\
%         \texttt{VEEGAN} & 0.109 &  \\
%         \texttt{CTGAN} & 0.704 &  \\ 
%         \texttt{TVAE} & \underline{0.100} &  \\
%         \texttt{TableGAN} & \textbf{0.046}  & \\ 
%         \texttt{OCT-GAN} & 26.926 &  \\ 
%         \texttt{RNODE} & 13.392 &  \\  \midrule
%         \texttt{Naive-STaSy} & 8.855  & \\ 
%         \texttt{STaSy} & 10.663   &\\ 
%         \bottomrule
%         \end{tabular}
% }
% \vspace{-1.2em}
% \end{wraptable}
\subsubsection{Sampling time}

\begin{table}[h]
    \centering
    \small
    \setlength\tabcolsep{2pt}
    \caption{Runtime evaluation results, averaged over all datasets}
    \label{tab:runtime}
    \begin{tabular}{lccccccccccc}
    % \begin{tabular}{llllllllllll}
    \toprule
    Methods  &  \texttt{MedGAN} &\texttt{VEEGAN}  &\texttt{CTGAN} &  \texttt{TVAE} & \texttt{TableGAN} & \texttt{OCT-GAN} & \texttt{RNODE} &  \texttt{Na\"ive-STaSy} &\texttt{STaSy} &  \\ 
    \midrule
    Runtime & 0.246 & 0.109 & 0.704 & \underline{0.100} & \textbf{0.046}&26.926 &  13.392 & 8.855  &  10.663  \\

    \bottomrule
    \end{tabular}
\end{table}

We summarize runtime in Table~\ref{tab:runtime}. In order to compare the runtime of all methods, we measure the wall-clock time taken to sample $N$ records, where $N$ is training size, 5 times, and average them. In general, simple GAN-based methods, especially \texttt{TableGAN} and \texttt{TVAE}, show faster runtime. On the other hand, SGMs, \texttt{OCT-GAN}, and \texttt{RNODE} take a relatively long time for sampling. Our proposed methods, \texttt{Na\"ive-STaSy} and \texttt{STaSy}, take a long sampling time compared to simple GAN-based methods but are faster than \texttt{OCT-GAN} and \texttt{RNODE}, which means a well-balanced trade-off between the sampling quality, diversity, and time.

\subsection{Ablation \& sensitivity studies}\label{sec:sen}

\begin{table}[h]
\centering
\setlength\tabcolsep{5pt}
\caption{Ablation study. We report F1 (resp. $R^2$) for classification (resp. regression).}
\label{tab:ablation}
\begin{tabular}{lcccccc}
\toprule
Datasets & \texttt{Na\"ive-STaSy} & w/o fine-tuning & w/o SPL & \texttt{STaSy}  \\ 
\midrule
 \texttt{Credit} &0.782\footnotesize{±0.042} & 0.782\footnotesize{±0.044} & 0.790\footnotesize{±0.028} & \textbf{0.794\footnotesize{±0.035}} & \\
\texttt{Default}  & 0.509\footnotesize{±0.014} & 0.517\footnotesize{±0.009} & 0.511\footnotesize{±0.016} & \textbf{0.523\footnotesize{±0.007}}  & \\
\texttt{Shoppers} & 0.635\footnotesize{±0.017} & 0.642\footnotesize{±0.009} & 0.635\footnotesize{±0.018} & \textbf{0.642\footnotesize{±0.015}}& \\
\midrule
\texttt{Contraceptive} & 0.418\footnotesize{±0.018} & 0.437\footnotesize{±0.012} & 0.434\footnotesize{±0.015} & \textbf{0.451\footnotesize{±0.017}} & \\
% & \texttt{Robot} & 0.965\footnotesize{±0.013} & 0.968\footnotesize{±0.002} & 0.969\footnotesize{±0.004} & \textbf{0.969\footnotesize{±0.003}} & \\
\texttt{Crowdsource} & 0.710\footnotesize{±0.104} & 0.738\footnotesize{±0.099} & 0.730\footnotesize{±0.108} & \textbf{0.743\footnotesize{±0.104}}  & \\
\texttt{Shuttle} & 0.791\footnotesize{±0.056} & 0.807\footnotesize{±0.088} & 0.816\footnotesize{±0.045} & \textbf{0.865\footnotesize{±0.072}} & \\
\midrule
 \texttt{Beijing} & 0.625\footnotesize{±0.141} & 0.670\footnotesize{±0.166} & 0.635\footnotesize{±0.145} & \textbf{0.672\footnotesize{±0.166}} & \\

\bottomrule
\end{tabular}
\end{table}

\begin{table}[h]
\centering
\caption{Sensitivity analyses. We report F1 (resp. $R^2$) for classification (resp. regression).}
\label{tab:sensitivity}
\begin{tabular}{llclclc}
\toprule
Datasets & \small{SDE Type} & Metric & $\alpha_0$ & Metric & $\beta_0$ & Metric   \\ 
\midrule
 \multirow{3}{*}{\texttt{Spambase}} & VE & 0.866\footnotesize{±0.036} & 0.05 & 0.890\footnotesize{±0.026} & 0.70 & 0.890\footnotesize{±0.023}\\ 
 &  VP & 0.875\footnotesize{±0.029} & 0.10 & 0.888\footnotesize{±0.024} & 0.80 & 0.893\footnotesize{±0.025} \\  
 & sub-VP & \textbf{0.893\footnotesize{±0.025}} & 0.30 & 0.888\footnotesize{±0.027} & 0.90 & 0.890\footnotesize{±0.028} \\  
\midrule
 \multirow{3}{*}{\texttt{Crowdsource}} & VE & \textbf{0.743\footnotesize{±0.104}} & 0.05 & 0.715\footnotesize{±0.103} & 0.75 & 0.712\footnotesize{±0.097}  \\ 
 & VP & 0.717\footnotesize{±0.102} & 0.10 & 0.713\footnotesize{±0.099} & 0.80 & 0.692\footnotesize{±0.127}   \\  
 & sub-VP & 0.667\footnotesize{±0.116} & 0.30 & 0.709\footnotesize{±0.105} & 0.95 & 0.700\footnotesize{±0.111} \\  
\midrule
 \multirow{3}{*}{\texttt{Beijing}} & VE& \textbf{0.672\footnotesize{±0.166}}& 0.05 & 0.663\footnotesize{±0.162} & 0.70 & 0.669\footnotesize{±0.165}  \\ 
 & VP & 0.594\footnotesize{±0.144} & 0.10 & 0.667\footnotesize{±0.164} & 0.75 & 0.668\footnotesize{±0.165}  \\  
 & sub-VP  & 0.525\footnotesize{±0.082} & 0.30 & 0.666\footnotesize{±0.164} & 0.95 & 0.668\footnotesize{±0.165} \\  
\bottomrule
\end{tabular}
% \vspace{-1em}
\end{table}

We define three ablation models: `\texttt{Na\"ive-STaSy}' without SPL and fine-tuning, `w/o fine-tuning' without fine-tuning but with SPL, and `w/o SPL' without SPL but with fine-tuning. All ablation models are inferior to \texttt{STaSy} in Table~\ref{tab:ablation}, showing the effectiveness of SPL and fine-tuning. In particular, SPL improves the sampling diversity as in Figure~\ref{fig:ablation}. \texttt{Na\"ive-STaSy} suffers from mild mode collapses, as highlighted in red.

Table~\ref{tab:sensitivity} shows sensitivity analyses w.r.t. some important hyperparameters. In general, all settings show reasonable results, outperforming the baselines. We recommend 0.2 and 0.25 for $\alpha_0$ and 0.9 and 0.95 for $\beta_0$.
% In some datasets, e.g., \texttt{Shoppers}, the model tends to show better performance as $\alpha_0$ gets larger, which means there exist more than 30\% of ``easy'' records.

We can adopt a variety of methods to solve the reverse SDE process in~\Eqref{eq:reverse}. Our method can generate fake records with the \emph{predictor-corrector} framework (Pred. Corr.) or the \emph{probability flow} (PF) method~\citep{songyang}. The former uses the ancestral sampling (AS), reverse diffusion (RD), or Euler-Maruyama (EM) method for solving the reverse SDE, and for the correction process, the Langevin corrector. In Table~\ref{tbl:sdesolver}, the \emph{probability flow} method in~\Eqref{eq:pfode} mostly leads to successful results, and other datasets also show similar results.

\begin{table}[h]
    \begin{minipage}{0.54\linewidth}
        % \vspace{-0.5em}
        \centering
        \includegraphics[width=0.65\textwidth]{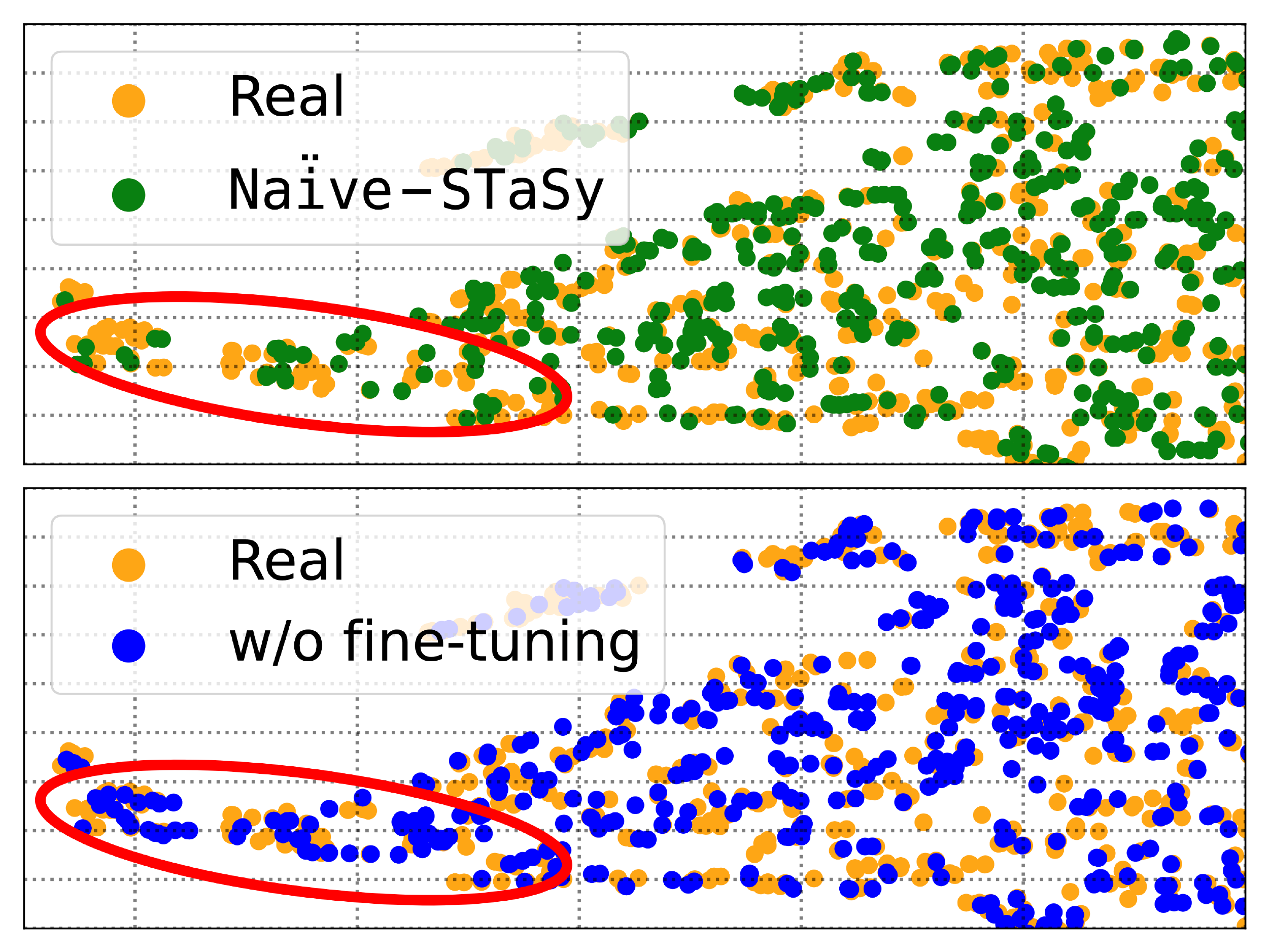}
        \captionof{figure}{t-SNE visualizations of the fake and original records in
        \texttt{Beijing}}\label{fig:ablation}
	\end{minipage}
    \hfill
	\begin{minipage}{0.42\linewidth}
        \small
        \centering
        \setlength{\tabcolsep}{2pt}
        \caption{Results of \texttt{Na\"ive-STaSy} by various reverse SDE solvers in \texttt{Magic}, where `Pred.' means the predictor-only method. We report F1.}
        \label{tbl:sdesolver}
        \begin{tabular}{lcc}
        \toprule
        Predictor & Pred.  & Pred. Corr. \\  %\cline{5-6} \cline{9-10}
        \midrule
        AS           &  0.752\footnotesize{±0.025} & 0.777\footnotesize{±0.048}\\ 
        RD            &  0.779\footnotesize{±0.052} & 0.775\footnotesize{±0.054}   \\  
        EM               &0.779\footnotesize{±0.052}& 0.771\footnotesize{±0.048}  \\  
        PF & \textbf{0.781±\footnotesize{0.054}} & No Corr. \\
        \bottomrule
        \end{tabular}
    \end{minipage}
\vspace{-1.5em}
\end{table}

\section{Conclusions and discussions}
Synthesizing tabular data is an important yet non-trivial task, as it requires modeling a joint probability of multi-modal columns. To this end, we presented our detailed designs and experimental results with thorough analyses. Our proposed method, \texttt{STaSy}, is a score-based model equipped with our proposed self-paced learning and fine-tuning methods. In our experiments with 15 benchmark datasets and 7 baselines, \texttt{STaSy} outperforms other deep learning methods in terms of the sampling quality and diversity (and with an acceptable sampling time). Based on these considerations, we believe that \texttt{STaSy} shows significant advancements in tabular data synthesis. We expect much follow-up work in utilizing SGMs for tabular data synthesis.

\noindent\textbf{Limitations.}\label{sec:limitation} Although our model shows the best balance for the deep generative task trilemma, we think that there exists room to improve runtime further --- existing simple GAN-based methods are faster than our method for sampling fake records. In addition, SGMs are known to be sometimes unstable for high-dimensional data, e.g., high-resolution images, but in general, stable for low-dimensional data, e.g., tabular data. Therefore, we think that SGMs have much potential for tabular data synthesis in the future. 

\clearpage

\section{Ethics statement}\label{sec:ethics}
Indeed, people do not always use artificial intelligence technology for righteous purposes. One can use our method to achieve his/her wrongful goals, e.g., selling high-quality fake data generated by our method, and retrieving private original data records from synthetic data. However, we believe that our research has much more beneficial points. One can use our method to generate fake data and share (after hiding the original data) to prevent potential privacy leakages. We, of course, need more studies to achieve the privacy protection goal based on our model. However, a research trend exists where researchers try to use a deep generative model to protect privacy~\citep{park2018data, lee2021invertible}.

\section{Reproducibility statement}\label{sec:reproducibility}
To reproduce the experimental results, we have made the following efforts: 1) Source codes used in the experiments are available in the supplementary material. By following the README guidance, the main results are easily reproducible. 2) All the experiments are repeated five times, and their mean and standard deviation values are reported in Appendix. 3) We provide extensive experimental details in Appendix~\ref{sec:apd_environments}.

% \textcolor{red}{Authors are strongly encouraged to include a paragraph-long Reproducibility Statement at the end of the main text (before references) to discuss the efforts that have been made to ensure reproducibility. This paragraph should not itself describe details needed for reproducing the results, but rather reference the parts of the main paper, appendix, and supplemental materials that will help with reproducibility. For example, for novel models or algorithms, a link to a anonymous downloadable source code can be submitted as supplementary materials; for theoretical results, clear explanations of any assumptions and a complete proof of the claims can be included in the appendix; for any datasets used in the experiments, a complete description of the data processing steps can be provided in the supplementary materials. Each of the above are examples of things that can be referenced in the reproducibility statement. This optional reproducibility statement will not count toward the page limit, but should not be more than 1 page.}

\bibliography{iclr2023_conference}
\bibliographystyle{iclr2023_conference}
\clearpage
\appendix

\section{VE, VP, and sub-VP SDEs}\label{fandg}

We introduce the definitions of $f$ and $g$ as follows:
% \begin{align}
%     f(t)&=\begin{cases}0,  &\textrm{ if VE,}\\
%     -\frac{1}{2}\beta(t)\mathbf{x},&\textrm{ if VP,}\\
%     -\frac{1}{2}\beta(t)\mathbf{x},&\textrm{ if sub-VP,}\\
%     \end{cases}\label{eq:drift}\\
%     g(t)&=\begin{cases}\sqrt{\frac{\texttt{d}[\sigma^2(t)]}{\texttt{d}t}},&\textrm{ if VE,}\\
%     \sqrt{\beta(t)},&\textrm{ if VP,}\\
%     \sqrt{\beta(t)(1-e^{-2\int_0^t \beta(s)\, \texttt{d}s})},&\textrm{ if sub-VP,}\\
%     \end{cases}\label{eq:diffusion}
% \end{align}where $\sigma(t)$ and $\beta(t)$ are noise functions w.r.t. time $t$. 
% $\sigma(t) = \sigma_{min} \left( \frac{\sigma_{max}}{\sigma_{min}} \right)^t$ for $t \in [0, 1]$, where $\sigma_{min}$ and $\sigma_{max}$ are hyperparameters, and we use $0.01$ and $10.0$, respectively, by default.

% $\beta(t) = \beta_{min} + t\left( \beta_{max}-\beta_{min}\right)$ for $t \in [0, 1]$, where $\beta_{min}$ and $\beta_{max}$ are hyperparameters, and we use $0.01$ and $10.0$, respectively, by default.

\begin{align}
    f(t)&=\begin{cases}0,  &\textrm{ if VE,}\\
    -\frac{1}{2}\gamma(t)\mathbf{x},&\textrm{ if VP,}\\
    -\frac{1}{2}\gamma(t)\mathbf{x},&\textrm{ if sub-VP,}\\
    \end{cases}\label{eq:drift}\\
    g(t)&=\begin{cases}\sqrt{\frac{\texttt{d}[\sigma^2(t)]}{\texttt{d}t}},&\textrm{ if VE,}\\
    \sqrt{\gamma(t)},&\textrm{ if VP,}\\
    \sqrt{\gamma(t)(1-e^{-2\int_0^t \gamma(s)\, \texttt{d}s})},&\textrm{ if sub-VP,}\\
    \end{cases}\label{eq:diffusion}
\end{align}where $\sigma(t)$ and $\gamma(t)$ are noise functions w.r.t. time $t$. 
$\sigma(t) = \sigma_{min} \left( \frac{\sigma_{max}}{\sigma_{min}} \right)^t$ for $t \in [0, 1]$, where $\sigma_{min}$ and $\sigma_{max}$ are hyperparameters, and we use $\sigma_{min}=\{0.01, 0.1\}$ and $\sigma_{max}=\{5.0, 10.0\}$. $\gamma(t) = \gamma_{min} + t\left( \gamma_{max}-\gamma_{min}\right)$ for $t \in [0, 1]$, where $\gamma_{min}$ and $\gamma_{max}$ are hyperparameters, and we use $\gamma_{min}=\{0.01, 0.1\}$ and $\gamma_{max}=\{5.0, 10.0\}$.

\section{Comparison between \texttt{SOS} and \texttt{STaSy} for the oversampling task} \label{sec:oversampling}

\begin{table}[h]
\caption{Comparison between \texttt{SOS} and \texttt{STaSy} in terms of Weighted-F1}
\centering
\label{tbl:oversampling}
\begin{tabular}{lcccccccccccc}
\toprule
% \multirow{2}{*}{Methods} && \multicolumn{2}{c}{\texttt{HTRU}} && \multicolumn{2}{c}{\texttt{Robot}} & \\ \cmidrule{3-4} \cmidrule{6-7}
Methods && \texttt{HTRU} && \texttt{Magic} & & \texttt{Robot} & \\ 
 \midrule
\texttt{Identity}  && {0.8657}\footnotesize{±0.0163}     && {0.7807}\footnotesize{±0.0348}     && {0.9031}\footnotesize{±0.0848}     & \\  \midrule
\texttt{SOS}  && {0.8767}\footnotesize{±0.0017}     && \small{0.7949}\footnotesize{±0.0011}  && \small{0.9197}\footnotesize{±0.0040}  &\\
\texttt{STaSy} && \textbf{{0.8803}\footnotesize{±0.0027}}       && \textbf{{0.7960}\footnotesize{±0.0015}}  && \textbf{{0.9270}\footnotesize{±0.0029}}  & \\ 
\bottomrule
\end{tabular}
% \vspace{-1em}
\end{table}

% \begin{table}[h]
% \caption{Comparison between \texttt{SOS} and \texttt{STaSy} in terms of Weighted-F1}
% \centering
% \label{tbl:oversampling}
% \begin{tabular}{lcccccccccccc}
% \toprule
% % \multirow{2}{*}{Methods} && \multicolumn{2}{c}{\texttt{HTRU}} && \multicolumn{2}{c}{\texttt{Robot}} & \\ \cmidrule{3-4} \cmidrule{6-7}
% Methods && \texttt{HTRU} && \texttt{Magic} & & \texttt{Shoppers} & & \texttt{Default} &\\ 
%  \midrule
% \texttt{Identity}  && {0.8657}\footnotesize{±0.0163}     && {0.7807}\footnotesize{±0.0348}     && {0.5654}\footnotesize{±0.0720}     & & {0.5123}\footnotesize{±0.0271}     &\\  \midrule
% \texttt{SOS}  && {0.8767}\footnotesize{±0.0017}     && \small{0.7949}\footnotesize{±0.0011}  && \small{0.6565}\footnotesize{±0.0045}  && \small{0.}\footnotesize{±0.}  &\\
% \texttt{STaSy} && \textbf{{0.8803}\footnotesize{±0.0027}}       && \textbf{{0.7960}\footnotesize{±0.0015}}  && \textbf{{0.6582}\footnotesize{±0.0038}}  && \textbf{{0.}\footnotesize{±0.}}  & \\ 
% \bottomrule
% \end{tabular}
% % \vspace{-1em}
% \end{table}

In this section, we discuss the difference between \texttt{SOS} and \texttt{STaSy}. They are both based on SGMs, but they are optimized towards different goals by using different objective functions and training strategies. \texttt{SOS} has many design points specialized to augment minor classes only (rather than synthesizing entire tabular data) --- for instance, \texttt{SOS} adopts a style-transfer-based idea to convert a major class sample to a minor one via their own SGM model without any consideration on the training difficulty of the denoising score matching. However, our \texttt{STaSy} more focuses on synthesizing entire tabular data by proposing special self-paced training and fine-tuning methods.

% Starting from \texttt{Na\"ive-STaSy}, a base model for \texttt{SOS} and a naive conversion of SGMs for images toward tabular data, we modify the objective to relieve the instability of training models on denoising score matching by applying the self-paced learning. 

Since \texttt{STaSy} can be converted to an oversampling method following the design guidance of \texttt{SOS}, we conduct oversampling experiments with \texttt{STaSy} to compare with \texttt{SOS}~\citep{sos}. We train, for fair comparison, \texttt{STaSy} w/o fine-tuning for each minor class and generate minority samples to be the same size of the majority class. We compare the two models in terms of the sampling quality using Weighted-F1 which is specialized in evaluating imbalanced data. We note that \texttt{Identity} means that we do not use any oversampling methods, which is, therefore, a minimum quality requirement upon oversampling.

As shown in Table~\ref{tbl:oversampling}, \texttt{STaSy} w/o fine-tuning outperforms \texttt{SOS}. The result shows that our proposed training strategy, i.e., the self-paced learning, improves the model training regardless of tasks.

\section{Network architecture}\label{networkarchi}
We propose the following score network $S_{\boldsymbol{\theta}}(\mathbf{x}(t), t)$:
% \begin{align*}
%     \mathbf{h}_0 &= \mathbf{x}(t),\\
%     \mathbf{h}_1 &=  \mathtt{H}_1(\mathbf{x}(t), \mathbf{e}_t),\\
%     \mathbf{h}_i &=  \mathbf{h}_{i-1} \oplus \mathtt{H}_i(\mathbf{h}_{i-1}, \mathbf{e}_t), 1 < i < d_N\\
%     % \mathbf{h}_2 &= \mathtt{H}_2(\mathbf{h}_1, \mathbf{e}_t),\\
%      \vdots \\
%     \mathbf{h}_{d_N} &= \mathbf{h}_{d_{N}-1} \oplus \mathtt{H}_{d_N}(\mathbf{h}_{d_{N}-1}, \mathbf{e}_t), \\
%     S_{\boldsymbol{\theta}}(\mathbf{x}(t), t) &= \mathtt{FC}(\mathbf{h}_{d_N}),
% \end{align*}
\begin{align*}
    \mathbf{h}_0     &=  \mathbf{x}(t),\\
    % \mathbf{h}_1     &=  \omega(\mathtt{H}_1(\mathbf{h}_0, t)),\\
    \mathbf{h}_i     &=  \omega( \mathtt{H}_i(\mathbf{h}_{i-1}, t) \oplus \mathbf{h}_{i-1} ), 1 \le i \le d_N\\
    %  \vdots \\
    % \mathbf{h}_{d_N} &=  \mathbf{h}_{d_{N}-1} \oplus \mathtt{H}_{d_N}(\mathbf{h}_{d_{N}-1}, t), \\
    S_{\boldsymbol{\theta}}(\mathbf{x}(t), t) &= \texttt{FC}(\mathbf{h}_{d_N}),
\end{align*}where $\mathbf{x}(t)$ is a record (or a row) at time $t$ in tabular data, $\mathbf{h}_i$ is the $i$-th hidden vector, and $\omega$ is an activation function. $d_N$ is the number of hidden layers. For various layer types of $\mathtt{H}_i(\mathbf{h}_{i-1}, t)$, we provide the following options:
% \begin{align*}
%     \mathtt{H}_i(\mathbf{h}_{i-1}, \mathbf{e}_t) &= 
%         \begin{cases}
%             \omega(\texttt{FC}_{i}(\mathbf{h}_{i-1}) \odot \psi(\texttt{FC}^{t}_{i}(\mathbf{e}_t))), &\textrm{ if Squash,}\\
%             \omega(\texttt{FC}_{i}(\mathbf{e}_t \oplus \mathbf{h}_{i-1})), &\textrm{ if Concat,} \\
%             \omega(\texttt{FC}_{i}(\mathbf{h}_{i-1}) \odot \psi(\texttt{FC}^{gate}_{i}(\mathbf{e}_t)) + \texttt{FC}^{bias}_{i}(\mathbf{e}_t)), &\textrm{ if Concatsquash,}
%         \end{cases}
% \end{align*}
\begin{align*}
    \mathtt{H}_i(\mathbf{h}_{i-1}, t) &= 
        \begin{cases}
            \texttt{FC}_{i}(\mathbf{h}_{i-1}) \odot \psi(\texttt{FC}^{t}_{i}(t)), &\textrm{ if Squash,}\\
            \texttt{FC}_{i}(t \oplus \mathbf{h}_{i-1}), &\textrm{ if Concat,} \\
            \texttt{FC}_{i}(\mathbf{h}_{i-1}) \odot \psi(\texttt{FC}^{gate}_{i}(t) + \texttt{FC}^{bias}_{i}(t)), &\textrm{ if Concatsquash,}
        \end{cases}
\end{align*}where we can choose one of the three possible layer types as a hyperparameter, $\odot$ means the element-wise multiplication, $\oplus$ means the concatenation operator, $\psi$ is the Sigmoid function, and $\texttt{FC}$ is a fully connected layer. We modify the architecture of~\citep{songyang} by using the layer types, being inspired by~\citep{grathwohl2018ffjord}.

% \begin{align*}
%     \bm{f}_{j,i} &= \mathtt{ReLU}(\mathtt{BN}(\mathtt{FC1}(\bm{z}_j))), \\
%     \texttt{G}_{j,i}(\bm{z}) &= \mathtt{FC3}(\bm{h}(0)), 
% \end{align*} where $\mathtt{FC}$ means a fully-connected layer, $\mathtt{BN}$ is a batch normalization, and $\mathtt{ReLU}$ is a rectified linear unit.
% \nspark{need to clear the meaning of thes subscripts.}

\section{Threshold controlling mechanism}\label{sec:threshold}
% $\alpha = \alpha_0 + \log (\frac{e}{\sfrac{c}{S}})(1-\alpha_0)$ and 
% $\beta = \beta_0 + \log (\frac{e}{\sfrac{c}{S}})(1-\beta_0)$. 

% The threshold controlling variable $\alpha$ and $\beta$, where $\alpha \le \beta \le 1$, are the probabilities that loss $l_i$ will take a value less than or equal to $Q(\alpha)$ and $Q(\beta)$, respectively. 

Our threshold controlling mechanism is designed to meet the following 3 requirements: i) it can control when the entire dataset is used for training, starting from a subset, ii) it should gradually increase the size of used training records while logarithmically decreasing the number of hard records (that are not involved in training), and iii) it should be a monotonically increasing/decreasing function to guarantee that the training difficulty gets more challenging as the training process goes on.

The threshold controlling variables $\alpha$ and $\beta$, where $0 \le \alpha \le \beta \le 1$, are gradually increased to 1 to involve the entire data records for training. We increase them proportionally to training steps, where $\alpha = \alpha_0 + \log \Big(1+ c \Big(\frac{e-1}{S} \Big)(1-\alpha_0)\Big)$ and $\beta = \beta_0 + \log \Big(1+ c \Big(\frac{e-1}{S} \Big) (1-\beta_0) \Big)$. $e$ is the base of the natural logarithm, $\alpha_0$ and $\beta_0$ are initial values of $\alpha$ and $\beta$, $c$ is the current training step, and $S$ determines when to utilize the entire data records. We use 10,000 for $S$.
% the number of total steps for the model training.
We set $\beta_0$ at least 0.8 to ensure 80\% of the data records are involved in training at the start of the training. 

\section{Proof of Theorem~\ref{theorem:v}}\label{sec:derivation}
% In this section, we show that optimal solution for $\mathbf{v}^*$, 
As defined in Section~\ref{sec:selfpacedlearning}, the \texttt{STaSy} objective is as follows: 

\begin{align}
    \underset{\boldsymbol{\theta}, \mathbf{v}}{min} \sum_{i=1}^{N} v_i l_i -\frac{Q(\alpha) - Q(\beta)}{2} \sum_{i=1}^N v_i^2 - Q(\beta) \sum_{i=1}^N v_i,
\end{align}where $l_i$ is the score matching loss for $i$-th training record as in~\Eqref{eq:L_i}. We can rewrite the optimal solution for each training record $v_i$ with respect to fixed $\boldsymbol{\theta}$ in the vertex form. Let $\mathcal{L}(v_i)$ be the objective with fixed $\boldsymbol{\theta}$, which is a quadratic function with respect to $v_i$. Then,
\begin{align}
\begin{split}
\mathcal{L}(v_i) &=  v_i l_i - \frac{Q(\alpha) - Q(\beta)}{2}  v_i^2 - Q(\beta) v_i \\
     &= - \frac{Q(\alpha) - Q(\beta)}{2} v_i^2 + (l_i - Q(\beta)) v_i  \\
     &= - \frac{Q(\alpha) - Q(\beta)}{2} \Big( v_i^2 - \frac{2( l_i - Q(\beta))}{Q(\alpha) - Q(\beta)} v_i \Big) \\
     &= - \frac{Q(\alpha) - Q(\beta)}{2} \Big\{ v_i^2 - \frac{2( l_i - Q(\beta))}{Q(\alpha) - Q(\beta)} v_i + \Big( \frac{ l_i -  Q(\beta) }{Q(\alpha)-Q(\beta) }\Big )^2 -  \Big( 
     \frac{ l_i -  Q(\beta)}{ Q(\alpha)-Q(\beta) }\Big )^2 \Big\} \\
     &= - \frac{Q(\alpha) - Q(\beta)}{2} \Big( v_i - \frac{  l_i - Q(\beta)}{ Q(\alpha) - Q(\beta)} \Big)^2 + \frac{ Q(\alpha) - Q(\beta)}{2} \Big( \frac{ l_i - Q(\beta)}{ Q(\alpha) - Q(\beta)} \Big)^2.  \label{18}
\end{split}
\end{align}

Because $\frac{Q(\alpha) - Q(\beta)}{2} $ is less than or equal to 0 and $\frac{ Q(\alpha) - Q(\beta)}{2} \Big( \frac{ l_i - Q(\beta)}{ Q(\alpha) - Q(\beta)} \Big)^2$ is a constant, the solution $v_i $ which minimizes~\Eqref{18} is $v_i = \frac{ l_i - Q(\beta)}{ Q(\alpha) - Q(\beta)}$. Considering $v_i \in [0, 1]$, we can get the optimal ${v_i}$ as follows:
\begin{align}
    v_{i}^* = \begin{dcases*}
        1,                                                            & if $l_i \le Q(\alpha)$,\\
        0,                                                            & if $l_i \ge Q(\beta)$,\\
        \frac{l_i - Q(\beta)}{ Q(\alpha)- Q(\beta)} ,  & otherwise.
        \end{dcases*}
\end{align}

% We use the denoising score199
% matching loss instead of the Hutchinson’s log-probability estimation at Line 11, since it provides200
% efficiency in enhancing log-probabilities as well as better sampling quality (see Appendix E

\section{The Hutchinson's estimation as a fine-tuning objective}\label{sec:exactlikelihood}
% We use the denoising score matching loss instead of the Hutchinson's log-probability estimation in  Line~\ref{a:fine} of Algorithm~\ref{alg1}. In this section, we describe the results of an additional experiment in which the Hutchinson's log-probability estimation is used for the fine-tuning objective. 
We use the denoising score matching loss in Line~\ref{a:fine} of Algorithm~\ref{alg1}. In this section, we describe the results of an additional experiment in which the Hutchinson's log-probability estimation is used for the fine-tuning objective.

Table~\ref{tab:hutchinson} summarizes the F1 score and the median of log-probabilities when we use the denoising score matching loss and the Hutchinson's estimation as the tine-tuning objective. In \texttt{Shoppers} and \texttt{Crowdsource}, there does not exist a clear winner between the two fine-tuning objectives, and similar results are also shown in other datasets. However, in some datasets, e.g., \texttt{Default} and \texttt{Contraceptive}, the former shows better F1 scores and better medians of the log-probabilities than the latter by large margins. In addition, when we update $\boldsymbol{\theta}$ using the Hutchinson's estimation, in \texttt{Default}, the sampling quality is lower than before fine-tuning. Considering these results, we use the denoising score matching loss, which shows the generalizability, as our default fine-tuning objective.

\begin{table}[h]
\caption{We report the F1 score and the median of the log-probabilities of testing records according to fine-tuning objective.}
\centering
\setlength\tabcolsep{2pt}
\resizebox{1\textwidth}{!}{
\label{tab:hutchinson}
\begin{tabular}{lccccccccc}
\toprule
\multirow{3}{*}{Datasets}  &  \multicolumn{2}{c}{\texttt{STaSy}}    & &  \multicolumn{2}{c}{Fine-tine with}  & & \multicolumn{2}{c}{Fine-tune with}  & \\ 
&  \multicolumn{2}{c}{w/o fine-tuning} & &  \multicolumn{2}{c}{the denoising score matching loss} & & \multicolumn{2}{c}{the Hutchinson's estimation} & \\ \cmidrule{2-3}\cmidrule{5-6} \cmidrule{8-9}
&  F1 &Log-probability & & F1 & Log-probability & &  F1 &Log-probability & \\ 
\midrule
{\texttt{Default}} & 0.517\footnotesize{±0.009} &131.011
&& \textbf{0.523\footnotesize{±0.007}} & \textbf{131.122}
&& 0.510\footnotesize{±0.012} & 106.184
\\ 
{\texttt{Shoppers}} & 0.642\footnotesize{±0.009} & 189.422 
&&0.642\footnotesize{±0.015} & \textbf{220.589} 
&&\textbf{0.643\footnotesize{±0.005}} & 206.924
\\ 
{\texttt{Contraceptive}} & 0.437\footnotesize{±0.012} & 225.334 
&& \textbf{0.451\footnotesize{±0.017}} & \textbf{225.638} 
&& 0.438\footnotesize{±0.012} & 225.392

\\ 
{\texttt{Crowdsource}} & 0.738\footnotesize{±0.099} & 47.567 
&& 0.743\footnotesize{±0.104} & \textbf{48.905} 
&& \textbf{0.744\footnotesize{±0.086}} & 48.683
\\ 
\bottomrule
% \label{tab:statistical1}
\end{tabular}
}
\end{table}

\section{Additional experimental results}\label{sec:additional_scores}

% \texttt{MedGAN} fails to generate all classes in multi-class classification datasets marked as `N/A', i.e., mode collapse happens.

% We also report statistical test results comparing the corresponding columns of real and fake data. We use KS test and CS test for numerical columns and for categorical columns, respectively. The statistical test shows higher scores when the distribution of real and fake data is similar. The test is conducted for each column, and the result in Table~\ref{tab:statistic} is the average across all columns. Our method, \texttt{STaSy}, outperforms all the baselines in 5 out of 8 datasets, and \texttt{CTGAN}, \texttt{TVAE}, and \texttt{Na\"ive-STaSy} also show reliable results.

% In Table~\ref{tab:additional}, we apply \texttt{Na\"ive-STaSy} and \texttt{STaSy} w/o fine-tuning to the 9 additional real-world datasets (that were not used in our main paper) to show the generalizability of our proposed method~\citep{shwartz2022tabular}. We compare our model to the top 3 baselines, i.e., \texttt{TVAE}, \texttt{OCT-GAN}, and \texttt{RNODE}. The top 3 models are selected with the average F1 score from Table~\ref{tbl:main}. Considering the results, \texttt{STaSy} generalizes well to other datasets.

\subsection{Sampling quality}\label{sec:additional_quality}
We mainly use F1 (resp. $R^2$) for the classfication (resp. regression) TSTR evaluation, and also report AUROC and Weighted-F1 (resp. RMSE) results. Full results for all datasets are in Tables~\ref{tab:binary_f1}, ~\ref{tab:AUROC_binary}, and ~\ref{tab:weighted_f1_binary} for binary classification, Tables~\ref{tab:macro_f1}, ~\ref{tab:AUROC_multi}, and ~\ref{tab:weighted_f1_multi} for multi-class classification, and Table~\ref{tab:regression} for regression. We train and test various base classifiers/regressors and report their mean and standard deviation. Moreover, we use the log-probability as another metric for the sampling quality. Full results are in Table~\ref{tab:probability}. The best results are highlighted in bold face and the second best results with underline. As shown, \texttt{Na\"ive-STaSy} and \texttt{STaSy} consistently show the best and the second best performances.

%  \jayoung{In particular, for \texttt{Credit}, which has a severe class imbalance problem, i.e., 99.8\% for class 0 and 0.2\% for class 1, almost all baselines fail to sample high-quality fake records, except \texttt{CTGAN} and \texttt{TableGAN}.}

\begin{table}[h]
\caption{Classification with real data. We report F1 for binary classification.}
\centering
\label{tab:binary_f1}
% \small
% \scriptsize
\setlength\tabcolsep{2.5pt}
% \label{tbl:main}
\begin{tabular*}{1\textwidth}{lcccccccccc}
\toprule
\multirow{2}{*}{Methods} &  & \multicolumn{7}{c}{{Binary classification}}   &    \\ \cmidrule{3-9}
&  & \small{\texttt{Credit}}& \small{\texttt{Default}}       &   \small{\texttt{HTRU}}  &  \small{\texttt{Magic}}  &  \small{\texttt{Phishing}}   &  \small{\texttt{Shoppers}} & \small{\texttt{Spambase}} &  \\ 
\midrule
\texttt{\small{Identity}}        & & \small{0.775}\scriptsize{±0.071} & \small{0.452}\scriptsize{±0.044} & \small{0.884}\scriptsize{±0.010} & \small{0.782}\scriptsize{±0.059} & \small{0.950}\scriptsize{±0.023} & \small{0.611}\scriptsize{±0.063} & \small{0.940}\scriptsize{±0.023}  & \\
\midrule
\texttt{\small{MedGAN}}       & & \small{0.000}\scriptsize{±0.000} & \small{0.000}\scriptsize{±0.000} & \small{0.033}\scriptsize{±0.066} & \small{0.566}\scriptsize{±0.069} & \small{0.615}\scriptsize{±0.001} & \small{0.285}\scriptsize{±0.161} & \small{0.489}\scriptsize{±0.218}  &  \\ 
\texttt{\small{VEEGAN}}        & & \small{0.002}\scriptsize{±0.005} & \small{0.368}\scriptsize{±0.000} & \small{0.487}\scriptsize{±0.265} & \small{0.571}\scriptsize{±0.049} & \small{0.815}\scriptsize{±0.061} & \small{0.398}\scriptsize{±0.064} & \small{0.586}\scriptsize{±0.045}  &
    \\ 
\texttt{\small{CTGAN}}        & &  \small{0.757}\scriptsize{±0.064} & \small{0.497}\scriptsize{±0.012} & \small{0.853}\scriptsize{±0.007} & \small{0.737}\scriptsize{±0.035} & \small{0.898}\scriptsize{±0.009} & \small{0.510}\scriptsize{±0.053} & \small{0.785}\scriptsize{±0.019}  &
   \\
\texttt{\small{TVAE}}        & & \small{0.000}\scriptsize{±0.000} & \small{0.444}\scriptsize{±0.029} & \small{0.854}\scriptsize{±0.005} & \small{0.701}\scriptsize{±0.010} & \small{0.913}\scriptsize{±0.007} & \small{0.585}\scriptsize{±0.039} & \small{0.758}\scriptsize{±0.033}  &
   \\ 
\texttt{\small{TableGAN}}       & &  \small{0.739}\scriptsize{±0.038} & \small{0.431}\scriptsize{±0.006} & \small{0.843}\scriptsize{±0.005} & \small{0.740}\scriptsize{±0.045} & \small{0.903}\scriptsize{±0.009} & \small{0.603}\scriptsize{±0.061} & \small{0.741}\scriptsize{±0.104}  & 
    \\ 
\texttt{\small{OCT-GAN}}        & & \small{0.127}\scriptsize{±0.187} & \small{0.486}\scriptsize{±0.018} & \small{0.865}\scriptsize{±0.010} & \small{0.728}\scriptsize{±0.018} & \small{0.905}\scriptsize{±0.007} & \small{0.622}\scriptsize{±0.038} & \small{0.859}\scriptsize{±0.022}  & 
\\
\texttt{\small{RNODE}}       & &  \small{0.117}\scriptsize{±0.110} & \small{0.407}\scriptsize{±0.014} & \small{0.632}\scriptsize{±0.047} & \small{0.745}\scriptsize{±0.035} & \small{0.899}\scriptsize{±0.003} & \small{0.541}\scriptsize{±0.024} & \small{0.825}\scriptsize{±0.049}  & 
   \\ 
\midrule
\texttt{\small{Na\"ive-STaSy}}       & & \underline{\small{0.782}\scriptsize{±0.042}} & \underline{\small{0.509}\scriptsize{±0.014}} & \underline{\small{0.885}\scriptsize{±0.009}} & \underline{\small{0.780}\scriptsize{±0.054}} & \underline{\small{0.925}\scriptsize{±0.010}} & \underline{\small{0.635}\scriptsize{±0.017}} & \underline{\small{0.884}\scriptsize{±0.026}}  &
     \\ 
\texttt{\small{STaSy}}  & &  \textbf{\small{0.794}\scriptsize{±0.035}} & \textbf{\small{0.523}\scriptsize{±0.007}} & \textbf{\small{0.885}\scriptsize{±0.007}} & \textbf{\small{0.781}\scriptsize{±0.054}} & \textbf{\small{0.932}\scriptsize{±0.013}} & \textbf{\small{0.642}\scriptsize{±0.015}} & \textbf{\small{0.893}\scriptsize{±0.025}}  &
   \\ 
\bottomrule
\end{tabular*}
\end{table}

\begin{table}[h]
\caption{Classification with real data. We report AUROC for binary classification.}
\centering
\small
\setlength\tabcolsep{2.5pt}
\label{tab:AUROC_binary}
\begin{tabular*}{1\textwidth}{lcccccccccc}
\toprule
\multirow{2}{*}{\small{Methods}}  &  & \multicolumn{7}{c}{\small{Binary classification}}   &    \\ \cmidrule{3-9}
&  &  \texttt{\small{Credit}} & \texttt{\small{Default}}       &   \texttt{\small{HTRU}}  &  \texttt{\small{Magic}}  &  \small{\texttt{Phishing}}   &  \small{\texttt{Shoppers}} & \small{\texttt{Spambase}} &  \\ 
\midrule
\texttt{\small{Identity}}        & & \small{0.955}\scriptsize{±0.052} & \small{0.769}\scriptsize{±0.015} & \small{0.971}\scriptsize{±0.004} & \small{0.909}\scriptsize{±0.039} & \small{0.991}\scriptsize{±0.007} & \small{0.922}\scriptsize{±0.015} & \small{0.986}\scriptsize{±0.014}  &
\\ 
\midrule
\texttt{\small{MedGAN}}       & & \small{0.500}\scriptsize{±0.000} & \small{0.500}\scriptsize{±0.000} & \small{0.577}\scriptsize{±0.179} & \small{0.704}\scriptsize{±0.061} & \small{0.677}\scriptsize{±0.149} & \small{0.752}\scriptsize{±0.089} & \small{0.825}\scriptsize{±0.059}  &
    \\ 
\texttt{\small{VEEGAN}}        & & \small{0.661}\scriptsize{±0.115} & \small{0.516}\scriptsize{±0.055} & \small{0.876}\scriptsize{±0.114} & \small{0.789}\scriptsize{±0.042} & \small{0.899}\scriptsize{±0.065} & \small{0.798}\scriptsize{±0.074} & \small{0.733}\scriptsize{±0.064}  & 
 \\ 
\texttt{\small{CTGAN}}       & & \small{0.957}\scriptsize{±0.034} & \small{0.749}\scriptsize{±0.007} & \small{0.954}\scriptsize{±0.015} & \small{0.871}\scriptsize{±0.027} & \small{0.969}\scriptsize{±0.007} & \small{0.849}\scriptsize{±0.029} & \small{0.908}\scriptsize{±0.026}  &
  \\ 
\texttt{\small{TVAE}}        & & \small{0.500}\scriptsize{±0.000} & \small{0.743}\scriptsize{±0.012} & \small{0.955}\scriptsize{±0.012} & \small{0.840}\scriptsize{±0.028} & \small{0.980}\scriptsize{±0.005} & \small{0.861}\scriptsize{±0.022} & \small{0.906}\scriptsize{±0.036}  & 
    \\ 
\texttt{\small{TableGAN}}       & & \small{0.941}\scriptsize{±0.031} & \small{0.662}\scriptsize{±0.026} & \small{0.959}\scriptsize{±0.011} & \small{0.888}\scriptsize{±0.028} & \small{0.972}\scriptsize{±0.007} & \small{0.859}\scriptsize{±0.037} & \small{0.930}\scriptsize{±0.046}  &
   \\ 
\texttt{\small{OCT-GAN}}        & & \small{0.813}\scriptsize{±0.106} & \small{0.726}\scriptsize{±0.011} & \small{0.959}\scriptsize{±0.012} & \small{0.868}\scriptsize{±0.017} & \small{0.973}\scriptsize{±0.006} & \small{0.890}\scriptsize{±0.016} & \small{0.950}\scriptsize{±0.020}  &
 \\ 
\texttt{\small{RNODE}}       & & \small{0.877}\scriptsize{±0.129} & \small{0.723}\scriptsize{±0.009} & \small{0.940}\scriptsize{±0.029} & \small{0.874}\scriptsize{±0.033} & \small{0.966}\scriptsize{±0.008} & \small{0.889}\scriptsize{±0.020} & \small{0.933}\scriptsize{±0.043}  & 
   \\ 
\midrule
\texttt{\small{Na\"ive-STaSy}}       & & \textbf{\small{0.972}\scriptsize{±0.020}} & \textbf{\small{0.749}\scriptsize{±0.019}} & \textbf{\small{0.969}\scriptsize{±0.009}} & \underline{\small{0.905}\scriptsize{±0.036}} & \underline{\small{0.984}\scriptsize{±0.005}}& \underline{\small{0.910}\scriptsize{±0.009}} & \underline{\small{0.958}\scriptsize{±0.025}}&
    \\ 
\texttt{\small{STaSy}}  & &  \underline{\small{0.967}\scriptsize{±0.036}} & \underline{\small{0.747}\scriptsize{±0.018}} & \underline{\small{0.966}\scriptsize{±0.010}} & \textbf{\small{0.905}\scriptsize{±0.037}} & \textbf{\small{0.986}\scriptsize{±0.006}} & \textbf{\small{0.910}\scriptsize{±0.009}} & \textbf{\small{0.961}\scriptsize{±0.027}}  &
   \\ 
\bottomrule

\end{tabular*}
\end{table}

\begin{table}[h]
\caption{Classification with real data. We report Weighted-F1, which is inversely weighted to its class size, for binary classification.}
\centering
\small
\setlength\tabcolsep{2.5pt}
\label{tab:weighted_f1_binary}
\begin{tabular*}{1\textwidth}{lccccccccc}
\toprule
\multirow{2}{*}{\small{Methods}}  &  & \multicolumn{7}{c}{\small{Binary classification}}   &    \\ \cmidrule{3-9}
&  &  \texttt{\small{Credit}} & \texttt{\small{Default}}       &   \texttt{\small{HTRU}}  &  \texttt{\small{Magic}}  &  \small{\texttt{Phishing}}   &  \small{\texttt{Shoppers}} & \small{\texttt{Spambase}} &  \\ 
\midrule
\texttt{\small{Identity}}        & & \small{0.775}\scriptsize{±0.071} & \small{0.549}\scriptsize{±0.034} & \small{0.894}\scriptsize{±0.009} & \small{0.822}\scriptsize{±0.047} & \small{0.954}\scriptsize{±0.021} & \small{0.659}\scriptsize{±0.055} & \small{0.949}\scriptsize{±0.019} & 
    \\ 
\midrule
\texttt{\small{MedGAN}}       & &  \small{0.004}\scriptsize{±0.005} & \small{0.286}\scriptsize{±0.003} & \small{0.514}\scriptsize{±0.257} & \small{0.549}\scriptsize{±0.091} & \small{0.833}\scriptsize{±0.041} & \small{0.461}\scriptsize{±0.065} & \small{0.616}\scriptsize{±0.053} & 
   \\ 
\texttt{\small{VEEGAN}}        & &  \small{0.002}\scriptsize{±0.000} & \small{0.197}\scriptsize{±0.000} & \small{0.115}\scriptsize{±0.061} & \small{0.498}\scriptsize{±0.180} & \small{0.345}\scriptsize{±0.008} & \small{0.369}\scriptsize{±0.139} & \small{0.584}\scriptsize{±0.131} & 
   \\ 
\texttt{\small{CTGAN}}        & & \small{0.758}\scriptsize{±0.064} & \small{0.579}\scriptsize{±0.009} & \small{0.866}\scriptsize{±0.007} & \small{0.784}\scriptsize{±0.027} & \small{0.908}\scriptsize{±0.008} & \small{0.562}\scriptsize{±0.048} & \small{0.814}\scriptsize{±0.016} &
  \\ 
\texttt{\small{TVAE}}        & & \small{0.002}\scriptsize{±0.000} & \small{0.540}\scriptsize{±0.021} & \small{0.866}\scriptsize{±0.005} & \small{0.740}\scriptsize{±0.008} & \small{0.921}\scriptsize{±0.006} & \small{0.635}\scriptsize{±0.034} & \small{0.777}\scriptsize{±0.033} &
  \\ 
\texttt{\small{TableGAN}}       & &  \small{0.739}\scriptsize{±0.038} & \small{0.510}\scriptsize{±0.009} & \small{0.856}\scriptsize{±0.004} & \small{0.789}\scriptsize{±0.035} & \small{0.913}\scriptsize{±0.008} & \small{0.649}\scriptsize{±0.054} & \small{0.792}\scriptsize{±0.076} &
    \\ 
\texttt{\small{OCT-GAN}}        & & \small{0.128}\scriptsize{±0.187} & \small{0.562}\scriptsize{±0.018} & \small{0.876}\scriptsize{±0.009} & \small{0.772}\scriptsize{±0.015} & \small{0.913}\scriptsize{±0.006} & \small{0.666}\scriptsize{±0.033} & \small{0.879}\scriptsize{±0.019} & 
  \\ 
\texttt{\small{RNODE}}       & & \small{0.119}\scriptsize{±0.110} & \small{0.512}\scriptsize{±0.010} & \small{0.663}\scriptsize{±0.043} & \small{0.789}\scriptsize{±0.029} & \small{0.908}\scriptsize{±0.003} & \small{0.597}\scriptsize{±0.021} & \small{0.855}\scriptsize{±0.039} &
  \\ 
\midrule
\texttt{\small{Na\"ive-STaSy}}       & & \underline{\small{0.782}\scriptsize{±0.042}} & \underline{\small{0.592}\scriptsize{±0.011}} & \underline{\small{0.895}\scriptsize{±0.008}} & \underline{\small{0.819}\scriptsize{±0.043}} & \underline{\small{0.932}\scriptsize{±0.009}} & \underline{\small{0.678}\scriptsize{±0.015}} & \underline{\small{0.902}\scriptsize{±0.021}} & 
   \\ 
\texttt{\small{STaSy}}  && \textbf{\small{0.794}\scriptsize{±0.035}} & \textbf{\small{0.600}\scriptsize{±0.005}} & \textbf{\small{0.895}\scriptsize{±0.006}} & \textbf{\small{0.820}\scriptsize{±0.044}} & \textbf{\small{0.938}\scriptsize{±0.012}} & \textbf{\small{0.681}\scriptsize{±0.013}} & \textbf{\small{0.910}\scriptsize{±0.020}} &  
  \\ 
\bottomrule

\end{tabular*}
\end{table}

\begin{table}[h]
\caption{Classification with real data. We report macro F1 for multi-class classification.}
\label{tab:macro_f1}
\centering
\small
\setlength\tabcolsep{2.5pt}
% \label{tbl:main}
\begin{tabular}{lcccccccccc}
\toprule
\multirow{2}{*}{\small{Methods}}  &  & \multicolumn{6}{c}{\small{Multi-class classification}}   & \\ \cmidrule{3-8} 
&  & \texttt{\small{Bean}}  & \texttt{\small{Contraceptive}} & \texttt{\small{Crowdsource}}       &   \texttt{\small{Obesity}}  &  \texttt{\small{Robot}}  &  \small{\texttt{Shuttle}}   &  \\ 
\midrule
\texttt{\small{Identity}}        & &\small{0.934}\scriptsize{±0.011} & \small{0.498}\scriptsize{±0.017} & \small{0.772}\scriptsize{±0.094} & \small{0.969}\scriptsize{±0.008} & \small{0.975}\scriptsize{±0.038} & \small{0.932}\scriptsize{±0.067}  &
    \\ 
\midrule
	
\texttt{\small{MedGAN}}       & & \small{0.058}\scriptsize{±0.000} & \underline{\small{0.423}\scriptsize{±0.010}} & \small{0.174}\scriptsize{±0.020} & \small{0.208}\scriptsize{±0.078} & \small{0.301}\scriptsize{±0.049} & \small{0.126}\scriptsize{±0.000}  &
\\ 
\texttt{\small{VEEGAN}}        & &\small{0.332}\scriptsize{±0.063} & \small{0.343}\scriptsize{±0.050} & \small{0.196}\scriptsize{±0.017} & \small{0.183}\scriptsize{±0.041} & \small{0.307}\scriptsize{±0.016} & \small{0.232}\scriptsize{±0.025}  &
   \\ 
\texttt{\small{CTGAN}}        & &\small{0.883}\scriptsize{±0.015} & \small{0.364}\scriptsize{±0.006} & \small{0.557}\scriptsize{±0.075} & \small{0.144}\scriptsize{±0.016} & \small{0.651}\scriptsize{±0.040} & \small{0.675}\scriptsize{±0.149}  & 
  \\ 
\texttt{\small{TVAE}}        & &\small{0.895}\scriptsize{±0.010} & \small{0.406}\scriptsize{±0.011} & \small{0.578}\scriptsize{±0.076} & \small{0.462}\scriptsize{±0.020} & \small{0.776}\scriptsize{±0.042} & \small{0.468}\scriptsize{±0.051}  &
 \\ 
\texttt{\small{TableGAN}}       & &\small{0.620}\scriptsize{±0.024} & \small{0.376}\scriptsize{±0.016} & \small{0.427}\scriptsize{±0.056} & \small{0.313}\scriptsize{±0.043} & \small{0.422}\scriptsize{±0.028} & \small{0.404}\scriptsize{±0.008}  &
\\ 
\texttt{\small{OCT-GAN}}        & &\small{0.916}\scriptsize{±0.014} & \small{0.394}\scriptsize{±0.013} & \small{0.594}\scriptsize{±0.113} & \small{0.274}\scriptsize{±0.031} & \small{0.782}\scriptsize{±0.036} & \small{0.637}\scriptsize{±0.079}  &
 \\ 
\texttt{\small{RNODE}}       & &\small{0.797}\scriptsize{±0.110} & \small{0.385}\scriptsize{±0.022} & \small{0.370}\scriptsize{±0.050} & \small{0.461}\scriptsize{±0.051} & \small{0.503}\scriptsize{±0.118} & \small{0.408}\scriptsize{±0.007}  & 
   \\ 
\midrule
\texttt{\small{Na\"ive-STaSy}}      & & \underline{\small{0.933}\scriptsize{±0.010}} & {\small{0.418}\scriptsize{±0.018}} & \underline{\small{0.710}\scriptsize{±0.104}} & \underline{\small{0.904}\scriptsize{±0.024}} & \underline{\small{0.943}\scriptsize{±0.028}} & \underline{\small{0.791}\scriptsize{±0.056}}  &
   \\ 
\texttt{\small{STaSy}} && \textbf{\small{0.935}\scriptsize{±0.009}} & \textbf{\small{0.451}\scriptsize{±0.017}}& \textbf{\small{0.743}\scriptsize{±0.104}} & \textbf{\small{0.910}\scriptsize{±0.025}} & \textbf{\small{0.944}\scriptsize{±0.032}} & \textbf{\small{0.865}\scriptsize{±0.072}}  &
\\ 

\bottomrule
\end{tabular}
\end{table}

\begin{table}[h]
\caption{Classification with real data. We report AUROC for multi-class classification.}
\centering
\small
\setlength\tabcolsep{2.5pt}
\label{tab:AUROC_multi}
\begin{tabular*}{1\textwidth}{lcccccccc}
\toprule
\multirow{2}{*}{\small{Methods}}  &  & \multicolumn{6}{c}{\small{Multi-class classification}}   & \\ \cmidrule{3-8}
&  & \texttt{\small{Bean}}  & \texttt{\small{Contraceptive}} & \texttt{\small{Crowdsource}}       &   \texttt{\small{Obesity}}  &  \texttt{\small{Robot}}  &  \small{\texttt{Shuttle}}   &  \\ 
\midrule
\texttt{\small{Identity}}        & &  \small{0.992}\scriptsize{±0.006} & \small{0.697}\scriptsize{±0.011} & \small{0.953}\scriptsize{±0.052} & \small{0.998}\scriptsize{±0.002} & \small{0.995}\scriptsize{±0.007} & \small{0.999}\scriptsize{±0.002} &  
    \\ 
\midrule
\texttt{\small{MedGAN}}       & &  \small{0.500}\scriptsize{±0.000} & \small{0.640}\scriptsize{±0.020} & \small{0.643}\scriptsize{±0.074} & \small{0.711}\scriptsize{±0.095} & \small{0.646}\scriptsize{±0.082} & \small{0.490}\scriptsize{±0.021} &
    \\ 
\texttt{\small{VEEGAN}}        & & \small{0.765}\scriptsize{±0.107} & \small{0.621}\scriptsize{±0.043} & \small{0.650}\scriptsize{±0.075} & \small{0.588}\scriptsize{±0.040} & \small{0.604}\scriptsize{±0.051} & \small{0.648}\scriptsize{±0.051} & 
   \\ 
\texttt{\small{CTGAN}}        & &\small{0.985}\scriptsize{±0.010} & \small{0.565}\scriptsize{±0.008} & \small{0.889}\scriptsize{±0.043} & \small{0.526}\scriptsize{±0.007} & \small{0.828}\scriptsize{±0.034} & \underline{\small{0.984}\scriptsize{±0.018}} & 
    \\ 
\texttt{\small{TVAE}}        & & \small{0.985}\scriptsize{±0.008} & \small{0.621}\scriptsize{±0.025} & \small{0.854}\scriptsize{±0.031} & \small{0.835}\scriptsize{±0.010} & \small{0.952}\scriptsize{±0.029} & \small{0.872}\scriptsize{±0.048} & 
 \\ 
\texttt{\small{TableGAN}}       & & \small{0.913}\scriptsize{±0.012} & \small{0.576}\scriptsize{±0.006} & \small{0.882}\scriptsize{±0.051} & \small{0.755}\scriptsize{±0.026} & \small{0.723}\scriptsize{±0.018} & \small{0.802}\scriptsize{±0.083} & 
   \\ 
\texttt{\small{OCT-GAN}}        & &  \small{0.989}\scriptsize{±0.009} & \small{0.588}\scriptsize{±0.015} & \small{0.915}\scriptsize{±0.054} & \small{0.686}\scriptsize{±0.036} & \small{0.927}\scriptsize{±0.027} & \small{0.968}\scriptsize{±0.037} &  
   \\ 
\texttt{\small{RNODE}}       & & \small{0.971}\scriptsize{±0.030} & \small{0.579}\scriptsize{±0.028} & \small{0.752}\scriptsize{±0.043} & \small{0.841}\scriptsize{±0.032} & \small{0.851}\scriptsize{±0.060} & \small{0.712}\scriptsize{±0.001} &
   \\ 
\midrule
\texttt{\small{Na\"ive-STaSy}}       & &
\underline{\small{0.991}\scriptsize{±0.008}} & 
\underline{\small{0.642}\scriptsize{±0.022}} & 
\underline{\small{0.948}\scriptsize{±0.048}} & 
\underline{\small{0.987}\scriptsize{±0.013}} & 
\textbf{\small{0.989}\scriptsize{±0.011}} & 
\small{0.984}\scriptsize{±0.027} & 
\\ 
\texttt{\small{STaSy}}  & & 
\textbf{\small{0.992}\scriptsize{±0.007}} & 
\textbf{\small{0.680}\scriptsize{±0.014}} & 
\textbf{\small{0.949}\scriptsize{±0.050}} & 
\textbf{\small{0.987}\scriptsize{±0.012}} & 
\underline{\small{0.989}\scriptsize{±0.011}} & 
\textbf{\small{0.991}\scriptsize{±0.014}} & 
  \\ 
\bottomrule
\end{tabular*}
\end{table}

\begin{table}[h]
\caption{Classification with real data. We report Weighted-F1, which is inversely weighted to its class size, for multi-class classification.}
\centering
\small
\setlength\tabcolsep{2.5pt}
\label{tab:weighted_f1_multi}
\begin{tabular*}{1\textwidth}{lcccccccc}
\toprule
\multirow{2}{*}{\small{Methods}}  &  & \multicolumn{6}{c}{\small{Multi-class classification}}   & \\ \cmidrule{3-8}
&  & \texttt{\small{Bean}}  & \texttt{\small{Contraceptive}} & \texttt{\small{Crowdsource}}       &   \texttt{\small{Obesity}}  &  \texttt{\small{Robot}}  &  \small{\texttt{Shuttle}}   &  \\ 
\midrule
\texttt{\small{Identity}}        & &\small{0.936}\scriptsize{±0.011} & \small{0.481}\scriptsize{±0.016} & \small{0.746}\scriptsize{±0.103} & \small{0.969}\scriptsize{±0.008} & \small{0.974}\scriptsize{±0.040} & \small{0.922}\scriptsize{±0.078} & 
\\ 
\midrule
\texttt{\small{MedGAN}}       & &\small{0.050}\scriptsize{±0.000} & \underline{\small{0.402}\scriptsize{±0.011}} & \small{0.096}\scriptsize{±0.024} & \small{0.209}\scriptsize{±0.079} & \small{0.257}\scriptsize{±0.056} & \small{0.032}\scriptsize{±0.000} &
   \\ 
\texttt{\small{VEEGAN}}        & &\small{0.335}\scriptsize{±0.066} & \small{0.318}\scriptsize{±0.056} & \small{0.119}\scriptsize{±0.020} & \small{0.182}\scriptsize{±0.041} & \small{0.268}\scriptsize{±0.016} & \small{0.147}\scriptsize{±0.027} &  
\\
\texttt{\small{CTGAN}}        & &\small{0.885}\scriptsize{±0.015} & \small{0.347}\scriptsize{±0.008} & \small{0.509}\scriptsize{±0.082} & \small{0.145}\scriptsize{±0.017} & \small{0.651}\scriptsize{±0.048} & \small{0.623}\scriptsize{±0.173} & 
    \\ 
\texttt{\small{TVAE}}        & &\small{0.896}\scriptsize{±0.011} & \small{0.387}\scriptsize{±0.010} & \small{0.528}\scriptsize{±0.080} & \small{0.461}\scriptsize{±0.020} & \small{0.767}\scriptsize{±0.044} & \small{0.380}\scriptsize{±0.059} & 
    \\ 
\texttt{\small{TableGAN}}       & &\small{0.620}\scriptsize{±0.025} & \small{0.366}\scriptsize{±0.018} & \small{0.361}\scriptsize{±0.058} & \small{0.316}\scriptsize{±0.043} & \small{0.397}\scriptsize{±0.032} & \small{0.314}\scriptsize{±0.010} &
    \\ 
\texttt{\small{OCT-GAN}}        & &\small{0.918}\scriptsize{±0.014} & \small{0.385}\scriptsize{±0.013} & \small{0.548}\scriptsize{±0.122} & \small{0.273}\scriptsize{±0.031} & \small{0.780}\scriptsize{±0.035} & \small{0.581}\scriptsize{±0.092} &  
  \\ 
\texttt{\small{RNODE}}       & &\small{0.795}\scriptsize{±0.116} & \small{0.370}\scriptsize{±0.022} & \small{0.294}\scriptsize{±0.050} & \small{0.461}\scriptsize{±0.050} & \small{0.474}\scriptsize{±0.137} & \small{0.316}\scriptsize{±0.006} &
   \\ 
\midrule
\texttt{\small{Na\"ive-STaSy}}       &&
\underline{\small{0.935}\scriptsize{±0.010}} & 
\small{0.398}\scriptsize{±0.019} & 
\underline{\small{0.677}\scriptsize{±0.114}} & 
\underline{\small{0.903}\scriptsize{±0.024}} & 
\underline{\small{0.941}\scriptsize{±0.029}} & 
\underline{\small{0.757}\scriptsize{±0.066}} & 
 \\ 
\texttt{\small{STaSy}}  & &\textbf{\small{0.936}\scriptsize{±0.009}} &
\textbf{\small{0.433}\scriptsize{±0.018}} & 
\textbf{\small{0.715}\scriptsize{±0.113}} & 
\textbf{\small{0.910}\scriptsize{±0.025}} & 
\textbf{\small{0.943}\scriptsize{±0.033}} & 
\textbf{\small{0.843}\scriptsize{±0.084}} & 
\\
\bottomrule
\end{tabular*}
\end{table}

\begin{table}[h]
\caption{Regression with real data. We report $R^2$ and RMSE for regression.}
\centering
\small
\label{tab:regression}
\begin{tabular}{lcccccccc}
\toprule
\multirow{2}{*}{\small{Methods}}  && \multicolumn{2}{c}{\small{$R^2$}} && \multicolumn{2}{c}{\small{RMSE}} & \\ \cmidrule{3-4} \cmidrule{6-7}
  & & \small{\texttt{Beijing}} & \small{\texttt{News}}   & & \small{\texttt{Beijing}} & \small{\texttt{News}} &  \\ 
\midrule
\texttt{\small{Identity}}     &&\small{0.728}\scriptsize{±0.196}&\small{0.004}\scriptsize{±0.038}&&\small{40.935}\scriptsize{±13.809}&\small{8177.913}\scriptsize{±156.310}

\\ 
\midrule
\texttt{\small{MedGAN}} &&\small{-48.753}\scriptsize{±26.307}&\small{-inf}&&\small{552.597}\scriptsize{±192.765}&\small{inf}

\\ 
\texttt{\small{VEEGAN}} &&\small{-1.039}\scriptsize{±0.002}&\small{-3.699}\scriptsize{±2.771}&&\small{116.869}\scriptsize{±0.056}&\small{16964.690}\scriptsize{±6020.151}

  \\ 
\texttt{\small{CTGAN}}   &&\small{0.207}\scriptsize{±0.017}&\small{0.019}\scriptsize{±0.011}&&\small{72.854}\scriptsize{±0.768}&\small{8119.705}\scriptsize{±45.940}

& 
  \\ 
\texttt{\small{TVAE}}  &&\small{0.256}\scriptsize{±0.033}&\small{-0.042}\scriptsize{±0.010}&&\small{70.598}\scriptsize{±1.553}&\small{8366.315}\scriptsize{±41.758}

 & 
 \\ 
\texttt{\small{TableGAN}}  && \small{0.155}\scriptsize{±0.027}&\small{-0.206}\scriptsize{±0.103}&&\small{75.208}\scriptsize{±1.194}&\small{8995.820}\scriptsize{±380.286}

 & 
\\ 
\texttt{\small{OCT-GAN}} &&\small{0.307}\scriptsize{±0.021}&\small{0.013}\scriptsize{±0.016}&&\small{68.133}\scriptsize{±1.011}&\small{8143.325}\scriptsize{±64.937}

 & 
 \\ 
\texttt{\small{RNODE}}   &&\small{0.250}\scriptsize{±0.029}&\small{0.010}\scriptsize{±0.008}&&\small{70.869}\scriptsize{±1.369}&\small{8154.987}\scriptsize{±33.918}

 & 
  \\ 
\midrule
\texttt{\small{Na\"ive-STaSy}}  &&
\underline{\small{0.625}\scriptsize{±0.141}}&
\underline{\small{0.023}\scriptsize{±0.011}}&&
\underline{\small{49.524}\scriptsize{±8.963}}&
\underline{\small{8103.668}\scriptsize{±45.072}}

 & 
  \\ 
\texttt{\small{STaSy}} &&
\textbf{\small{0.672}\scriptsize{±0.166}}&
\textbf{\small{0.029}\scriptsize{±0.008}}&&
\textbf{\small{46.123}\scriptsize{±10.830}}&
\textbf{\small{8097.163}\scriptsize{±33.137}}

 & \\ 
\bottomrule
\end{tabular}
\end{table}

\begin{table}[h]
\caption{The median of the log-probabilities of testing records are reported.}
\centering
\label{tab:probability}
\begin{tabular}{llcccccccc}
\toprule
& Datasets  &  &  \texttt{RNODE}  & \texttt{STaSy} w/o fine-tuning & \texttt{STaSy} &  \\ 
\midrule
\multirow{7}{*}{\rotatebox[origin=c]{90}{Binary}} &\texttt{Credit}  & & 71.248&117.001&\textbf{117.404}
 \\ 
&\texttt{Default} & & 55.284&131.011  &  \textbf{131.122}\\ 
&\texttt{HTRU} & &  \textbf{27.852}&5.435   &     5.326 \\
&\texttt{Magic} & & 17.413&30.247  &     \textbf{30.216}\\
&\texttt{Phishing} & & 53.478&228.912 &  \textbf{228.914}\\ 
&\texttt{Shoppers} & &  62.843&189.422 & \textbf{220.589}\\ 
&\texttt{Spambase} & & 140.686&253.935 & \textbf{253.798}
 \\ 
\midrule
\multirow{6}{*}{\rotatebox[origin=c]{90}{Multi.}}
&\texttt{Bean}  & & 46.764&92.555&              \textbf{92.635} \\ 
&\texttt{Contraceptive} & & 26.797&225.334&     \textbf{225.638} \\ 
&\texttt{Crowdsource} & & 23.323&47.567&        \textbf{48.905} \\ 
&\texttt{Obesity}& &130.018&330.003&            \textbf{333.017}     \\ 
&\texttt{Robot} & & 14.238&79.233&              \textbf{79.355}\\ 
&\texttt{Shuttle} & &48.875&114.247&            \textbf{114.294}
\\
\midrule
\multirow{2}{*}{\rotatebox[origin=c]{90}{Reg.}} &\texttt{Beijing} & &  \textbf{42.614}&17.720&18.192
\\ 
&\texttt{News} & &  \textbf{128.472}&76.770&76.600
\\ 
\bottomrule
% \label{tab:statistical1}
\end{tabular}
\end{table}

\clearpage
\subsection{Sampling diversity}\label{sec:additional_diversity}
We use the coverage score as a metric for the sampling diversity. Full results for all datasets are in Tables~\ref{tab:coverage_binary}, ~\ref{tab:coverage_multi}, and ~\ref{tab:coverage_reg}. We measure the coverage score 5 times with different fake records and report their mean and standard deviation.

Coverage is bounded between 0 and 1, and higher coverage means more diverse samples. This k-NN-based measurement is expected to achieve 100\% performance when the real and fake records are identical, but in practice, this is not always the case. For the dataset whose coverage score does not show 1 for two same data records, we choose the hyperparameter $k$ to achieve at least greater than 0.95. In our experiments, $k$ for \texttt{Phishing} is 7, and for others, $k$ is 5. As shown in Tables~\ref{tab:coverage_binary},~\ref{tab:coverage_multi}, and ~\ref{tab:coverage_reg}, in 12 out of 15 datasets, our methods outperform others by large margins.

\begin{table}[h]
\caption{Sampling diversity in terms of coverage for binary classification datasets}
\centering
\small
\setlength\tabcolsep{2.5pt}
\label{tab:coverage_binary}
\resizebox{1\textwidth}{!}{
\begin{tabular}{lccccccccc}
\toprule
\multirow{2}{*}{\small{Methods}}  &  & \multicolumn{7}{c}{\small{Binary classification}}   &    \\ \cmidrule{3-9}
&  &  \texttt{\small{Credit}} & \texttt{\small{Default}}       &   \texttt{\small{HTRU}}  &  \texttt{\small{Magic}}  &  \small{\texttt{Phishing}}   &  \small{\texttt{Shoppers}} & \small{\texttt{Spambase}} &  \\ 
\midrule
\texttt{\small{MedGAN}}       & & 
\small{0.000}\scriptsize{±0.000}&\small{0.000}\scriptsize{±0.000}&\small{0.000}\scriptsize{±0.000}&\small{0.001}\scriptsize{±0.000}&\small{0.002}\scriptsize{±0.001}&\small{0.000}\scriptsize{±0.000}&\small{0.002}\scriptsize{±0.000}&

    \\ 
\texttt{\small{VEEGAN}}        &&
\small{0.000}\scriptsize{±0.000}&\small{0.000}\scriptsize{±0.000}&\small{0.000}\scriptsize{±0.000}&\small{0.003}\scriptsize{±0.000}&\small{0.035}\scriptsize{±0.000}&\small{0.141}\scriptsize{±0.002}&\small{0.201}\scriptsize{±0.004}&

 \\ 
\texttt{\small{CTGAN}}        &&
\small{0.174}\scriptsize{±0.001}&\small{0.190}\scriptsize{±0.001}&\small{0.461}\scriptsize{±0.004}&\small{0.655}\scriptsize{±0.003}&\small{0.416}\scriptsize{±0.001}&\small{0.723}\scriptsize{±0.006}&\small{0.471}\scriptsize{±0.007}&

  \\ 
\texttt{\small{TVAE}}        & & 
\underline{\small{0.373}\scriptsize{±0.001}}&\underline{\small{0.272}\scriptsize{±0.002}}&\small{0.741}\scriptsize{±0.001}&\small{0.650}\scriptsize{±0.003}&\small{0.623}\scriptsize{±0.002}&\small{0.737}\scriptsize{±0.001}&\small{0.698}\scriptsize{±0.002}&

    \\ 
\texttt{\small{TableGAN}}       & &
\textbf{\small{0.458}\scriptsize{±0.002}}&\small{0.216}\scriptsize{±0.001}&\small{0.299}\scriptsize{±0.002}&\small{0.748}\scriptsize{±0.002}&\small{0.512}\scriptsize{±0.007}&\small{0.698}\scriptsize{±0.005}&\small{0.711}\scriptsize{±0.005}&

  \\ 
\texttt{\small{OCT-GAN}}        & &
\small{0.000}\scriptsize{±0.000}&\small{0.171}\scriptsize{±0.000}&\small{0.375}\scriptsize{±0.004}&\small{0.634}\scriptsize{±0.001}&\small{0.450}\scriptsize{±0.003}&\small{0.714}\scriptsize{±0.003}&\small{0.470}\scriptsize{±0.016}&

 \\ 
\texttt{\small{RNODE}}       & & 
\small{0.025}\scriptsize{±0.000}&\textbf{\small{0.293}\scriptsize{±0.003}}&\small{0.751}\scriptsize{±0.006}&\small{0.612}\scriptsize{±0.003}&\small{0.272}\scriptsize{±0.008}&\small{0.447}\scriptsize{±0.004}&\small{0.326}\scriptsize{±0.009}&

  \\ 
\midrule
\texttt{\small{Na\"ive-STaSy}}       && 
\small{0.014}\scriptsize{±0.000}&\small{0.149}\scriptsize{±0.001}&\textbf{\small{0.921}\scriptsize{±0.002}}&\small{0.919}\scriptsize{±0.002}&\small{0.655}\scriptsize{±0.006}&\textbf{\small{0.832}\scriptsize{±0.005}}&\small{0.684}\scriptsize{±0.011}&

    \\ 
\small{\texttt{STaSy} w/o fine-tuning } & &
\small{0.013}\scriptsize{±0.000}&\small{0.083}\scriptsize{±0.001}&\small{0.907}\scriptsize{±0.002}&\underline{ \small{0.943}\scriptsize{±0.003}}&\textbf{\small{0.780}\scriptsize{±0.005}}&\small{0.796}\scriptsize{±0.006}&\textbf{\small{0.727}\scriptsize{±0.006}}&

\\
\small{\texttt{STaSy}}  & & \small{0.014}\scriptsize{±0.000}&\small{0.101}\scriptsize{±0.001}&\underline{\small{0.911}\scriptsize{±0.003}}&\textbf{\small{0.944}\scriptsize{±0.001}}&\underline{\small{0.779}\scriptsize{±0.005}}&\underline{\small{0.798}\scriptsize{±0.004}}&\underline{\small{0.727}\scriptsize{±0.013}}&

  \\ 
\bottomrule
\end{tabular}}
\end{table}

\begin{table}[h]
\caption{Sampling diversity in terms of coverage for multi-class classification datasets}
\label{tab:coverage_multi}
\centering
\small
\setlength\tabcolsep{2.5pt}
% \label{tbl:main}
\resizebox{1\textwidth}{!}{
\begin{tabular}{lccccccccc}
\toprule
\multirow{2}{*}{\small{Methods}}  &  & \multicolumn{6}{c}{\small{Multi-class classification}}   &  \\ \cmidrule{3-8}
&  & \texttt{\small{Bean}}  & \texttt{\small{Contraceptive}} & \texttt{\small{Crowdsource}}       &   \texttt{\small{Obesity}}  &  \texttt{\small{Robot}}  &  \small{\texttt{Shuttle}}   & \\ 
\midrule
\texttt{\small{MedGAN}}       & & \small{0.000}\scriptsize{±0.000}&\small{0.538}\scriptsize{±0.007}&\small{0.000}\scriptsize{±0.000}&\small{0.007}\scriptsize{±0.001}&\small{0.000}\scriptsize{±0.000}&\small{0.000}\scriptsize{±0.000}&

    \\ 
\texttt{\small{VEEGAN}}        && \small{0.004}\scriptsize{±0.001}&\small{0.082}\scriptsize{±0.002}&\small{0.000}\scriptsize{±0.000}&\small{0.100}\scriptsize{±0.003}&\small{0.005}\scriptsize{±0.001}&\small{0.001}\scriptsize{±0.000}&

 \\ 
\texttt{\small{CTGAN}}        && \small{0.053}\scriptsize{±0.001}&\small{0.753}\scriptsize{±0.011}&\small{0.064}\scriptsize{±0.000}&\small{0.252}\scriptsize{±0.015}&\small{0.140}\scriptsize{±0.004}&\small{0.031}\scriptsize{±0.000}&

  \\ 
\texttt{\small{TVAE}}        & & \underline{\small{0.118}\scriptsize{±0.001}}&\small{0.680}\scriptsize{±0.011}&\small{0.254}\scriptsize{±0.003}&\small{0.297}\scriptsize{±0.003}&\small{0.472}\scriptsize{±0.014}&\small{0.111}\scriptsize{±0.002}&

    \\ 
\texttt{\small{TableGAN}}       && \small{0.116}\scriptsize{±0.003}&\small{0.751}\scriptsize{±0.001}&\small{0.339}\scriptsize{±0.001}&\small{0.440}\scriptsize{±0.006}&\small{0.259}\scriptsize{±0.003}&\small{0.005}\scriptsize{±0.000}&

   \\ 
\texttt{\small{OCT-GAN}}        && \small{0.106}\scriptsize{±0.001}&\small{0.764}\scriptsize{±0.011}&\small{0.133}\scriptsize{±0.000}&\small{0.345}\scriptsize{±0.010}&\small{0.250}\scriptsize{±0.004}&\small{0.023}\scriptsize{±0.000}&

 \\ 
\texttt{\small{RNODE}}       & & \textbf{\small{0.292}\scriptsize{±0.006}}&\small{0.547}\scriptsize{±0.011}&\small{0.104}\scriptsize{±0.003}&\small{0.375}\scriptsize{±0.008}&\small{0.113}\scriptsize{±0.004}&\small{0.005}\scriptsize{±0.000}&

   \\ 
\midrule
\texttt{\small{Na\"ive-STaSy}}       & & \small{0.092}\scriptsize{±0.003}&\small{0.839}\scriptsize{±0.012}&\small{0.920}\scriptsize{±0.004}&\textbf{\small{0.825}\scriptsize{±0.006}}&\small{0.935}\scriptsize{±0.007}&\small{0.133}\scriptsize{±0.003}&
\\
\small{\texttt{STaSy} w/o fine-tuning}       & &
\small{0.095}\scriptsize{±0.005}&\underline{\small{0.879}\scriptsize{±0.007}}&\underline{\small{0.970}\scriptsize{±0.002}}&\small{0.777}\scriptsize{±0.013}&\underline{\small{0.937}\scriptsize{±0.004}}&\underline{\small{0.207}\scriptsize{±0.001}}&

    \\ 
\texttt{\small{STaSy}}  & &\small{0.100}\scriptsize{±0.005}&\textbf{\small{0.894}\scriptsize{±0.010}}&\textbf{\small{0.971}\scriptsize{±0.005}}&\underline{\small{0.778}\scriptsize{±0.008}}&\textbf{\small{0.937}\scriptsize{±0.004}}&\textbf{\small{0.209}\scriptsize{±0.001}}&

   \\ 
\bottomrule
\end{tabular}}
\end{table}

\begin{table}[h]
\caption{Sampling diversity in terms of coverage for regression datasets}
\label{tab:coverage_reg}
\centering
% \small
% \setlength\tabcolsep{2.5pt}
% \label{tbl:main}
\begin{tabular}{lcccc}
\toprule
\multirow{2}{*}{\small{Methods}}  &  & \multicolumn{2}{c}{\small{Regression}}   &  \\ \cmidrule{3-4}
&  & \texttt{\small{Beijing}}  & \texttt{\small{News}} &  \\ 
\midrule
\texttt{\small{MedGAN}}       & & 0.000\footnotesize{±0.000}&0.000\footnotesize{±0.000}&

    \\ 
\texttt{\small{VEEGAN}}        & &0.000\footnotesize{±0.000}&0.002\footnotesize{±0.000}&

 \\ 
\texttt{\small{CTGAN}}        & &0.532\footnotesize{±0.002}&0.366\footnotesize{±0.001}&

  \\ 
\texttt{\small{TVAE}}        & & 0.720\footnotesize{±0.002}&0.665\footnotesize{±0.001}&

    \\ 
\texttt{\small{TableGAN}}       & &0.803\footnotesize{±0.003}&0.154\footnotesize{±0.003}&

   \\ 
\texttt{\small{OCT-GAN}}        & &0.693\footnotesize{±0.000}&0.582\footnotesize{±0.000}&
 \\ 
\texttt{\small{RNODE}}       & &  0.501\footnotesize{±0.003}&0.255\footnotesize{±0.002}&

   \\ 
\midrule
\texttt{\small{Na\"ive-STaSy}}       & &0.876\footnotesize{±0.003}&0.755\footnotesize{±0.003}&

    \\ 
{\small{\texttt{STaSy}} w/o fine-tuning}  & &  \textbf{0.943\footnotesize{±0.003}}&\underline{0.762\footnotesize{±0.004}}&

   \\
\texttt{\small{STaSy}}  & &\underline{0.941\footnotesize{±0.003}}&\textbf{0.762\footnotesize{±0.002}}&

   \\ 
\bottomrule
\end{tabular}
\end{table}

% \clearpage
\subsection{Sampling time}
Tables~\ref{tab:runtime_binary}, ~\ref{tab:runtime_multi}, and ~\ref{tab:runtime_reg} show runtime evaluation results of each method. We measure the wall-clock time taken to sample fake records 5 times, and report their mean and standard deviation. In almost all datasets, \texttt{Na\"ive-STaSy} and \texttt{STaSy} show faster runtime than \texttt{OCT-GAN} and \texttt{RNODE}. \texttt{TableGAN} and \texttt{TVAE} take a short sampling time, but considering their inferior sampling quality and diversity, only our proposed model resolves the problems of \textit{the generative learning trilemma}.

\begin{table}[h]
\caption{Wall-clock runtime for binary classification datasets}
\centering
\small
\setlength\tabcolsep{2.5pt}
\label{tab:runtime_binary}
\resizebox{1\textwidth}{!}{
\begin{tabular}{lccccccccc}
\toprule
\multirow{2}{*}{\small{Methods}}  &  & \multicolumn{7}{c}{\small{Binary classification}}   &    \\ \cmidrule{3-9}
&  &  \texttt{\small{Credit}} & \texttt{\small{Default}}       &   \texttt{\small{HTRU}}  &  \texttt{\small{Magic}}  &  \small{\texttt{Phishing}}   &  \small{\texttt{Shoppers}} & \small{\texttt{Spambase}} &  \\ 
\midrule
\texttt{\small{MedGAN}}       & 
&\underline{\small{0.756}\scriptsize{±0.328}}
&\small{0.223}\scriptsize{±0.334}
&\small{0.201}\scriptsize{±0.332}
&\small{0.205}\scriptsize{±0.333}
&\small{0.189}\scriptsize{±0.334}
&\small{0.203}\scriptsize{±0.332}
&\small{0.187}\scriptsize{±0.328}&
\\				
\texttt{\small{VEEGAN}}        & 
&\small{0.800}\scriptsize{±0.031} 
&\small{0.112}\scriptsize{±0.008}
&\small{0.051}\scriptsize{±0.005}
&\small{0.061}\scriptsize{±0.007} 
&\small{0.041}\scriptsize{±0.009} 
&\small{0.041}\scriptsize{±0.006} 
&\small{0.019}\scriptsize{±0.002}&
 \\ 					
\texttt{\small{CTGAN}}  & 
&\small{6.342}\scriptsize{±0.300} 
&\small{0.546}\scriptsize{±0.054} 
&\small{0.299}\scriptsize{±0.085}
&\small{0.284}\scriptsize{±0.002}
&\small{0.163}\scriptsize{±0.005}
&\small{0.204}\scriptsize{±0.005}
&\small{0.107}\scriptsize{±0.008}&
  \\ 
\texttt{\small{TVAE}}        &
&\small{0.937}\scriptsize{±0.185} 
&\underline{\small{0.052}\scriptsize{±0.005}} 
&\underline{\small{0.028}\scriptsize{±0.001}}
&\underline{\small{0.049}\scriptsize{±0.004}} 
&\underline{\small{0.022}\scriptsize{±0.004}} 
&\underline{\small{0.039}\scriptsize{±0.003}} 
&\underline{\small{0.014}\scriptsize{±0.005}} &
    \\ 					
\texttt{\small{TableGAN}}       & 
&\textbf{\small{0.381}\scriptsize{±0.025}}
&\textbf{\small{0.043}\scriptsize{±0.007}}
&\textbf{\small{0.013}\scriptsize{±0.001}} 
&\textbf{\small{0.027}\scriptsize{±0.007}} 
&\textbf{\small{0.019}\scriptsize{±0.007}}
&\textbf{\small{0.020}\scriptsize{±0.007}} 
&\textbf{\small{0.011}\scriptsize{±0.005}} &
   \\ 
\texttt{\small{OCT-GAN}}        & 
&\small{241.104}\scriptsize{±4.724}
&\small{28.206}\scriptsize{±3.598}
&\small{10.312}\scriptsize{±3.688}
&\small{11.083}\scriptsize{±3.701}
&\small{8.622}\scriptsize{±3.704}
&\small{7.741}\scriptsize{±3.714}
&\small{4.224} \scriptsize{±3.680} 
&
 \\ 
\texttt{\small{RNODE}}       & 					
&\small{10.613}\scriptsize{±0.139}
&\small{12.901}\scriptsize{±0.731}
&\small{4.346}\scriptsize{±0.073}
&\small{2.576}\scriptsize{±0.622}
&\small{11.385}\scriptsize{±0.067}
&\small{16.461}\scriptsize{±1.326}
&\small{21.270}\scriptsize{±0.121}
&
   \\ 
\midrule
\texttt{\small{Na\"ive-STaSy}}       & 				
&\small{79.202}\scriptsize{±0.350}
&\small{7.192}\scriptsize{±0.320}
&\small{3.275}\scriptsize{±0.245}
&\small{3.466}\scriptsize{±0.254}
&\small{0.733}\scriptsize{±0.239}
&\small{3.774}\scriptsize{±0.327}
&\small{2.234}\scriptsize{±0.229}
&
    \\ 
\texttt{\small{STaSy}}  & 
&\small{79.085}\scriptsize{±0.449}
&\small{9.073}\scriptsize{±0.319}
&\small{11.903}\scriptsize{±0.201}
&\small{3.788}\scriptsize{±0.284}
&\small{3.287}\scriptsize{±0.267}
&\small{4.015}\scriptsize{±0.285}
&\small{2.173}\scriptsize{±0.225}&
   \\ 						

\bottomrule
\end{tabular}}
\end{table}

\begin{table}[h]
\caption{Wall-clock runtime for multi-class classification datasets}
\centering
\small
\setlength\tabcolsep{1.5pt}
\label{tab:runtime_multi}
\resizebox{1\textwidth}{!}{
\begin{tabular*}{1\textwidth}{lccccccccccc}
\toprule
\multirow{2}{*}{\small{Methods}}  &  & \multicolumn{6}{c}{\small{Multi-class classification}}   &\\ \cmidrule{3-8}
&  & \texttt{\small{Bean}}  & \texttt{\small{Contraceptive}} & \texttt{\small{Crowdsource}}       &   \texttt{\small{Obesity}}  &  \texttt{\small{Robot}}  &  \small{\texttt{Shuttle}}  & \\ 
\midrule
\texttt{\small{MedGAN}}      &
&\small{0.218}\scriptsize{±0.347}
&\small{0.174}\scriptsize{±0.335}
&\small{0.191}\scriptsize{±0.336}
&\small{0.196}\scriptsize{±0.334}	
&\small{0.181}\scriptsize{±0.338}
&\small{0.271}\scriptsize{±0.330}
&
    \\ 
\texttt{\small{VEEGAN}}  & 
&\small{0.029}\scriptsize{±0.006}
&\small{0.006}\scriptsize{±0.000}
&\underline{\small{0.033}\scriptsize{±0.006}}
&\underline{\small{0.009}\scriptsize{±0.001}}
&\small{0.024}\scriptsize{±0.010}
&\underline{\small{0.102}\scriptsize{±0.004}}	
&

 \\ 

\texttt{\small{CTGAN}}       & & 
\small{0.227}\scriptsize{±0.009}&	
\small{0.025}\scriptsize{±0.002}&	
\small{0.252}\scriptsize{±0.027}&	
\small{0.036}\scriptsize{±0.002}&	
\small{0.113}\scriptsize{±0.014}&	
\small{0.712}\scriptsize{±0.009}	&
  \\ 
\texttt{\small{TVAE}}        & &
\textbf{\small{0.023}\scriptsize{±0.002}}&	
\underline{\small{0.004}\scriptsize{±0.001}}&	
\small{0.038}\scriptsize{±0.005}&	
\small{0.039}\scriptsize{±0.040}&	
\underline{\small{0.009}\scriptsize{±0.000}}&	
\small{0.138}\scriptsize{±0.008}	&

    \\ 

\texttt{\small{TableGAN}}       & &
\underline{\small{0.026}\scriptsize{±0.008}}&	
\textbf{\small{0.002}\scriptsize{±0.000}}&	
\textbf{\small{0.021}\scriptsize{±0.007}}&	
\textbf{\small{0.004}\scriptsize{±0.001}}&	
\textbf{\small{0.006}\scriptsize{±0.000}}&	
\textbf{\small{0.041}\scriptsize{±0.007}}	&

   \\ 

\texttt{\small{OCT-GAN}}        && 
\small{6.595}\scriptsize{±3.725}&	
\small{2.849}\scriptsize{±3.678}&	
\small{6.917}\scriptsize{±3.743}&	
\small{2.885}\scriptsize{±3.613}&	
\small{5.013}\scriptsize{±3.670}&	
\small{29.469}\scriptsize{±3.677}&
 \\ 
\texttt{\small{RNODE}}       & & 
\small{5.210}\scriptsize{±0.234}&
\small{12.452}\scriptsize{±0.270}&
\small{7.985}\scriptsize{±0.072}&
\small{23.675}\scriptsize{±0.432}&
\small{10.861}\scriptsize{±0.801}&
\small{6.819}\scriptsize{±0.257}&
&					

   \\

\midrule
\texttt{\small{Na\"ive-STaSy}}       & & 
\small{2.797}\scriptsize{±0.263}&
\small{1.346}\scriptsize{±0.220}&
\small{3.090}\scriptsize{±0.237}&
\small{1.387}\scriptsize{±0.232}&
\small{1.692}\scriptsize{±0.215}&
\small{12.935}\scriptsize{±0.853}&
&					

    \\ 
\texttt{\small{STaSy}}  & &
\small{2.750}\scriptsize{±0.241}&
\small{1.045}\scriptsize{±0.247}&
\small{3.126}\scriptsize{±0.229}&
\small{1.023}\scriptsize{±0.237}&
\small{1.371}\scriptsize{±0.227}&
\small{13.352}\scriptsize{±0.382}&
   \\ 					
\bottomrule
\end{tabular*}}
\end{table}

\begin{table}[h]
\caption{Wall-clock runtime for regression datasets}
\centering
% \small
% \setlength\tabcolsep{1pt}
\label{tab:runtime_reg}
\begin{tabular}{lccccc}
\toprule
\multirow{2}{*}{{Methods}}  &  &  \multicolumn{2}{c}{{Regression}}  \\ \cmidrule{3-4}
  &&   \texttt{{Beijing}} &  \texttt{\small{News}} & \\ 
\midrule
\texttt{{MedGAN}}      	&&
0.212\footnotesize{±0.334}	&
0.284\footnotesize{±0.341}&
    \\ 
\texttt{{VEEGAN}} 	&&
0.067\footnotesize{±0.005}	&
0.234\footnotesize{±0.037}&

 \\
\texttt{{CTGAN}}      	&&
0.346\footnotesize{±0.023}	&
0.902\footnotesize{±0.013}&
  \\ 
\texttt{{TVAE}}       	&&
\underline{0.036\footnotesize{±0.006}}	&
\underline{0.075\footnotesize{±0.003}}&

    \\ 

\texttt{{TableGAN}}       & &
\textbf{0.023\footnotesize{±0.008}}	&
\textbf{0.052\footnotesize{±0.005}}&

   \\ 

\texttt{{OCT-GAN}}  &&	
18.008\footnotesize{±3.566}&	
20.866\footnotesize{±3.648}&
 \\ 
\texttt{{RNODE}}  &&
19.380\footnotesize{±0.500}&
34.955\footnotesize{±0.276}&
   \\ 
\midrule
\texttt{{Na\"ive-STaSy}}  &&
7.338\footnotesize{±0.146}&
2.362\footnotesize{±0.241}&
    \\ 
\texttt{{STaSy}}  &&
7.818\footnotesize{±1.074}&
16.132\footnotesize{±1.241}&
   \\ 
\bottomrule
\end{tabular}
\end{table}

% \clearpage
\section{Experimental environments}\label{sec:apd_environments}

Our software and hardware environments are as follows: \textsc{Ubuntu} 18.04 LTS, \textsc{Python} 3.8.2, \textsc{Pytorch} 1.8.1, \textsc{CUDA} 11.4, and \textsc{NVIDIA} Driver 470.42.01, i9 CPU, and \textsc{NVIDIA RTX 3090}. 
% \jayoung{The SGM backbone for all \texttt{STaSy} is ~\citep{songyang} \footnote{\url{https://github.com/yang-song/score_sde_pytorch}  (Apache License 2.0)}.}
Our code for the experiments is mainly based on \url{https://github.com/yang-song/score_sde_pytorch} (Apache License 2.0).
% The VAE backbone for all LSGM models is NVAE [20], one of the best-performing VAEs in the literature.

\subsection{Baselines}
We utilize a set of baselines that includes various generative models.
\begin{itemize}
    % \item \texttt{Identity} is a case where we do not synthesize but use original data. 
    \item \texttt{Identity} is a case where we do not synthesize but use original data.
    \item  \texttt{MedGAN}\footnote{\url{https://github.com/sdv-dev/SDGym} (MIT License)}~\citep{DBLP:journals/corr/ChoiBMDSS17} is a GAN that incorporates non-adversarial losses to generate discrete medical records. 
    \item \texttt{VEEGAN}\footnotemark[1]~\citep{srivastava2017veegan} is a GAN for tabular data that avoids mode collapse by adding a reconstructor network. 
    \item \texttt{CTGAN}\footnotemark[1]~\citep{NIPS2019_8953} and \texttt{TVAE}\footnotemark[1]~\citep{NIPS2019_8953} are a conditional GAN and a VAE for tabular data with mixed types of variables. 
    % \item \texttt{TVAE}~\citep{NIPS2019_8953} is a variational autoencoder that considers column-type-specific processes. 
    \item \texttt{TableGAN}\footnotemark[1]~\citep{park2018data} is a GAN for tabular data using convolutional neural networks. 
    \item \texttt{OCT-GAN}\footnote{\url{https://github.com/bigdyl-yonsei/OCTGAN}}~\citep{10.1145/3442381.3449999} is a GAN that has a generator and discriminator based on neural ordinary differential equations. 
    \item \texttt{RNODE}\footnote{\url{https://github.com/cfinlay/ffjord-rnode} (MIT License)}~\citep{Finlay2020HowTT} is an advanced flow-based model with two regularization terms added to the training objective of FFJORD~\citep{grathwohl2018ffjord}. 
    % \texttt{Na\"ive-STaSy} is a score-based generative model for image based on SDE. We adopt this model on tabular data and customize the score network for tabular data synthesis.
\end{itemize}

\subsection{Datasets}\label{sec:datasets}
In this section, we describe 15 real-world tabular datasets for our experiments. We select the datasets for experiments with two metrics: 1) how many times a dataset has been cited/used in previous papers, and 2) how many times a dataset has been viewed/downloaded in famous repositories, such as UCI Machine Learning Repository and Kaggle. Among them, we choose the datasets that can be used for classification and regression tasks, with more than 5 columns and 1,000 rows. 

\begin{itemize}
    % \item \texttt{Cardio} is a binary classification dataset on the records of patients with cardiovascular disease.
    \item \texttt{Credit} is a binary classification dataset collected from European cardholders for credit card fraud detection. 
    \item \texttt{Default}~\citep{default} is a binary classification dataset describing the information on credit card clients in Taiwan regarding default payments. 
    \item \texttt{HTRU}~\citep{misc_htru2_372} is a binary classification dataset that describes a sample of pulsar candidates collected during the High Time Resolution Universe Survey.
    \item \texttt{Magic}~\citep{misc_magic_gamma_telescope_159} is a binary classification dataset that simulates the registration of high-energy gamma particles in the atmospheric telescope.
    \item \texttt{Phishing}~\citep{misc_phishing_websites_327} is a binary classification dataset used to distinguish between phishing and legitimate web pages.
    \item \texttt{Shoppers}~\citep{shoppers} is a binary classification dataset about online shoppers' intention. 
    \item \texttt{Spambase}~\citep{misc_spambase_94} is a binary classification dataset that indicates whether an email is spam or non-spam.
    \item \texttt{Bean}~\citep{bean} is a multi-class classification dataset that includes types of beans with their characteristics. 
    \item \texttt{Contraceptive}~\citep{contraceptive} is a multi-class classification dataset about Indonesia contraceptive prevalence.
    \item \texttt{Crowdsource}~\citep{crowdsource} is a multi-class classification dataset used to classify satellite images into different land cover classes. 
    \item \texttt{Obesity}~\citep{obesity} is a multi-class classification dataset describing obesity levels based on eating habits and physical condition. 
    \item \texttt{Robot}~\citep{robot} is a multi-class classification dataset collected as the robot moves around the room, following the wall using ultrasound sensors.
    \item \texttt{Shuttle}~\citep{shuttle} is a multi-class classification dataset for extracting conditions in which automatic landing is preferred over manual control of the spacecraft.
    \item \texttt{Beijing}~\citep{beijing_pm2.5} is a regression dataset about PM2.5 air quality in the city of Beijing. 
    \item \texttt{News}~\citep{news} is a regression dataset about online news articles to predict the number of shares in social networks. 
    % \item \texttt{RiceType} is a binary classification dataset about many features extracted from two kinds of rice.
    % \item \texttt{Musk}~\citep{misc_musk_(version_2)_75} is a binary classification dataset that describes a set of molecules depending on their features.
\end{itemize}

The statistical information of datasets used in our experiments is in Table~\ref{tab:dataset}. \#train, \#test, \#continuous, \#categorical, and \#class mean the number of training data, testing data, continuous columns, categorical columns, and class, respectively. 
% We randomly split 80\% of the datasets into a training split and use the remaining 20\% as a test split.

\begin{table}[t]
% \vspace{-1em}
\caption{Datasets used for our experiments}
\label{tab:dataset}
\centering
\setlength\tabcolsep{3pt}
\begin{tabular}{lcccccc}
\toprule
Datasets & \#train & \#test & \#continuous & \#categorical & task (\#class) \\
\midrule
% \texttt{Cardio}& 56K & 14K & 5 & 7 & Binary classification & \\
\texttt{Credit}& 264.8K&20K&29&1&Binary classification&\\
\texttt{Default} & 24K&6K&13&11&Binary classification&\\
\texttt{HTRU}& 14.3K & 3.6K & 8 & 1 & Binary classification & \\
\texttt{Magic} & 15.2K & 3.8K & 10 & 1 & Binary classification & \\
\texttt{Phishing}& 8.8K & 2.2K & 0 & 31 & Binary classification & \\
\texttt{Shoppers} & 9.8K&2.4K&10&8&Binary classification& \\
\texttt{Spambase}& 3.7K & 0.9K & 57 & 1 & Binary classification & \\
% \texttt{Musk}& 5.3K & 1.3K & 165 & 1 & Binary classification & \\
\texttt{Bean} & 10.8K&2.7K&16&1&Multi-class classification (7)& \\
\texttt{Contraceptive}& 1.2K & 0.3K & 0 & 10 & Multi-class classification (3) & \\
\texttt{Crowdsource} & 8.6K&2.1K&28&1&Multi-class classification (6)&\\
\texttt{Obesity} & 1.6K&0.4K&7&10&Multi-class classification (7)&\\
\texttt{Robot}& 4.4K & 1.1K & 24 & 1 & Multi-class classification (4) & \\
\texttt{Shuttle}& 46.4K & 11.6K & 9 & 1 & Multi-class classification (7) & \\
\texttt{Beijing} & 15.2K&3.8K&8&6&Regression& \\
\texttt{News} & 31.6K&8K&45&14&Regression&\\
% \texttt{RiceType} & 14.5K&3.6K&10&1&Binary classification&\\
\bottomrule
\end{tabular}
% \vspace{-1em}
\end{table}

The raw data of 15 datasets are available online:
\begin{itemize}
    % \item \texttt{Cardio}: \url{https://www.kaggle.com/sulianova/cardiovascular-disease-dataset}
    \item \texttt{Credit}: \url{https://www.kaggle.com/mlg-ulb/creditcardfraud} (DbCL 1.0)
    \item \texttt{Default}: \url{https://archive.ics.uci.edu/ml/datasets/default+of+credit+card+clients} (CC BY 4.0)
    \item \texttt{HTRU}: \url{https://archive.ics.uci.edu/ml/datasets/HTRU2} (CC BY 4.0)
    \item \texttt{Magic}: \url{https://archive.ics.uci.edu/ml/datasets/magic+gamma+telescope} (CC BY 4.0)
    \item \texttt{Phishing}: \url{https://archive.ics.uci.edu/ml/datasets/phishing+websites} (CC BY 4.0)
    \item \texttt{Shoppers}: \url{https://archive.ics.uci.edu/ml/datasets/Online+Shoppers+Purchasing+Intention+Dataset} (CC BY 4.0)
    \item \texttt{Spambase}: \url{https://archive.ics.uci.edu/ml/datasets/spambase} (CC BY 4.0)
    \item \texttt{Bean}: \url{https://archive.ics.uci.edu/ml/datasets/Dry+Bean+Dataset} (CC BY 4.0)
    \item \texttt{Contraceptive}: \url{https://archive.ics.uci.edu/ml/datasets/Contraceptive+Method+Choice} (CC BY 4.0)
    \item \texttt{Crowdsource}: \url{https://archive.ics.uci.edu/ml/datasets/Crowdsourced+Mapping#} (CC BY 4.0)
    \item \texttt{Obesity}: \url{https://archive.ics.uci.edu/ml/datasets/Estimation+of+obesity+levels+based+on+eating+habits+and+physical+condition+} (CC BY 4.0)
    \item \texttt{Robot}: \url{https://archive.ics.uci.edu/ml/datasets/Wall-Following+Robot+Navigation+Data} (CC BY 4.0)
    \item \texttt{Shuttle}: \url{https://archive.ics.uci.edu/ml/datasets/Statlog+(Shuttle)} (CC BY 4.0)
    \item \texttt{Beijing}: \url{https://archive.ics.uci.edu/ml/datasets/Beijing+PM2.5+Data} (CC BY 4.0)
    \item \texttt{News}: \url{https://archive.ics.uci.edu/ml/datasets/online+news+popularity} (CC BY 4.0)
    % \item \texttt{RiceType}: \url{https://www.kaggle.com/datasets/mssmartypants/rice-type-classification} (GPL 2.0)
    % \item \texttt{Musk}: \url{https://archive.ics.uci.edu/ml/datasets/Musk+(Version+2)} (CC BY 4.0)
\end{itemize}

% \subsection{Evaluation methodology}\label{sec:methodology}
%  We randomly split 80\% of the dataset into a training split and use the remaining 20\% as a test split, which is used for evaluation.
% % We separate the datasets into training/testing sets and use them for evaluation.
% We first train generative models with the training sets. After that, we i) generate a fake table, ii) train \texttt{Decision tree}~\citep{breiman1984classification}, \texttt{Logistic regession}~\citep{cox1958regression}, \texttt{Adaboost}~\citep{10.5555/1624312.1624417}, and \texttt{MLP}~\citep{10.5555/1162264} with the fake table for the binary classification, \texttt{Decision tree} and \texttt{MLP} for the multi-class classification, and \texttt{Linear regression} and \texttt{MLP} for the regression tasks, and iii) evaluate them with testing sets. We use F1 (resp. Macro F1) score for the binary (resp. the multi-class) classification and $R^2$ score for the regression tasks. This evaluation framework is 

% \subsection{Processing of tabular data}\label{sec:processing}
% To handle mixed types of data, which is a challenge in tabular data generation, we pre/post-process columns. We use the min-max scaler to pre-process numerical columns, and its reverse scaler is used for post-processing after generation. We also apply one-hot encoding to pre-process categorical columns, and use the softmax function, followed by the rounding function, when generating.

\subsection{Evaluation methods}\label{sec:evaluation}
The reported scores for TSTR results in the paper are calculated as follows:
\begin{enumerate}
    \item We download a dataset. If used previously, we use their train-test split. If not used before, we perform a new train-test split. The train-test split ratio is 80\% and 20\%, respectively.
    \item Generate fake records which has the same number of records as the original training set for other fake data generation methods.
    \item Using the training records from Step2, we train base classifiers/regressors to predict. We search the best hyperparameter set for each classifier/regressor. In Table~\ref{tab:binary_tasks}, considered hyperparameters and their candidate settings are summarized. We use DecisionTree, AdaBoost, Logistic Regression, MLP classifiers, RamdomForest, and XGBoost for binary classification tasks; DecisionTree, MLP classifiers, RandomForest, and XGBoost for multi-class classification tasks; MLP regressor, RandomForest, XGBoost, and Linear Regression for regression tasks. % Hyperparameter settings for the classifiers and regressors are in Table~\ref{tab:binary_tasks}, and we search for the best classifiers/regressors.}
    \item Test the classifiers/regressors with a testing data. We use various evaluation metrics for rigorous evaluations as reported earlier.
    % \chaejeong{
    % \item Repeat Step2 to Step5 5 times and average them.
    % \item Take an average score of the classifiers/regressors.}
\end{enumerate}

We repeat Step2 to Step4 5 times for all datasets. We then calculate the average score for each method and for each evaluation metric. Detailed metrics for our experiment are as follows:
\begin{enumerate}
    \item Binary F1 for binary classification datasets: f1\_score from sklearn.metrics after setting the ‘average’ option to ‘binary’.
    \item Macro F1 for multi-class classification datasets: f1\_score from sklearn.metrics after setting the ‘average’ option to ‘macro’.
    \item Weighted-F1 for classification datasets: Weighted-F1 = $\sum_{i=0}^{N} w_i s_i$, where $N$ is the number of classes, the weight of $i$-th class $w_i$ is $ \frac{1-p_i}{N-1} $, $p_i$ is the proportion of $i$-th class’s cardinality in a total dataset, and score $s_i$ is a per-class F1 of $i$-th class (in a One-vs-Rest manner). This formula allows us to evaluate synthesized tables with more focus on mode collapse by giving a higher weight to a smaller class, which is more likely to be forgotten by the model. 
    \item AUROC: roc\_auc\_score from sklearn.metrics.
    \item Coverage: compute\_prdc from \url{https://github.com/clovaai/generative-evaluation-prdc}.
\end{enumerate}

\begin{table}[t]
\caption{Hyperparameters of the base classifiers/regressors}
\label{tab:binary_tasks}
\centering
% \begin{tabular}{llcccc}
\begin{tabular}{lllll}
\toprule
Models & Hyperparameters & Values & \\
\midrule
\multirow{3}{*}{DecisionTree} & max\_depth & 4, 8, 16, 32&\\
 & min\_samples\_split& 2, 4, 8&\\
 & min\_samples\_leaf & 1, 3, 5&\\
\midrule
AdaBoost & n\_esimators & -10, 50, 100 & \\
\midrule
\multirow{5}{*}{Logistic Regression} & solver & lbfgs  & \\
& n\_jobs & -1 & \\
 & max\_iter & 10, 50, 100& \\
& C & 0.01, 0.1, 1.0 & \\
& tol & 0.0001, 0.01, 0.1&  \\
\midrule
 \multirow{3}{*}{MLP} & hidden\_layer\_sizes  & (100, ), (200, ), (100, 100)  & \\
 & max\_iter & 50, 100 & \\
 & alpha & 0.0001, 0.001 & \\
\midrule
\multirow{4}{*}{RandomForest} & max\_depth & 8, 16, Inf  & \\
 & min\_samples\_split & 2, 4 & \\
 & min\_samples\_leaf & 1, 3& \\
 & n\_jobs & -1 & \\
\midrule
\multirow{5}{*}{XGBoost} & n\_estimators & 10, 50, 100  & \\
 &min\_child\_weight & 1, 10 & \\
 & max\_depth & 5, 10& \\
 & gamma & 0.0, 1.0 & \\
% & & objective & binary:logistic & \\
 & nthread & -1 & \\
 \midrule
 Linear Regression & - & - & \\
\bottomrule
\end{tabular}
\end{table}

% \begin{table}[h]
% \caption{\textcolor{red}{Hyperparameters for the models used in multi-class classification tasks.}}
% \label{tab:multi_tasks}
% \centering
% \begin{tabular}{lcccc}
% \toprule
% Models & Hyperparameters & Values & \\
% \midrule
% DecisionTree & max\_depth & 30&\\
% MLP Classifier& hidden\_layer / max\_iter & (100, ) / 50 & \\
% \bottomrule
% \end{tabular}
% \end{table}

% \begin{table}[h]
% \caption{\textcolor{red}{Hyperparameters for the models used in regression tasks.}}
% \label{tab:reg_tasks}
% \centering
% \begin{tabular}{lcccc}
% \toprule
% Models & Hyperparameters & Values & \\
% \midrule
% Linear Regression&-& - &\\
% MLP Regressor & hidden\_layer / max\_iter & (100, ) / 50 & \\
% \bottomrule
% \end{tabular}
% \end{table}

\section{Hyperparameters}\label{sec:hyperparameters}
% In this section, we report the hyperparameters used for our experiments.

% \begin{table}[h]
% \vspace{-1em}
% \caption{The best hyperparameters of \jayoung{SGM with SPL} used in Table~\ref{tbl:main}.}
% \label{tab:hyperparameters}
% \centering
% % \setlength\tabcolsep{1.5pt}
% \begin{tabular}{lcccccccc}
% \toprule
% \multirow{2}{*}{Dataset} &  \multicolumn{6}{c}{Hyperparameters for \jayoung{SGM with SPL}} & \\ \cmidrule{2-7}
% &SDE Type & Layer Type & Activation & Lean. Rate &$\alpha_0$ &$\beta_0$ & \\
% \midrule
% \texttt{Default} &VP&concatsquash&ReLU &2e-03&0.30&0.90& \\
% \texttt{Shoppers} &VE&concatsquash&ELU &2e-03&0.30&0.95& \\
% \texttt{Credit} &VP&concatsquash&LeakyReLU &2e-03&0.25&0.90& \\
% \texttt{Bean} &VP&squash&ReLU &2e-03&0.25&0.80& \\
% \texttt{Crowdsource} &VP&concat&ReLU &2e-04&0.25&0.95&\\
% \texttt{Obesity} &VP&squash&ELU &2e-03&0.25&0.90& \\
% \texttt{Beijing} &VE&concatsquash&LeakyReLU &2e-03&0.25&0.90& \\
% \texttt{News} &sub-VP&concatsquash&LeakyReLU &2e-03&0.30&0.95& \\

% \bottomrule
% \end{tabular}
% \end{table}
% % \end{center}

\begin{table}[t]
% \vspace{-0.5em}
\caption{The best hyperparameters used in Table~\ref{tbl:main}}
\label{tab:hyperparameters}
% \small
\centering
\setlength\tabcolsep{1.5pt}
\resizebox{1\textwidth}{!}{
\begin{tabular}{lccccccccccc}
\toprule
\multirow{2}{*}{Datasets} &  \multicolumn{6}{c}{Hyperparameters for SPL of \texttt{STaSy}} &&\multicolumn{2}{c}{Hyperparameters for fine-tuning} \\ \cmidrule{2-7} \cmidrule{9-10}
&SDE Type & Layer Type & Activation & Learn. Rate &$\alpha_0$ &$\beta_0$ && Hutchinson Type & Learn. Rate & \\
\midrule
% \texttt{Cardio} & & &  & & & &&   & &\\
\texttt{Credit} &VP&concatsquash&LeakyReLU &2e-03&0.25&0.90&& Rademacher &2e-05& \\
\texttt{Default} &VP&concatsquash&ReLU &2e-03&0.30&0.90&&  Rademacher&2e-04&\\
\texttt{HTRU} & VE & concatsquash& LeakyReLU & 2e-04 & 0.25 & 0.95&& Rademacher  & 2e-05 &\\
\texttt{Magic} & sub-VP& squash& ReLU & 2e-03& 0.3& 0.95&& Rademacher  &2e-07 &\\
\texttt{Phishing} & VP& squash& LeakyReLU & 2e-03& 0.2& 0.95&& Rademacher  &2e-07 &\\
\texttt{Shoppers} &VE&concatsquash&ELU &2e-03&0.30&0.95&& Rademacher&2e-07&\\
\texttt{Spambase} &sub-VP & concat& LeakyReLU &2e-04 &0.2 &0.95 &&  Rademacher & 2e-07&\\
\midrule
\texttt{Bean} &VP&squash&ReLU &2e-03&0.25&0.80&& Rademacher &2e-05& \\
\texttt{Contraceptive} & VP & concatsquash& LeakyReLU & 2e-03& 0.2&0.95 && Rademacher  &2e-07 &\\
\texttt{Crowdsource} &VE&squash&LeakyReLU &2e-03&0.25&0.90&&Gaussian&2e-07&\\
\texttt{Obesity} &VP&squash&ELU &2e-03&0.25&0.90& & Gaussian&2e-05&\\
\texttt{Robot} & sub-VP & concat& ELU & 2e-03& 0.3& 0.95&&   Rademacher& 2e-07&\\
\texttt{Shuttle} & sub-VP& squash& ReLU &2e-03 & 0.3& 0.95&& Rademacher  &2e-07 &\\
\midrule
\texttt{Beijing} &VE&concatsquash&LeakyReLU &2e-03&0.25&0.90&& Rademacher&2e-07&  \\
\texttt{News} &VE&concatsquash&LeakyReLU &2e-03&0.25&0.90& &Gaussian&2e-07& \\
\bottomrule
\end{tabular}
}
% \vspace{-0.5em}
\end{table}
Hyperparameter settings for the best models are in Table~\ref{tab:hyperparameters}. We have three SDE types, which are VE, VP, and sub-VP, and three layer types as shown in Appendix~\ref{networkarchi}: Concat, Squash, and Concatsquash. We use a learning rate in $\{2e-03, 2e-04\}$. We search for $\alpha_0$ and $\beta_0$, in total, with 9 combinations using $\alpha_0 = \{0.20, 0.25, 0.30\}$ and $\beta_0 = \{0.80, 0.90, 0.95\}$.

We also consider the hyperparameters for the fine-tuning process. To compute the exact log-probability with the Hutchinson's estimation~\citep{hutchinson, grathwohl2018ffjord}, we use a Gaussian or a Rademacher distribution for $p(\boldsymbol{\epsilon})$, where $\boldsymbol{\epsilon}$ is a noise vector. The fine-tuning epoch is $\{1, \dots, 20\}$, and the fine-tuning learning rate is $\{2\times 10^{-i} | i=\{4,5,6,7\}\}$.

\clearpage
\section{Additional visualizations}\label{sec:additional_visualization}
% \subsection{A comparison between \texttt{STaSy} and others}
% \jayoung{Add more figures like fig2. (right).}
We show several visualizations that are missing in the main paper. In each subsection, we show column-wise histograms and t-SNE visualizations on \texttt{HTRU}, \texttt{Robot}, and \texttt{News}, respectively. 
% We omit visualizations of \texttt{MedGAN} and \texttt{VEEGAN} due to their poor generation quality.

% \chaejeong{In Fig.~\ref{fig:histo_shoppers}, the fake data by \texttt{TableGAN} and \texttt{STaSy} show almost similar distributions to that of real data, and so do \texttt{TVAE} and \texttt{STaSy} in Fig.~\ref{fig:histo_obesity}, and \texttt{STaSy} in Fig.~\ref{fig:histo_news}. On the other hand, other baselines fail to find the distribution of real. In Figs.~\ref{fig:tsne_shoppers}, ~\ref{fig:tsne_obesity}, and ~\ref{fig:tsne_news}, almost all baselines generate out-of-distribution records or have mode collapse, highlighted in red, but \texttt{STaSy} generates fake data that is nearly identical to the real distribution. Especially, they show that our score-based ablation models have reliable sampling diversity and SPL can improve it in \texttt{Shoppers}.}

\subsection{Additional visualizations in \texttt{HTRU}}

\begin{figure}[h]
        \centering
        \begin{subfigure}{\includegraphics[width=0.23\textwidth]{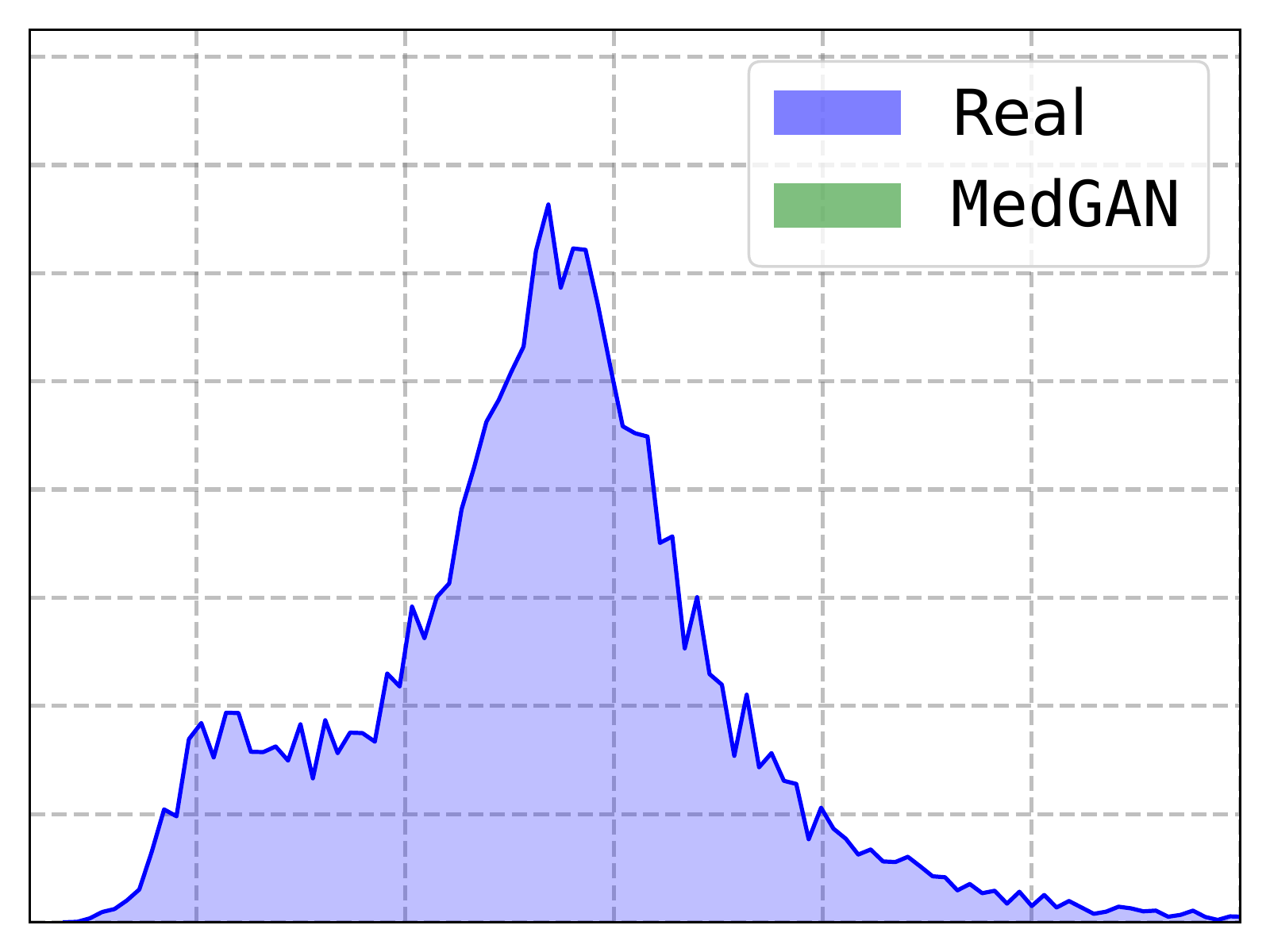}}
        \end{subfigure}
        \begin{subfigure}{\includegraphics[width=0.23\textwidth]{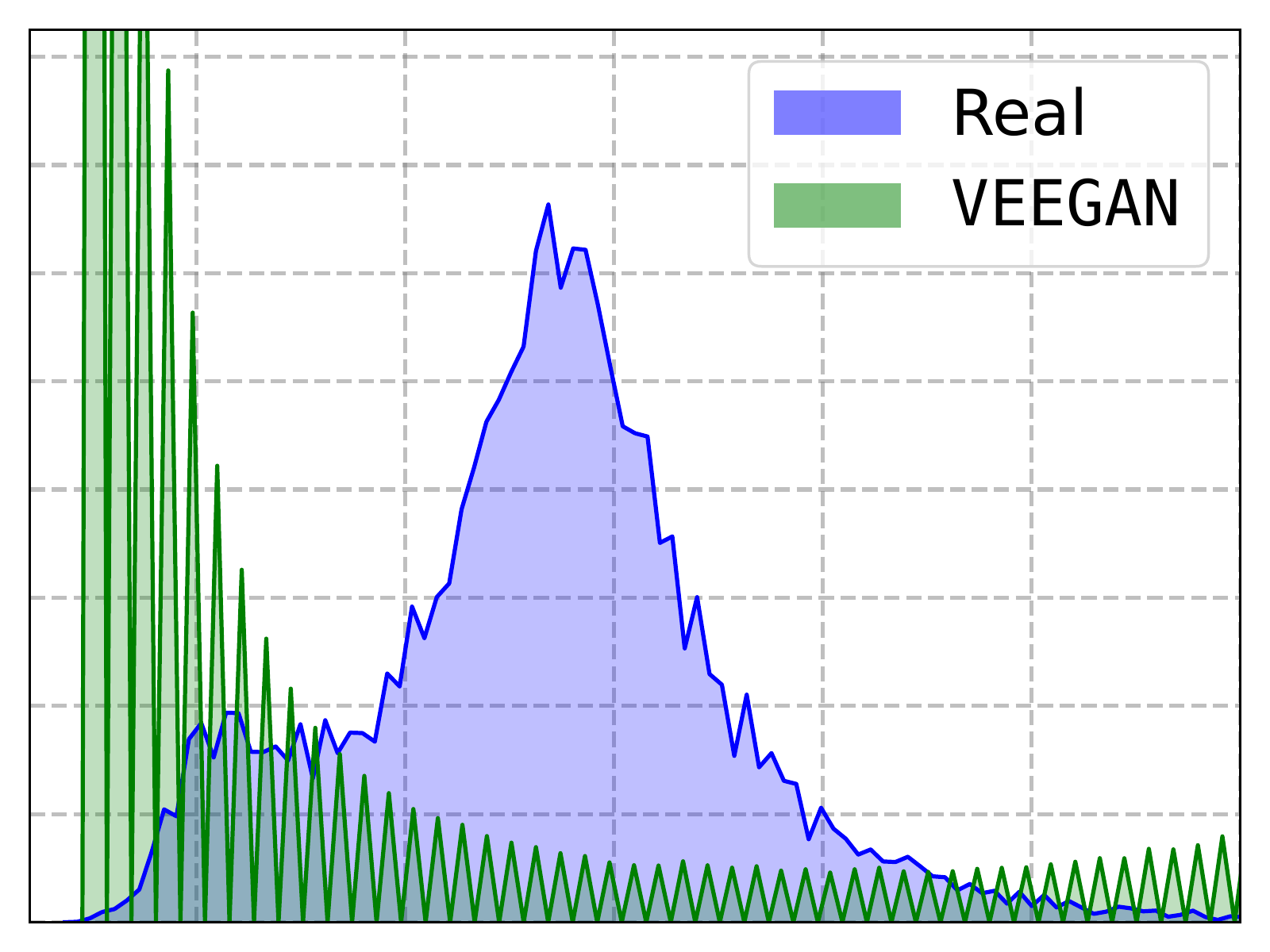}}
        \end{subfigure}
        \begin{subfigure}{\includegraphics[width=0.23\textwidth]{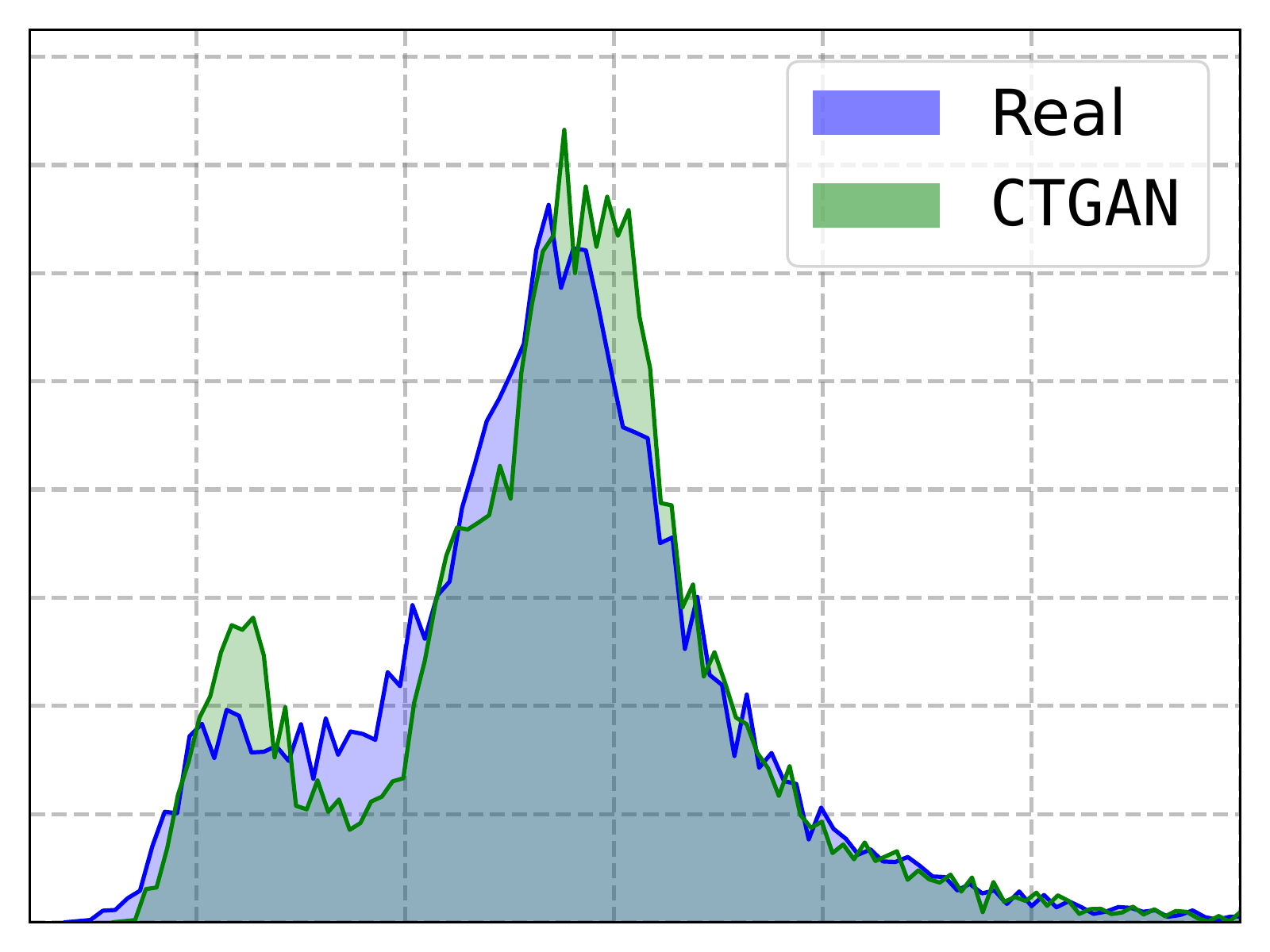}}
        \end{subfigure}
        \begin{subfigure}{\includegraphics[width=0.23\textwidth]{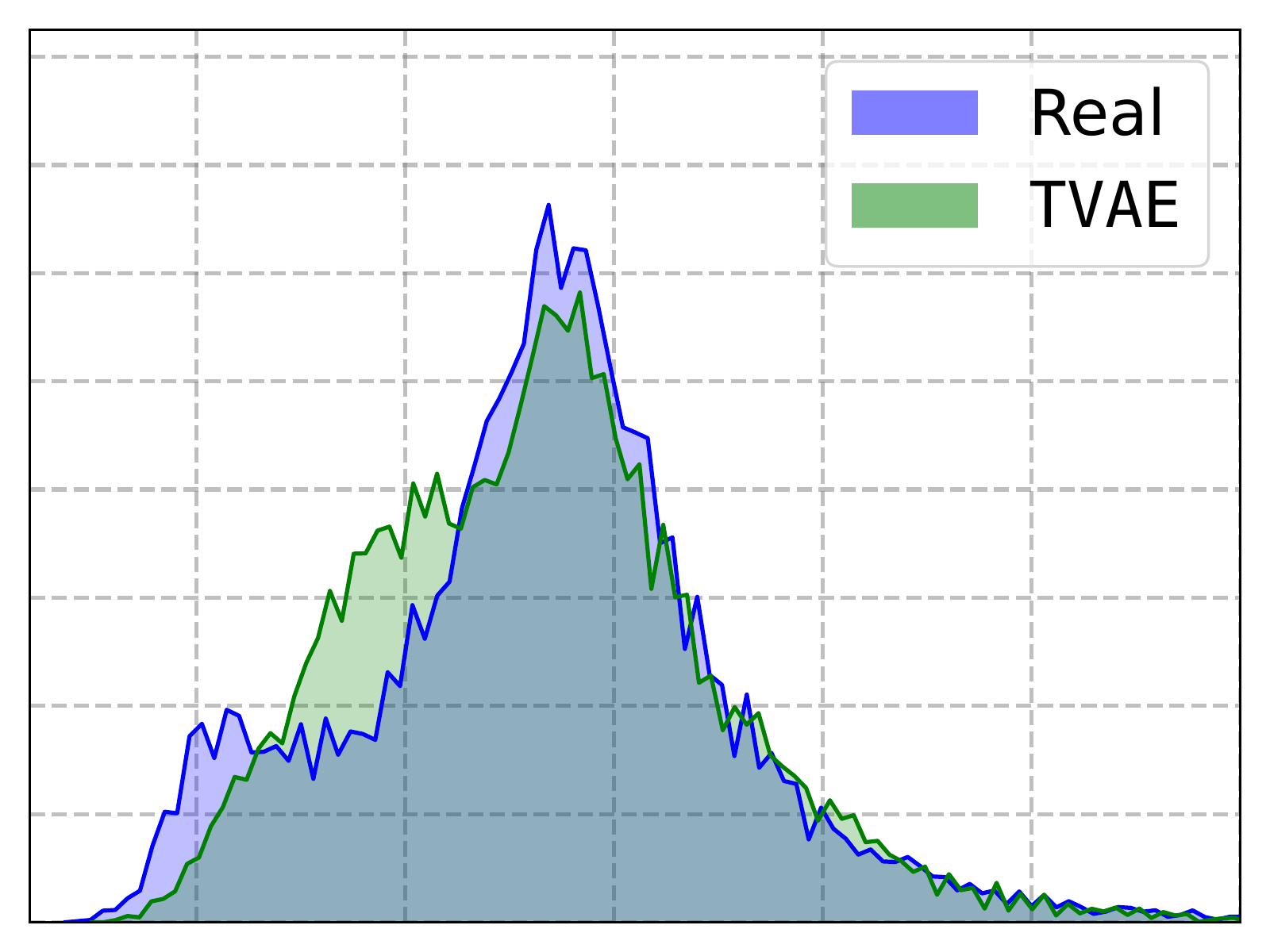}}
        \end{subfigure}
        \begin{subfigure}{\includegraphics[width=0.23\textwidth]{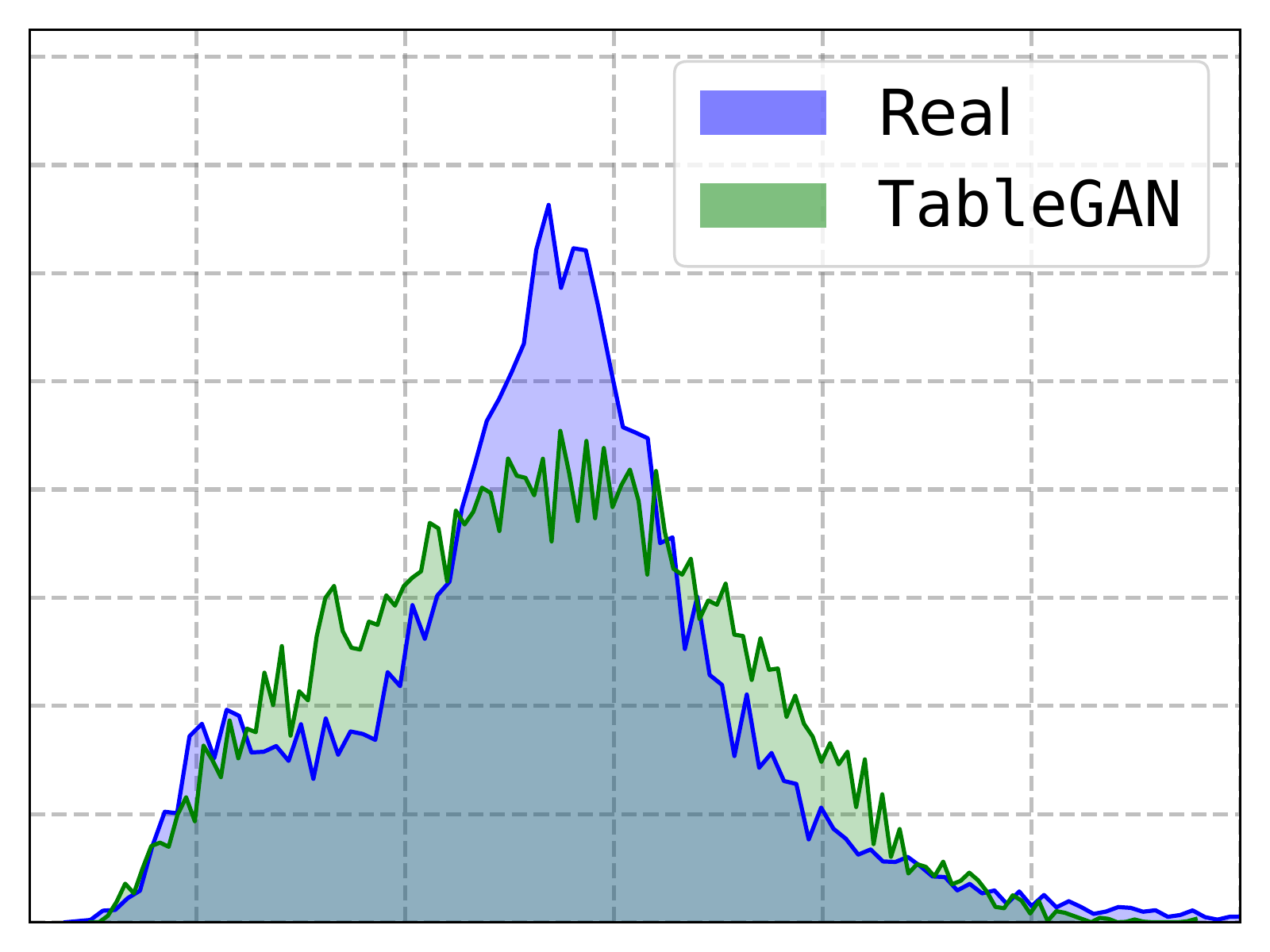}}
        \end{subfigure}
        % \\[\baselineskip]
        % \vspace{-1em}
        \begin{subfigure}{\includegraphics[width=0.23\textwidth]{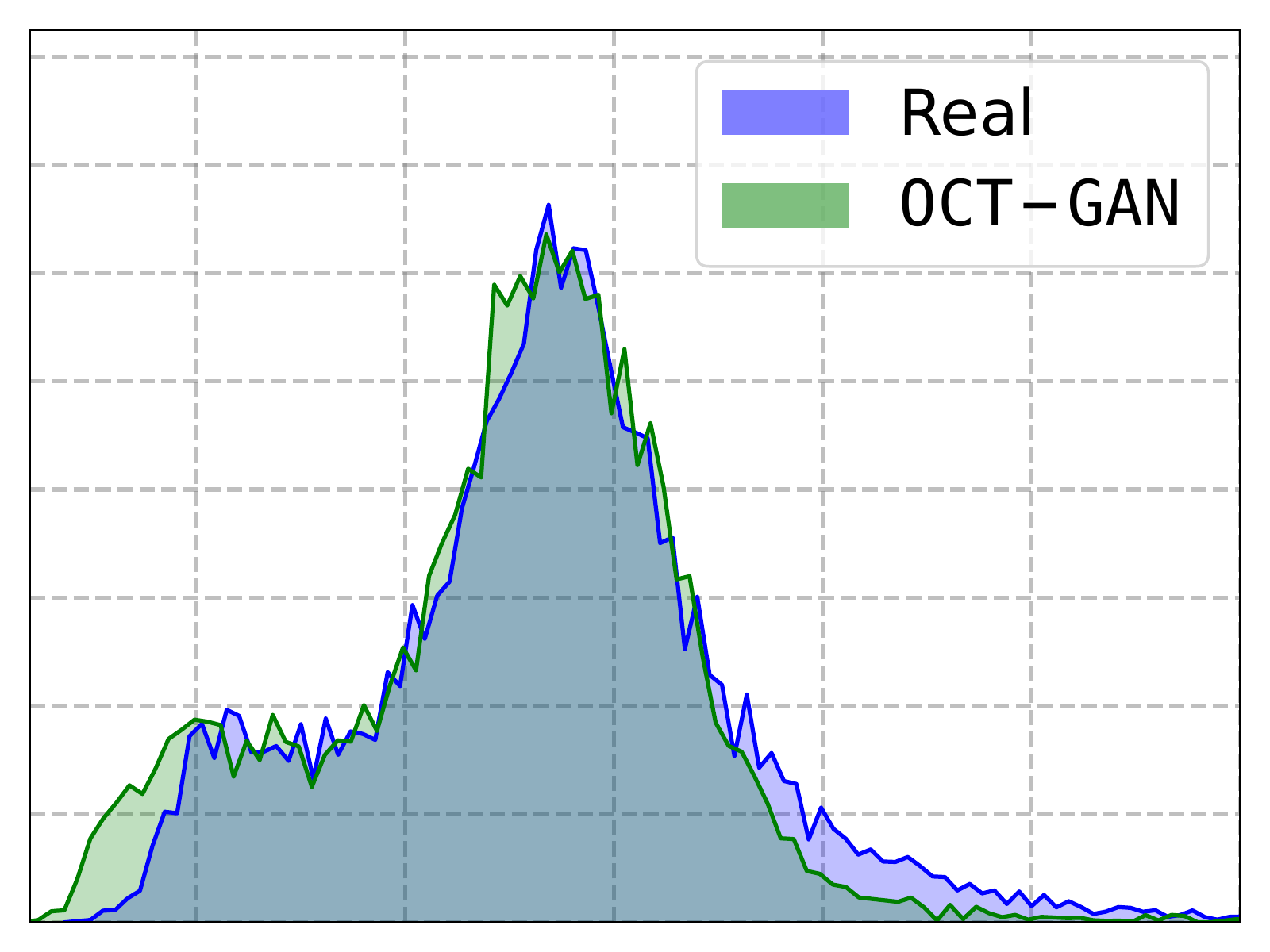}}
        \end{subfigure}
        \begin{subfigure}{\includegraphics[width=0.23\textwidth]{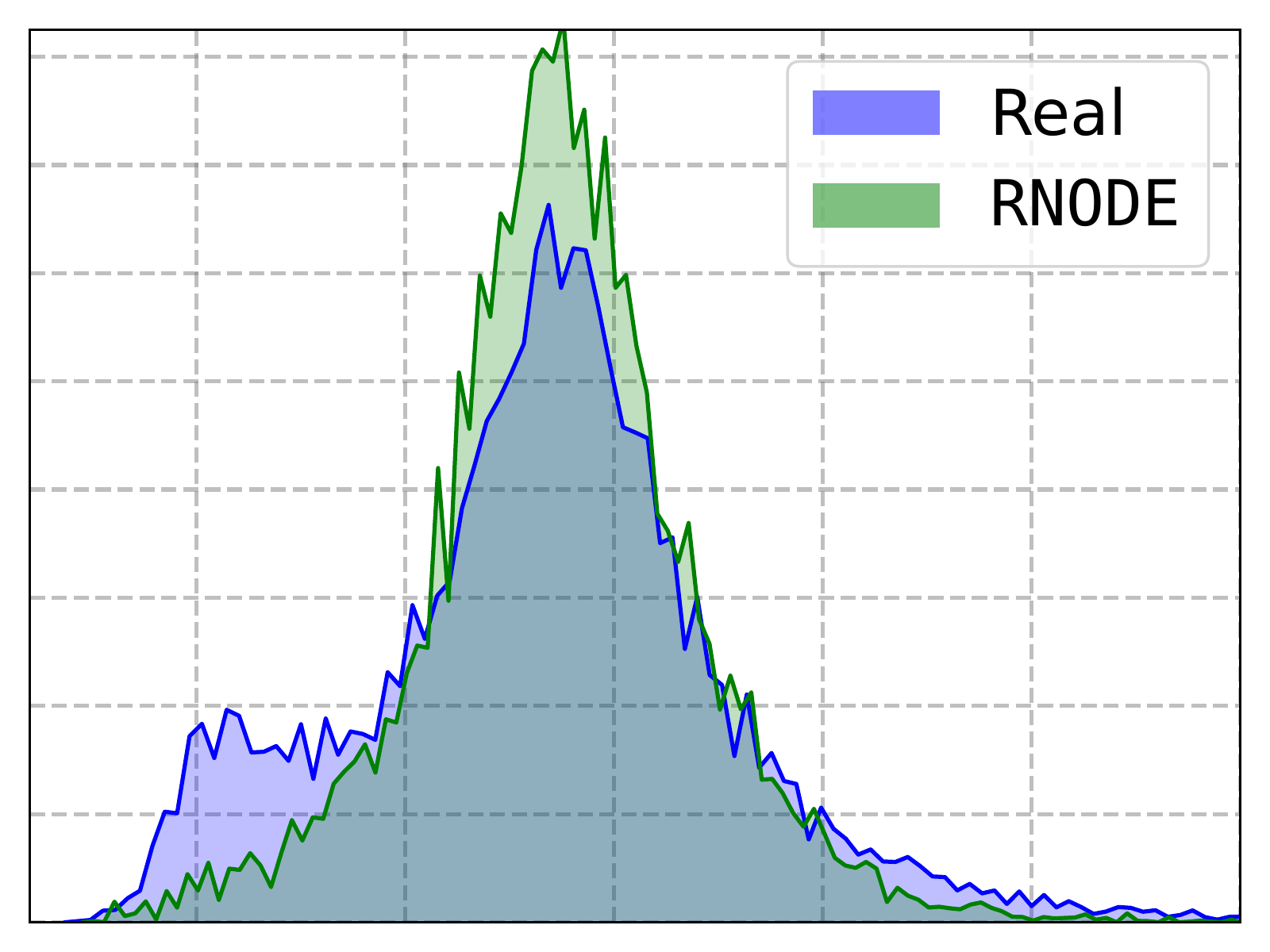}}
        \end{subfigure}
        \begin{subfigure}{\includegraphics[width=0.23\textwidth]{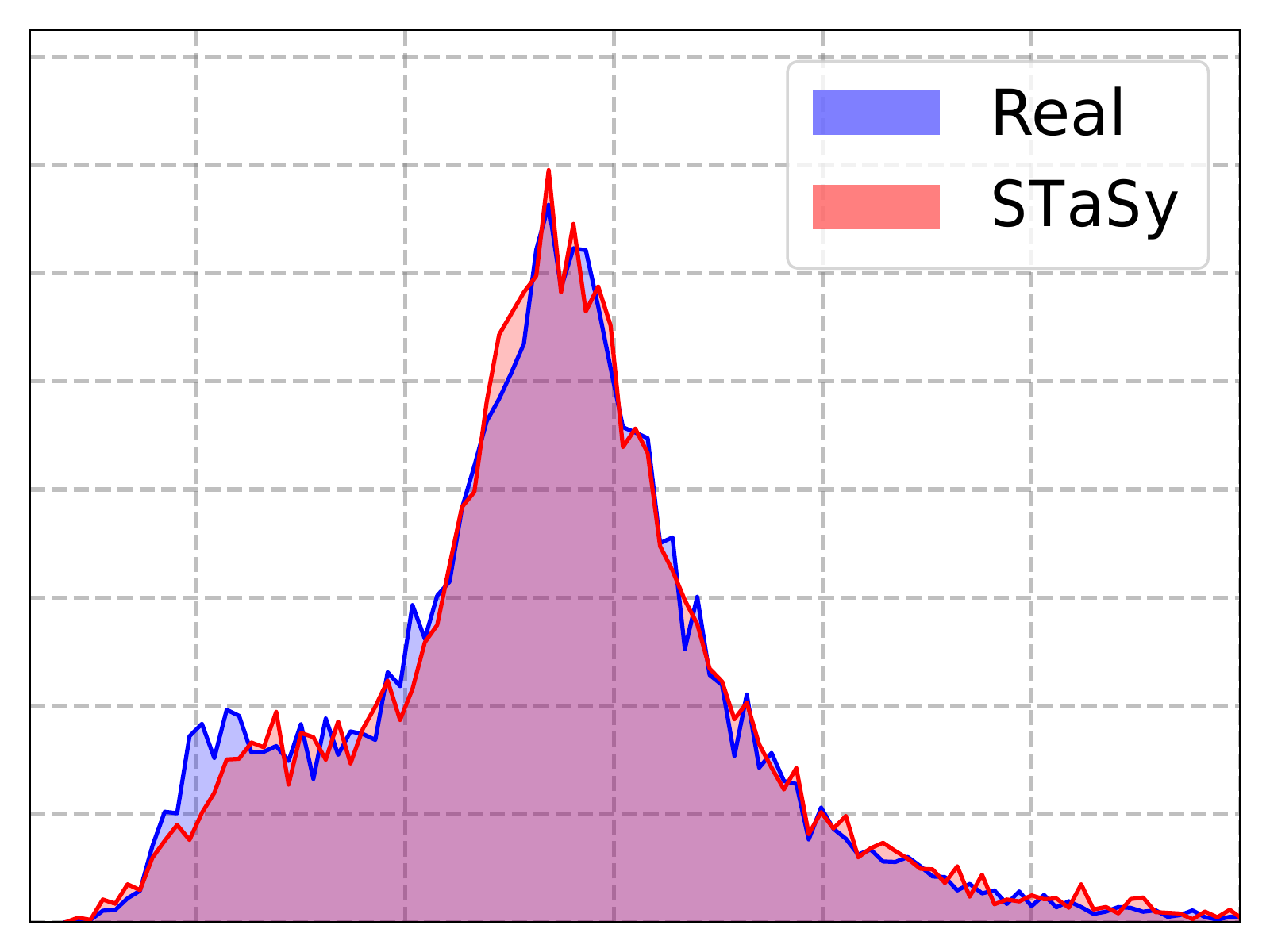}}
        \end{subfigure}
    \caption{Histograms of values in the \textit{excess kurtosis} column of \texttt{HTRU}}
    \label{fig:histo_htru}
% \vspace{-2em}
\end{figure}

\begin{figure}[h]
% \begin{subfigure}
        \centering
        \begin{subfigure}{\includegraphics[width=0.23\textwidth]{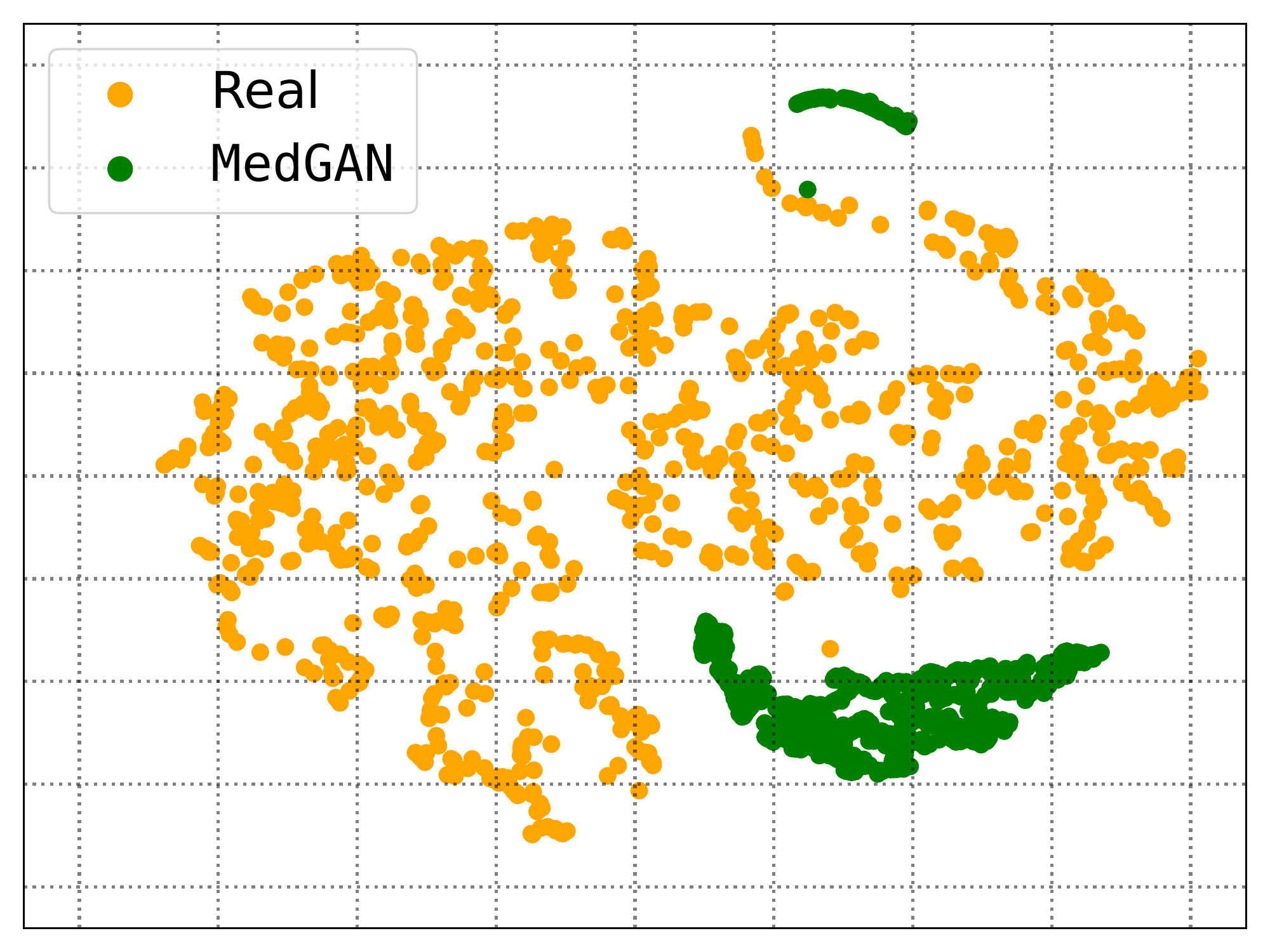}}
        \end{subfigure}
        \begin{subfigure}{\includegraphics[width=0.23\textwidth]{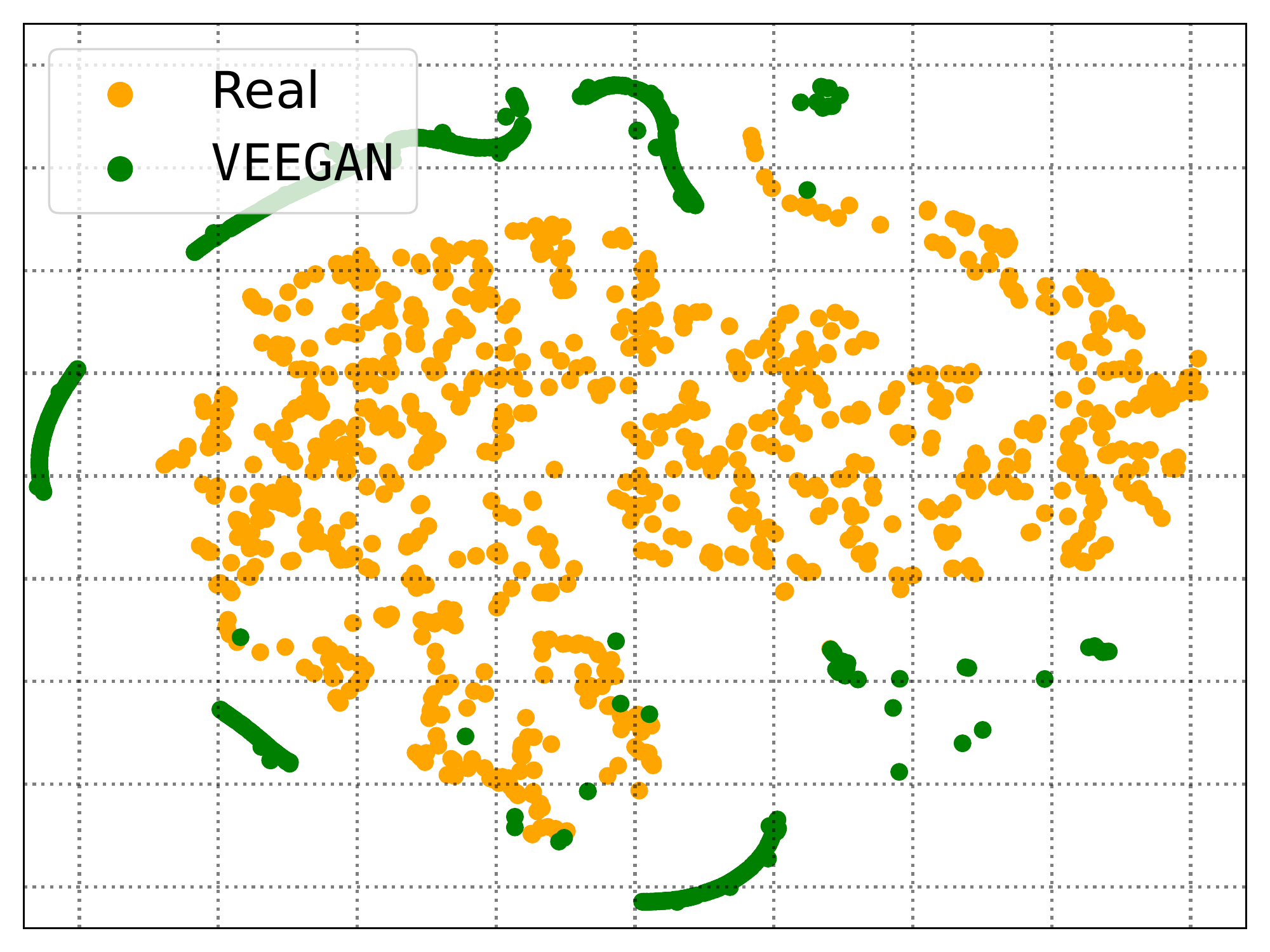}}
        \end{subfigure}
        \begin{subfigure}{\includegraphics[width=0.23\textwidth]{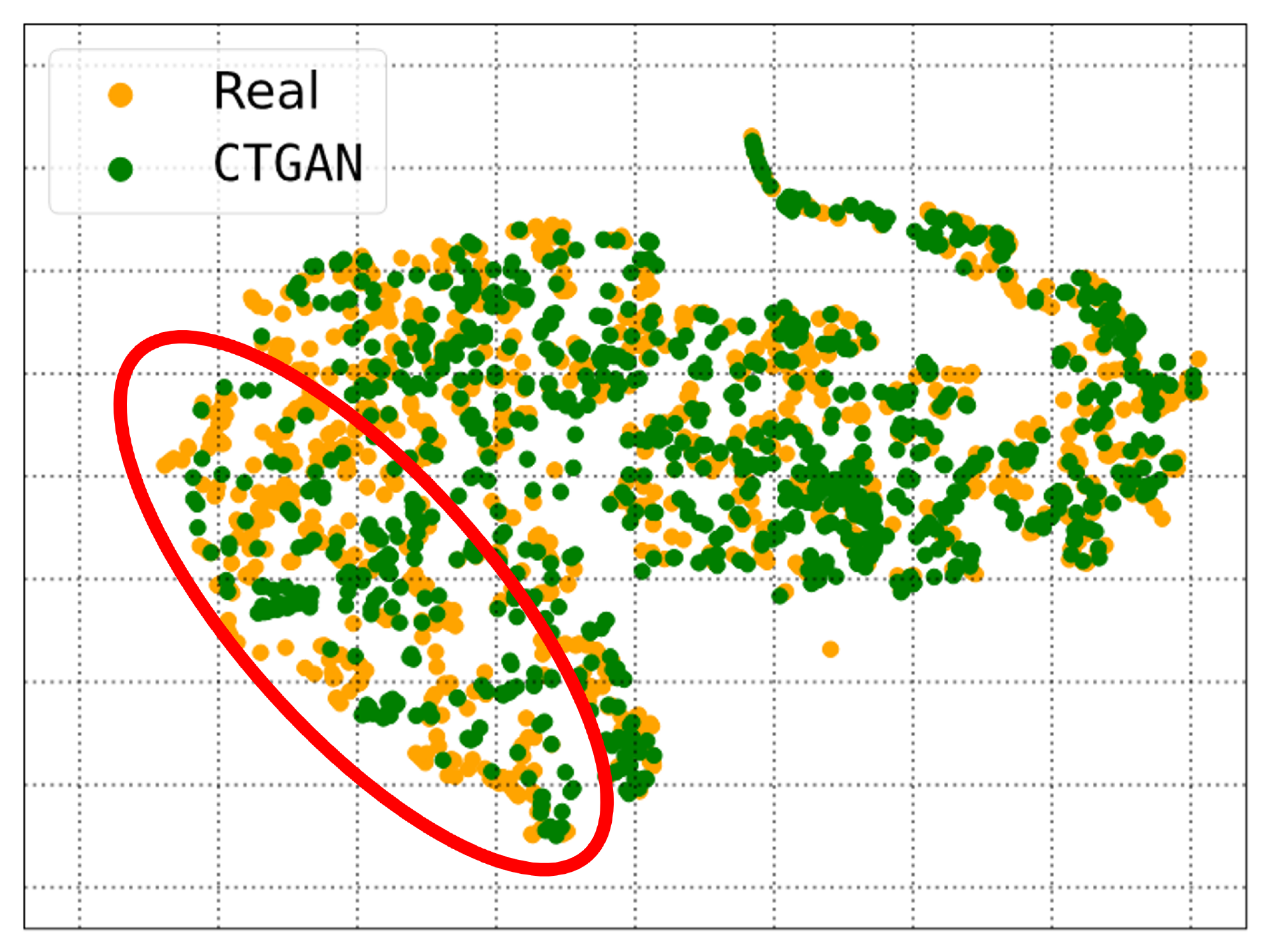}}
        \end{subfigure}
        % \\[\baselineskip]
        % \vspace{-1em}
        \begin{subfigure}{\includegraphics[width=0.23\textwidth]{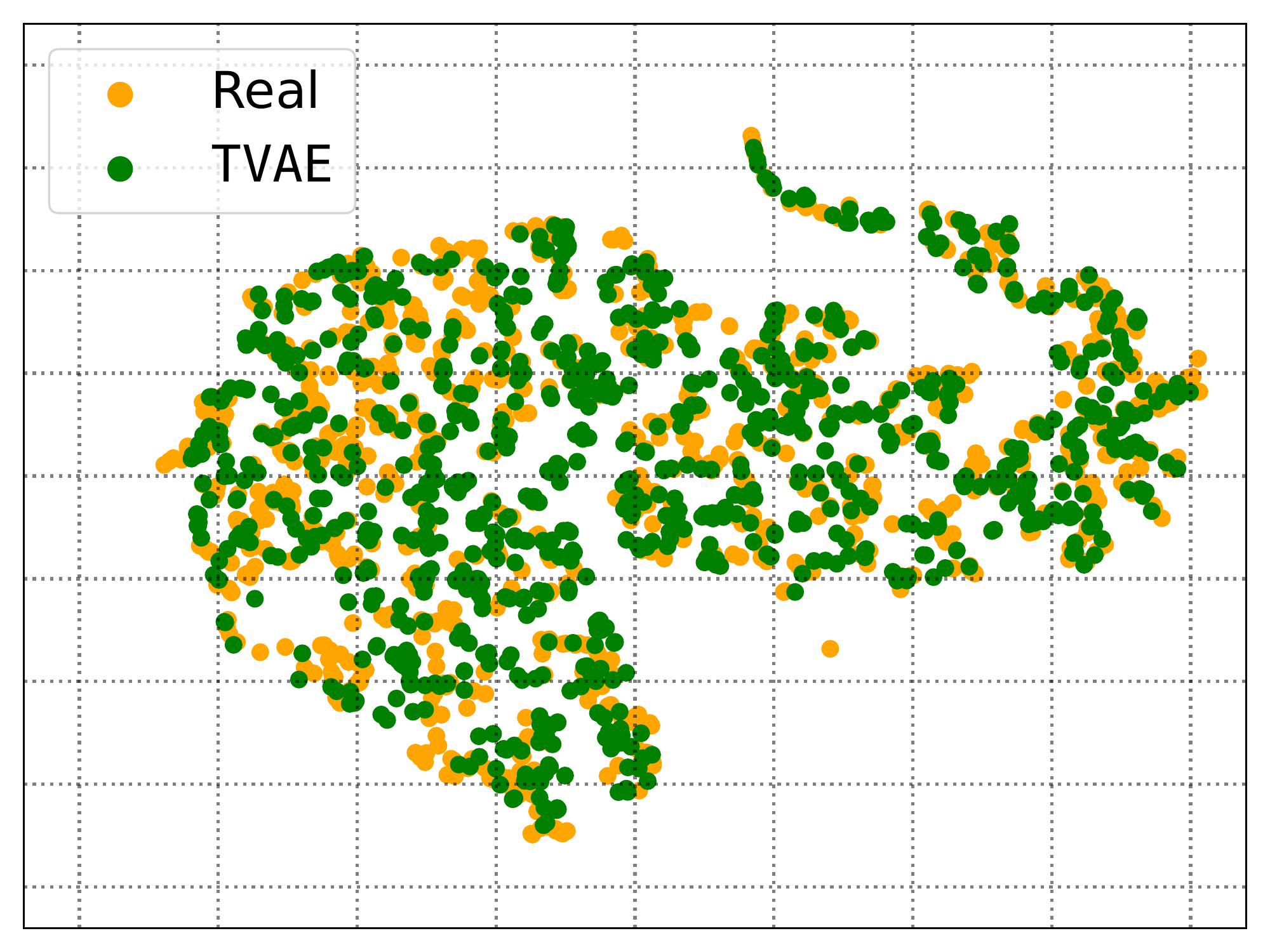}}
        \end{subfigure}
        \begin{subfigure}{\includegraphics[width=0.23\textwidth]{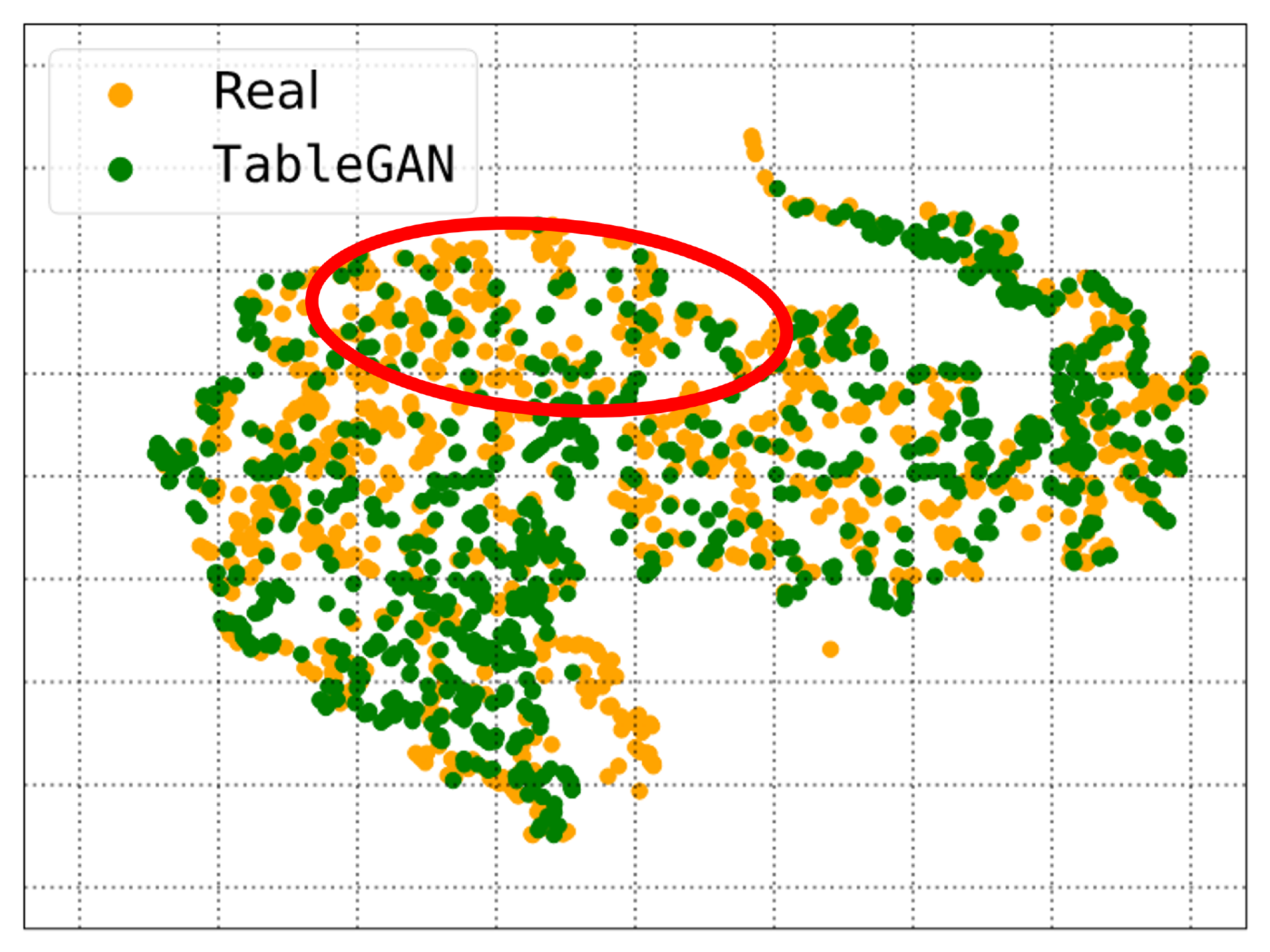}}
        \end{subfigure}
        \begin{subfigure}{\includegraphics[width=0.23\textwidth]{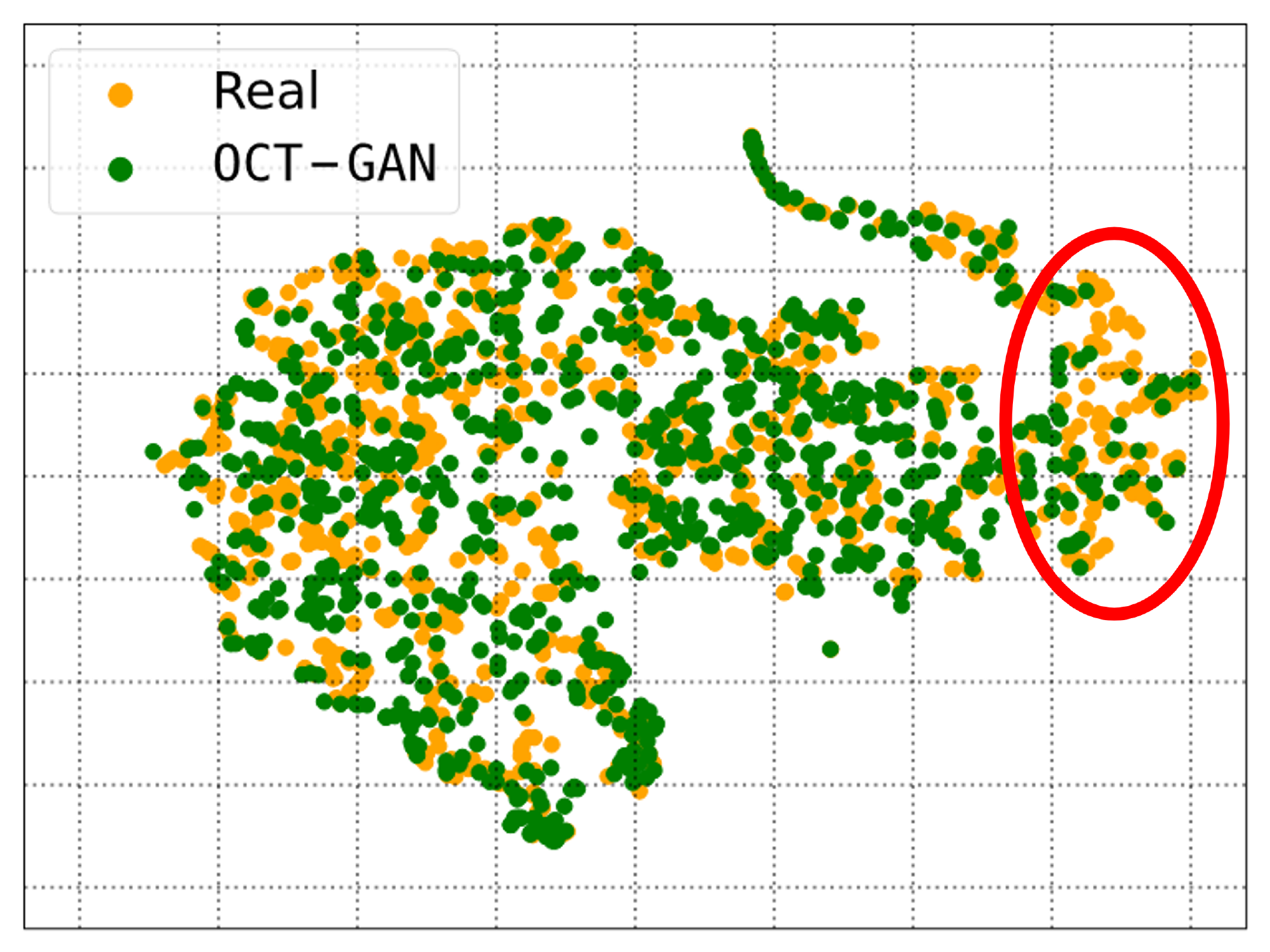}}
        \end{subfigure}
        \begin{subfigure}{\includegraphics[width=0.23\textwidth]{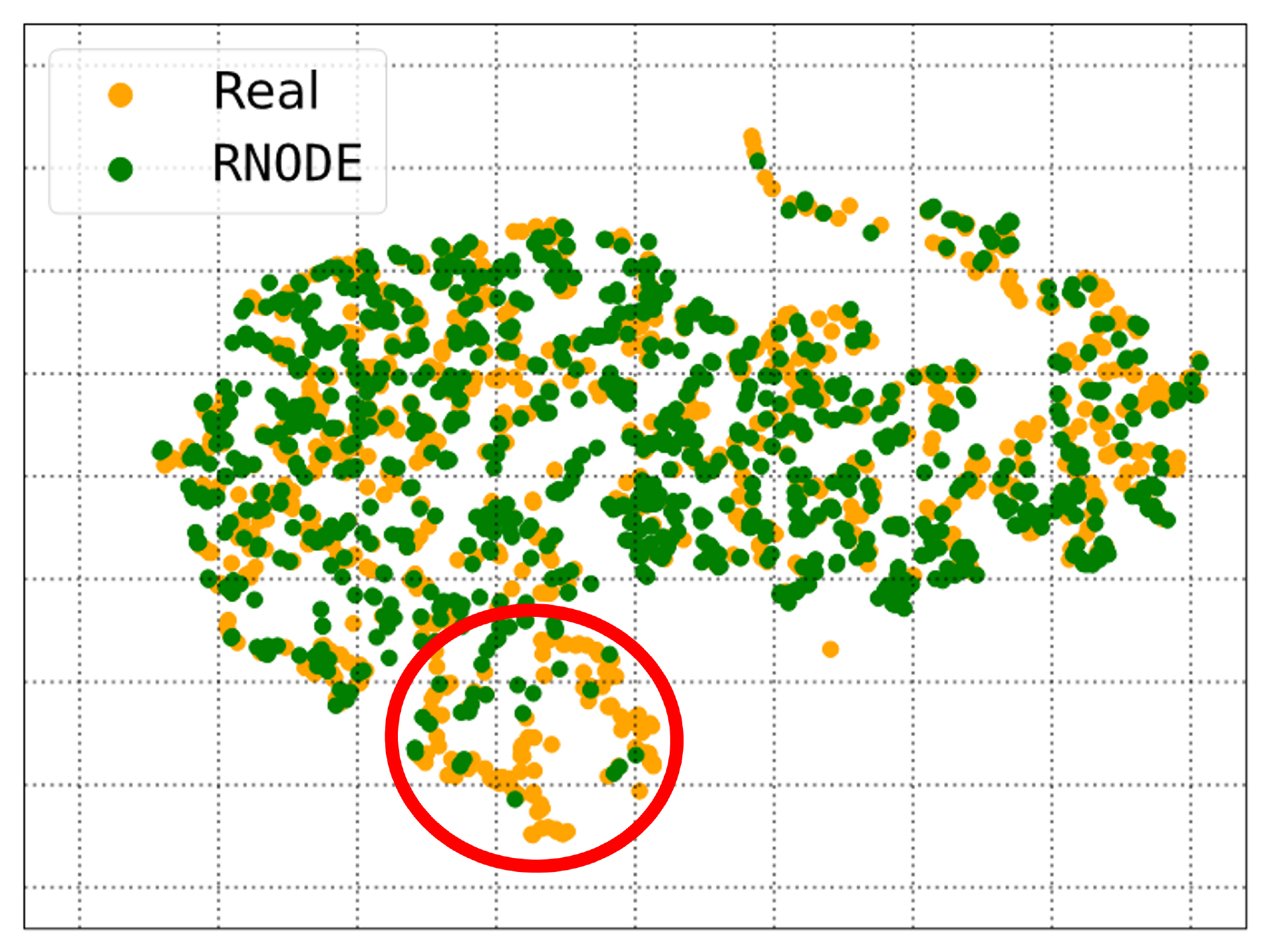}}
        \end{subfigure}
        \begin{subfigure}{\includegraphics[width=0.23\textwidth]{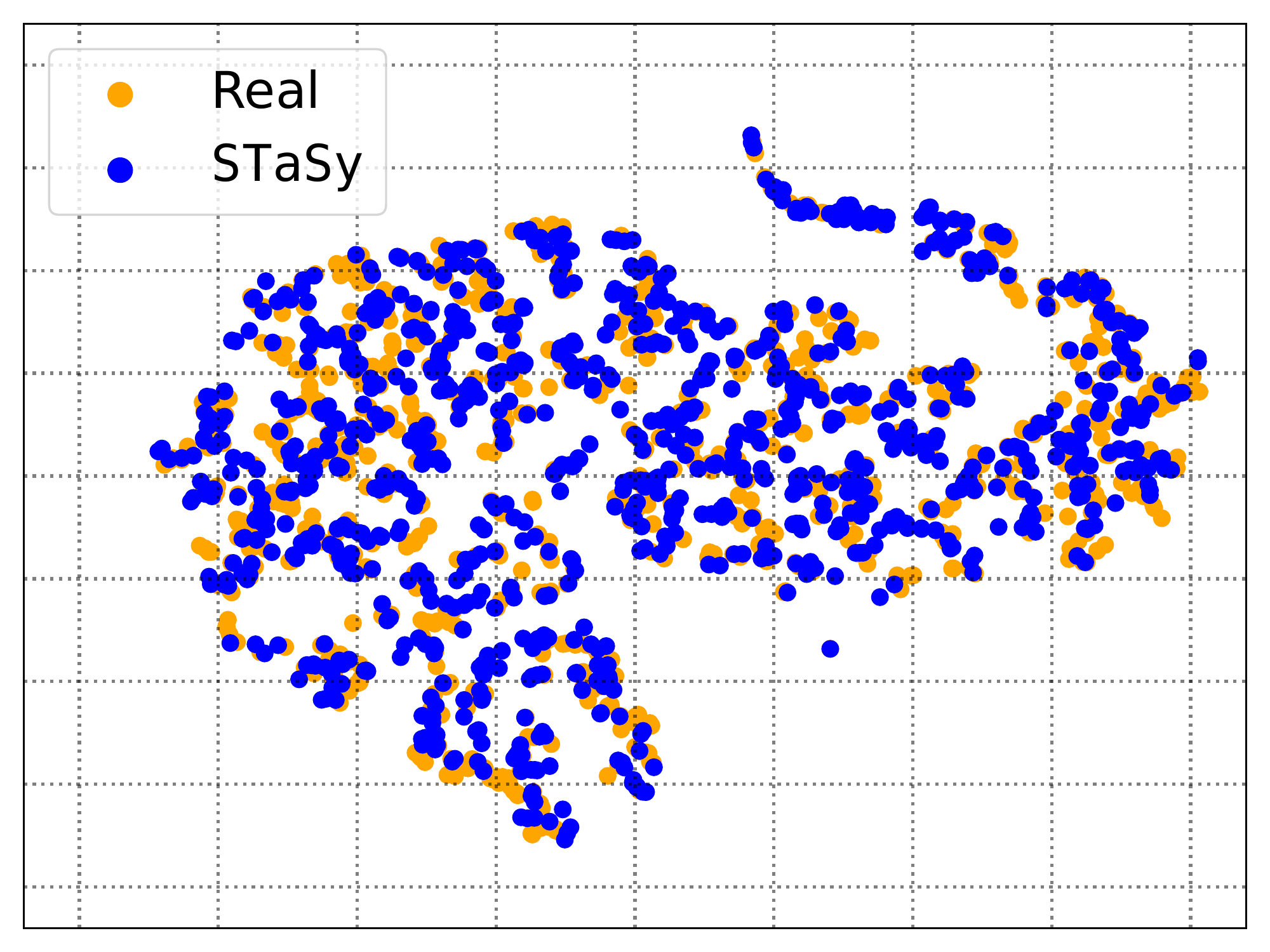}}
        \end{subfigure}
        \caption{t-SNE visualizations of fake and original records in \texttt{HTRU}}
        \label{fig:tsne_htru}
% \end{subfigure}
% \vspace{-5em}
\end{figure}

As shown in Figure~\ref{fig:histo_htru}, the fake data distributions of \texttt{TVAE}, \texttt{TableGAN}, and \texttt{RNODE} are dissimilar to the real data distributions, and these baselines fail to sample high-quality fake records.
% although they show a good performance in terms of task-dependant evaluations compared to \texttt{CTGAN} and \texttt{TVAE}.
In Figure~\ref{fig:tsne_htru}, \texttt{CTGAN}, \texttt{TableGAN}, \texttt{OCT-GAN}, and \texttt{RNODE} suffer from mode collapses as highlighted in red. In \texttt{STaSy}, however, the mode collapse problem is clearly alleviated, which means that \texttt{STaSy} is effective in enhancing the diversity.

\clearpage
\subsection{Additional visualizations in \texttt{Robot}}\label{sec:visualization_robot}

\begin{figure}[h]
% \vspace{-1em}
% \begin{subfigure}
        \centering
        \begin{subfigure}{\includegraphics[width=0.23\textwidth]{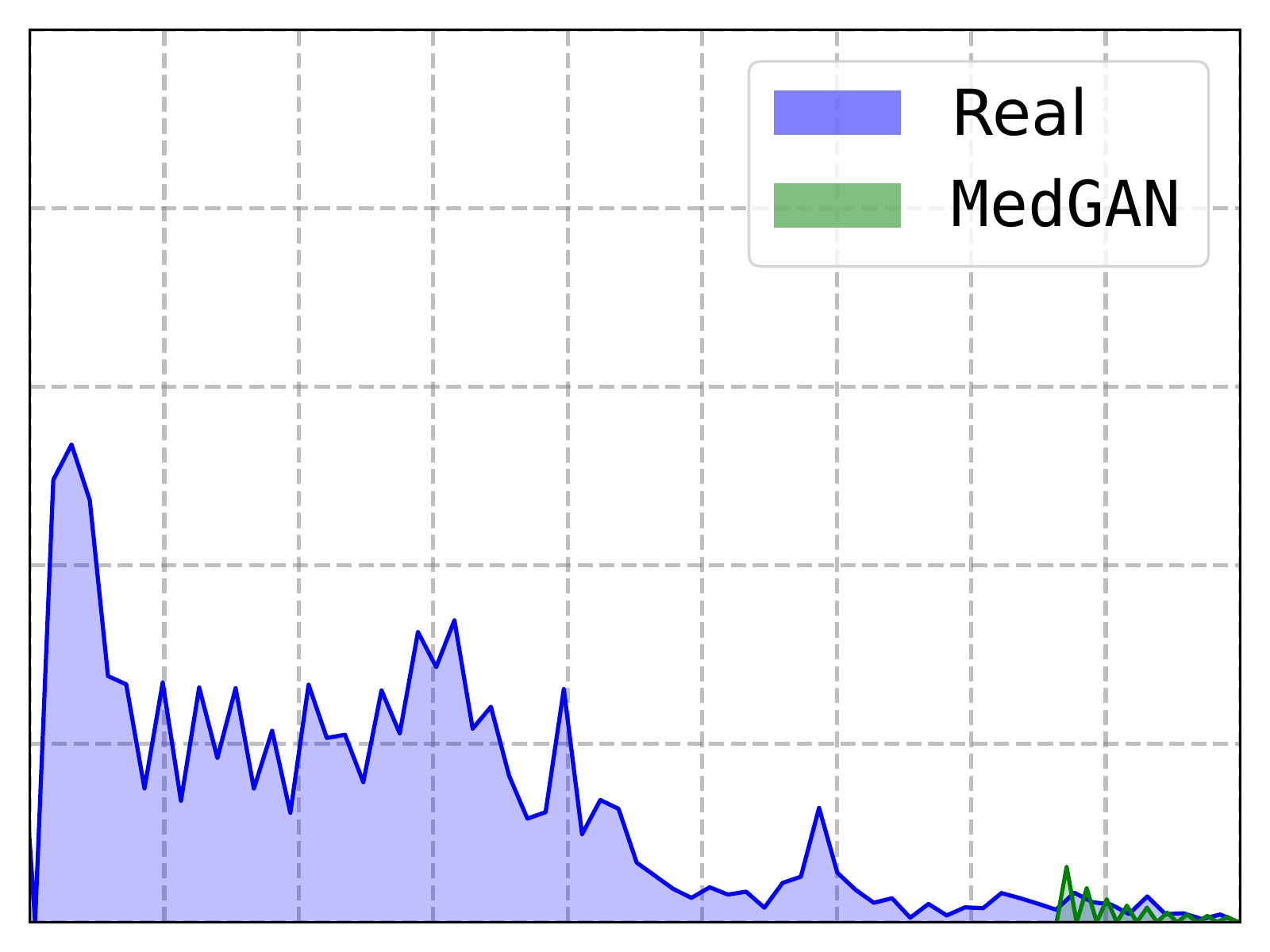}}
        \end{subfigure}
        \begin{subfigure}{\includegraphics[width=0.23\textwidth]{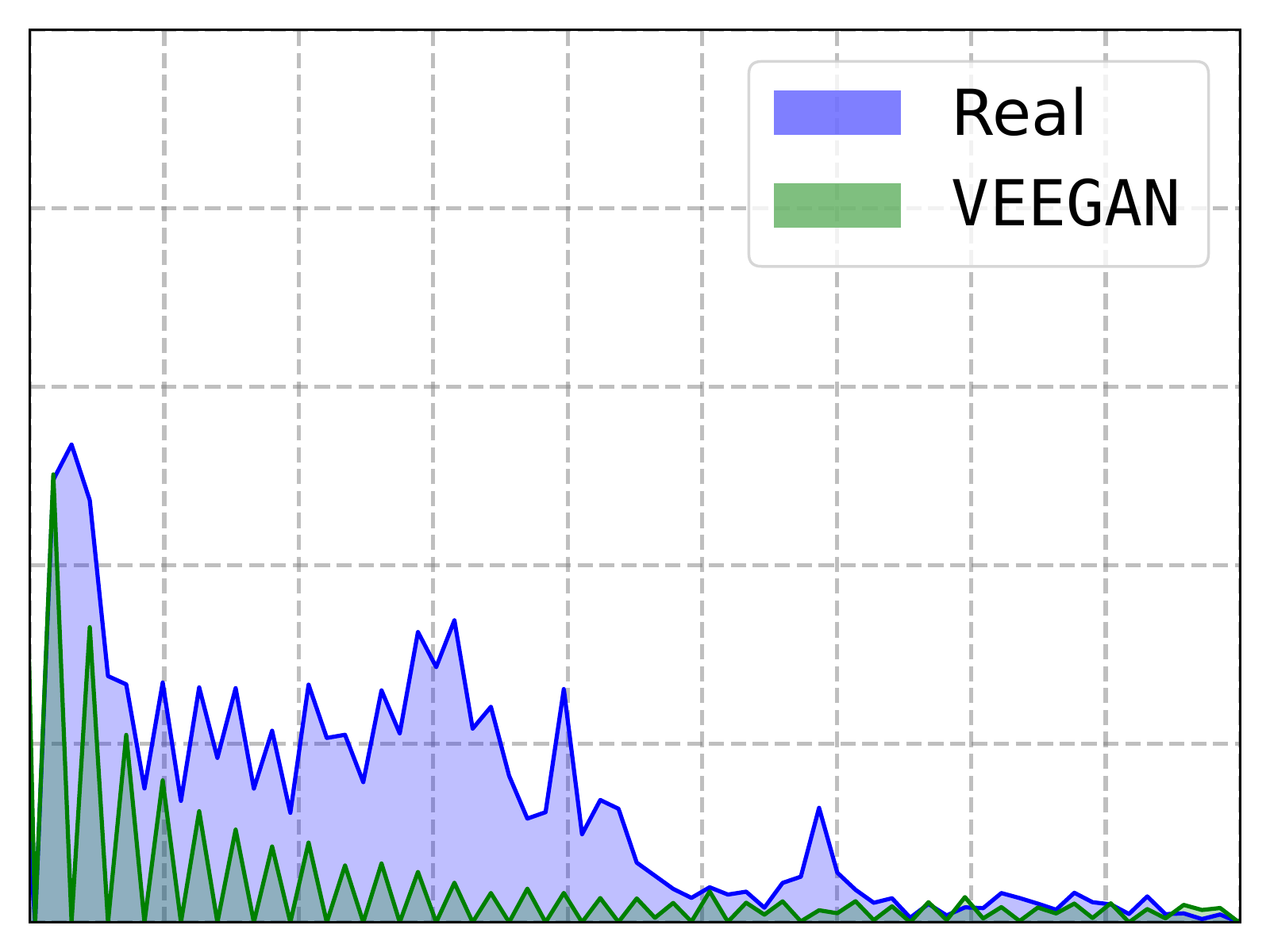}}
        \end{subfigure}
        \begin{subfigure}{\includegraphics[width=0.23\textwidth]{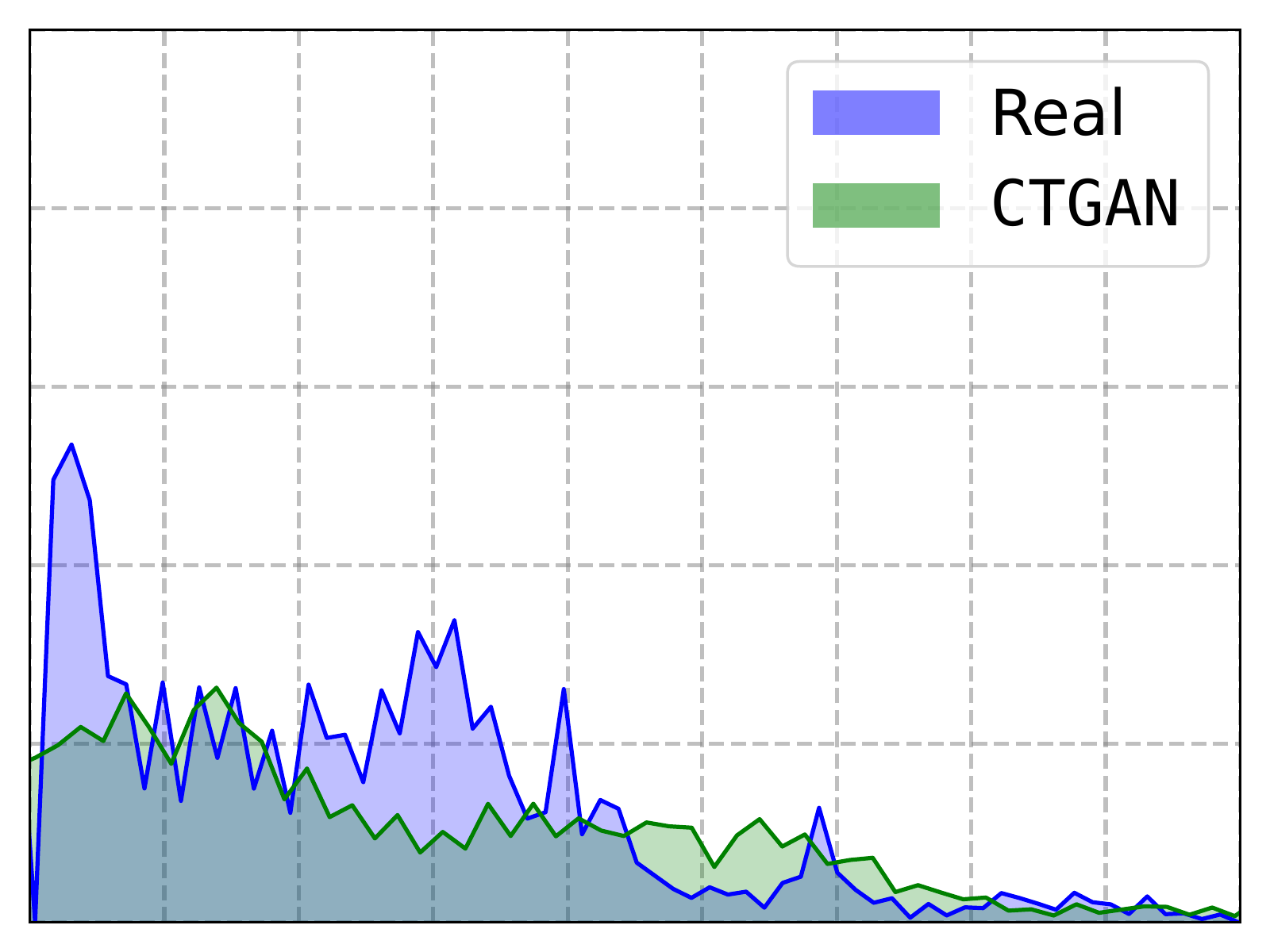}}
        \end{subfigure}
        \begin{subfigure}{\includegraphics[width=0.23\textwidth]{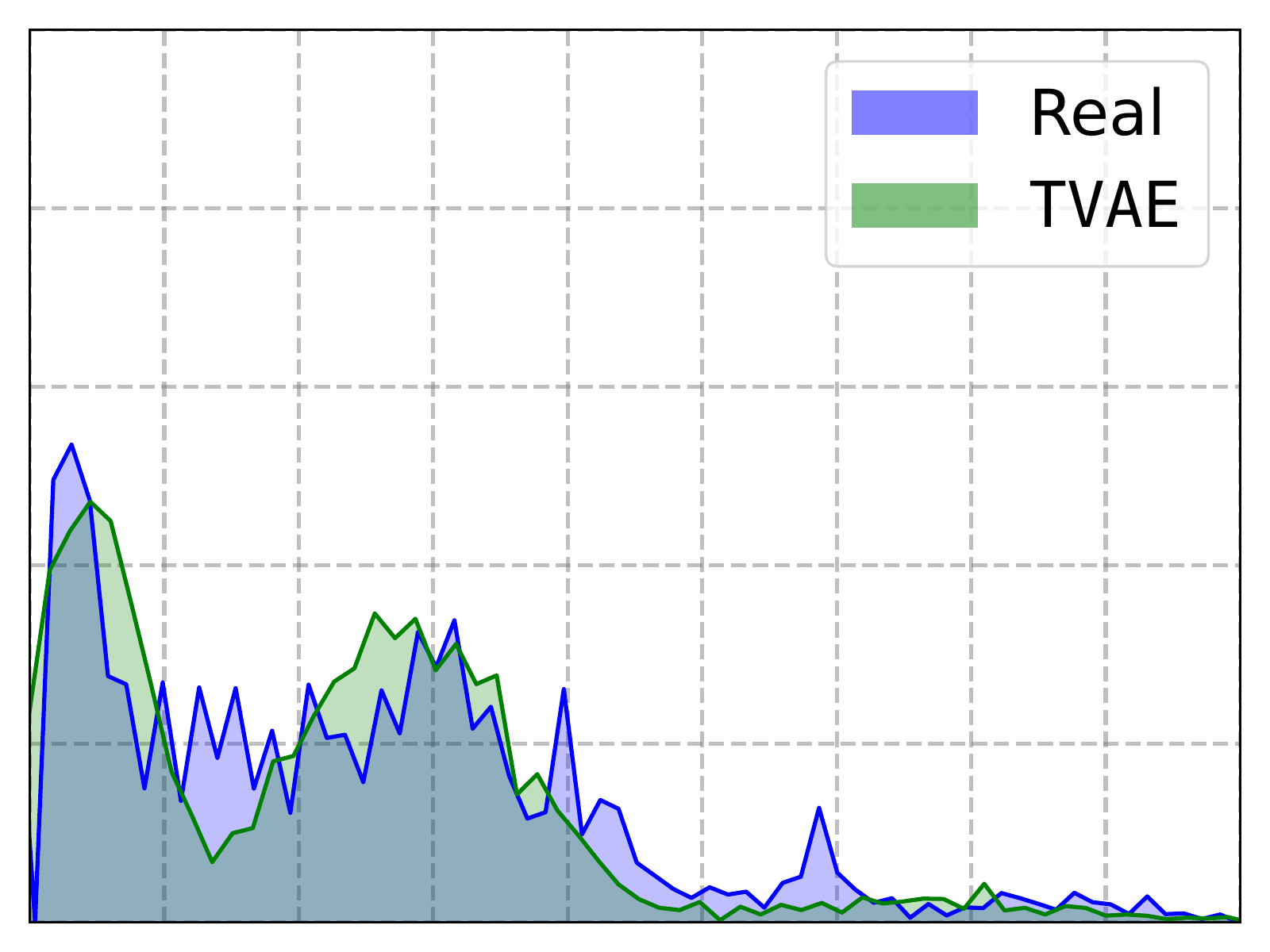}}
        \end{subfigure}
        \begin{subfigure}{\includegraphics[width=0.23\textwidth]{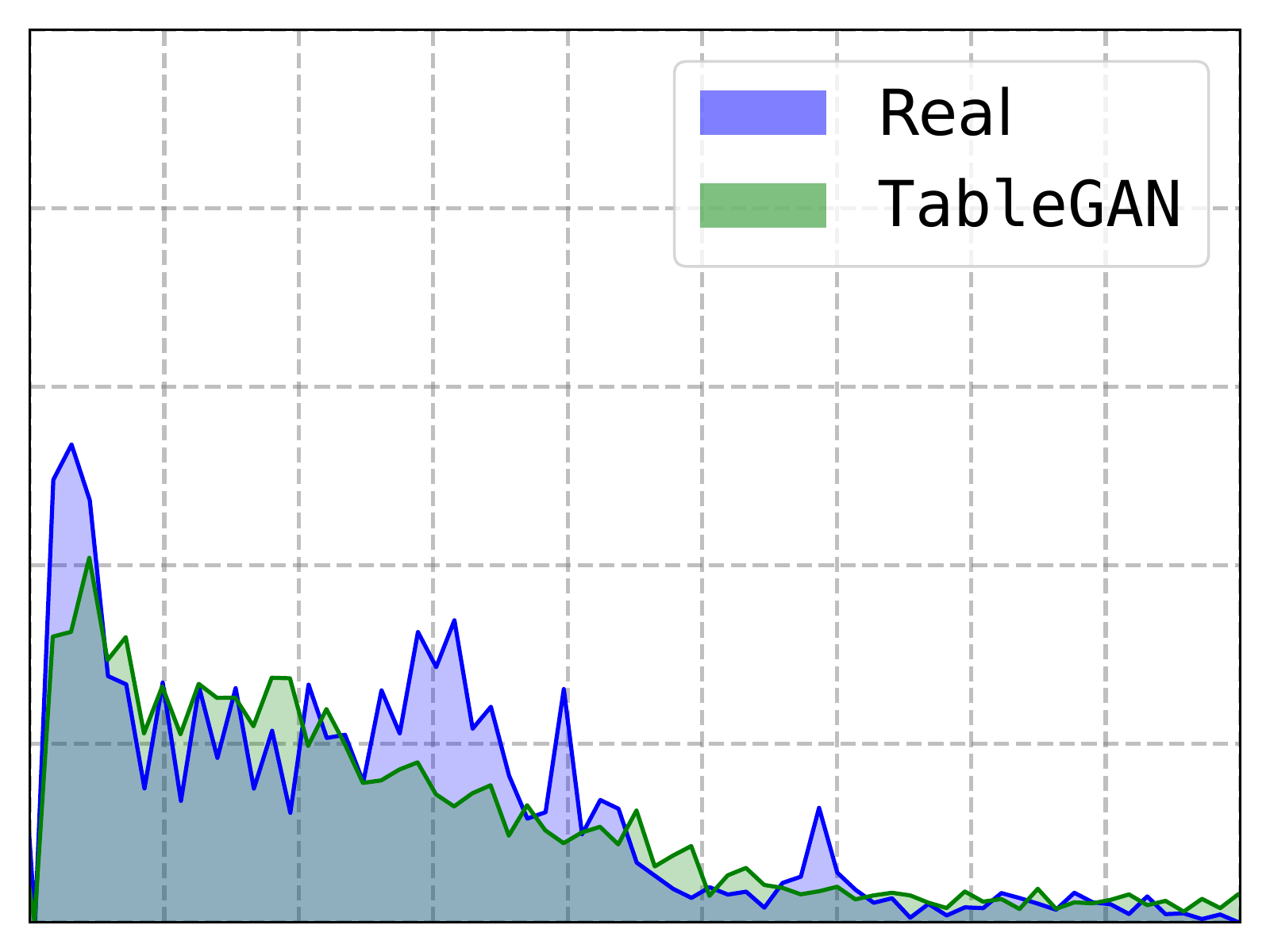}}
        \end{subfigure}
        \begin{subfigure}{\includegraphics[width=0.23\textwidth]{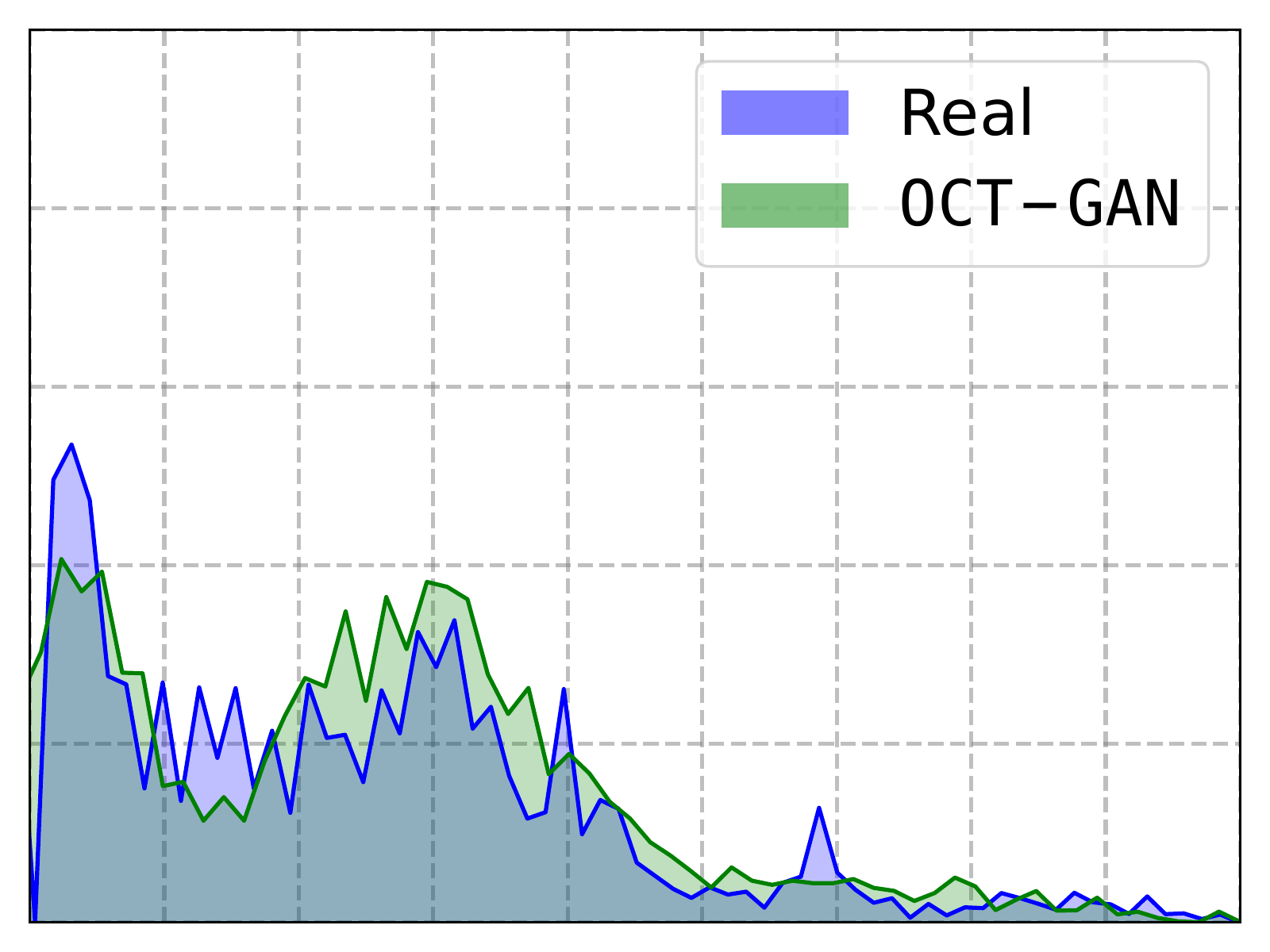}}
        \end{subfigure}
        \begin{subfigure}{\includegraphics[width=0.23\textwidth]{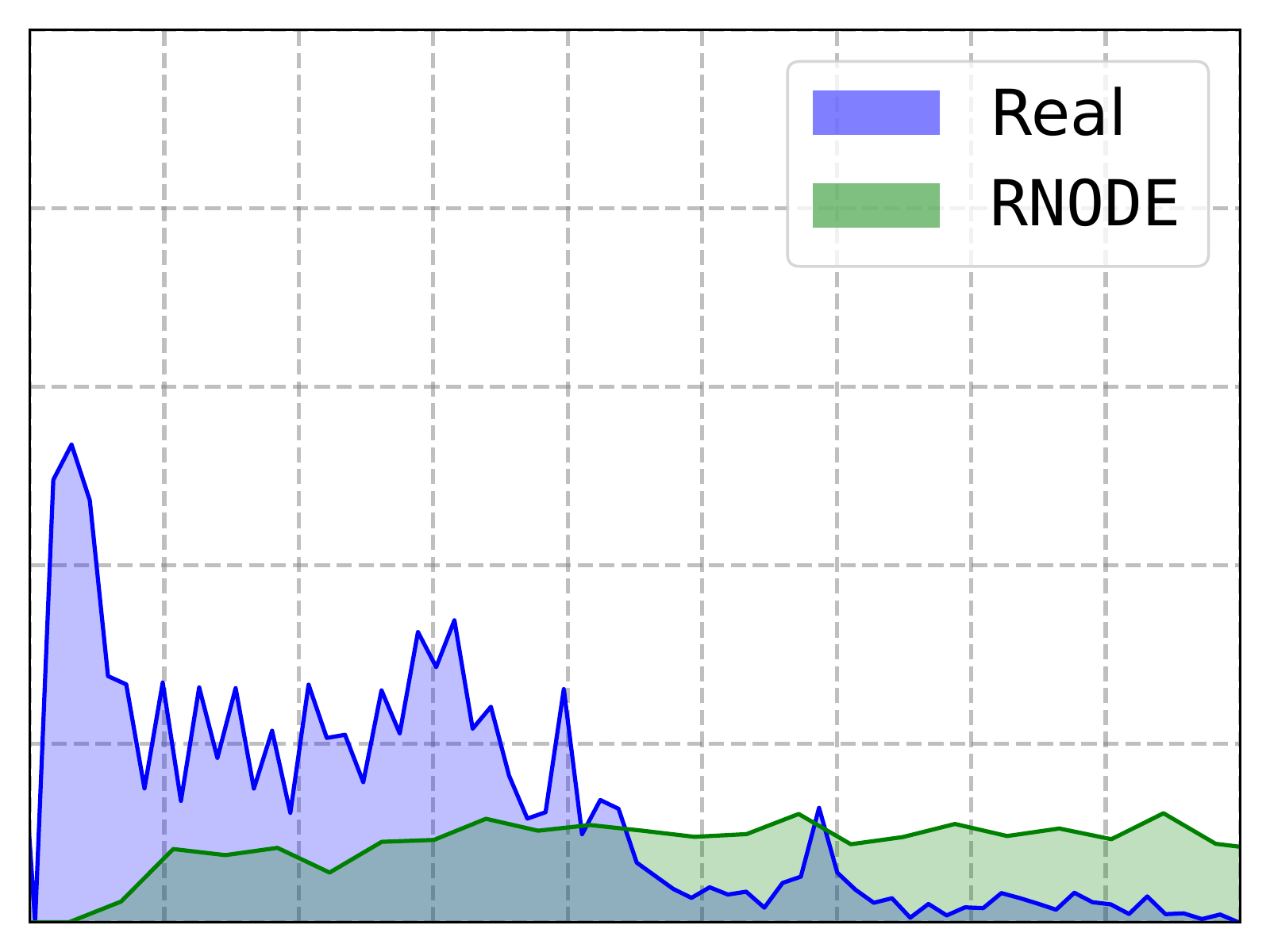}}
        \end{subfigure}
        \begin{subfigure}{\includegraphics[width=0.23\textwidth]{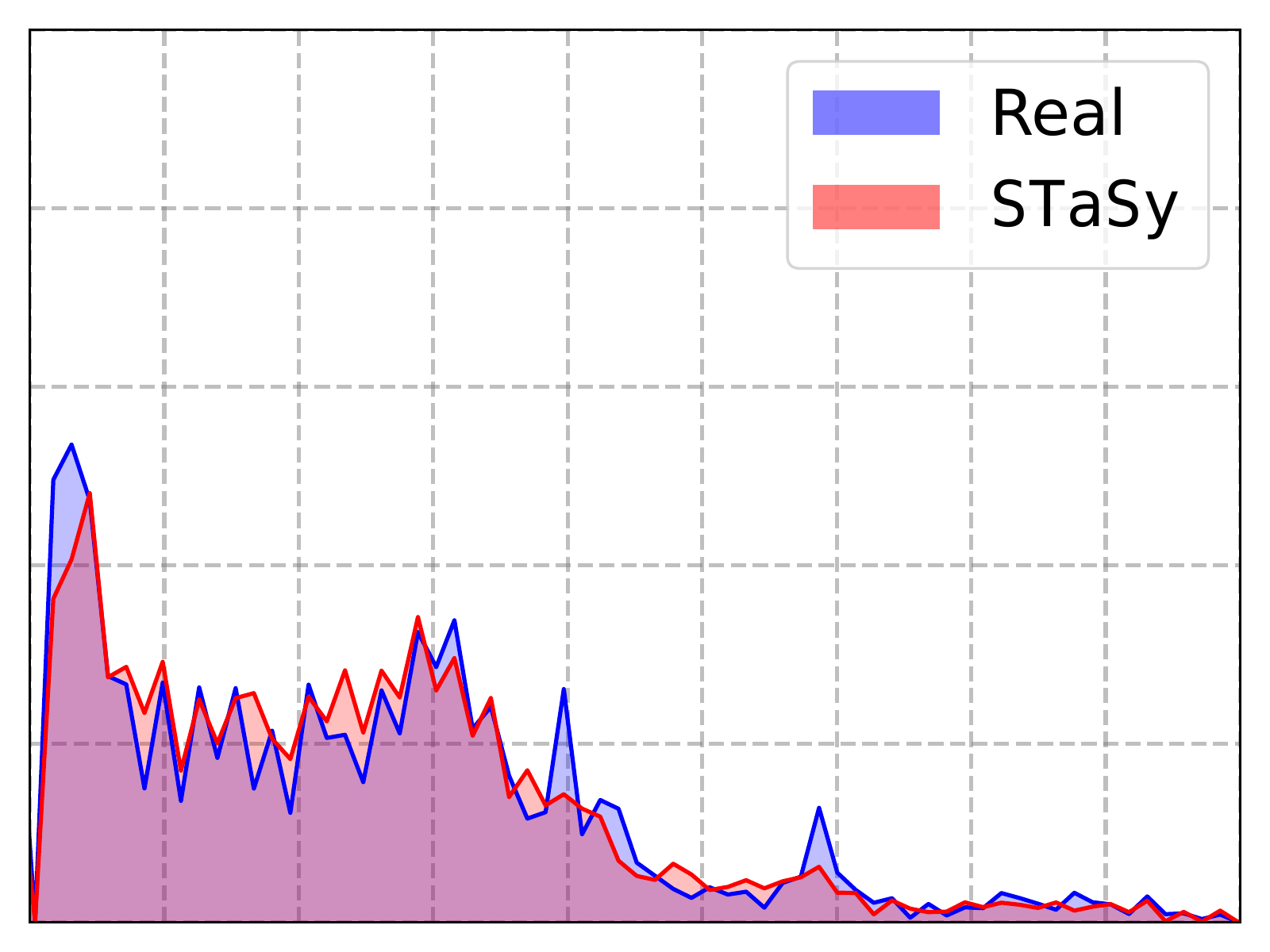}}
        \end{subfigure}
        \caption{Histograms of values in the \textit{ultrasound sensor} column of \texttt{Robot}}
        \label{fig:histo_robot}
% \end{subfigure}
% \vspace{-2em}
\end{figure}

\begin{figure}[h]
% \vspace{-2em}
% \begin{subfigure}
        \centering
        \begin{subfigure}{\includegraphics[width=0.23\textwidth]{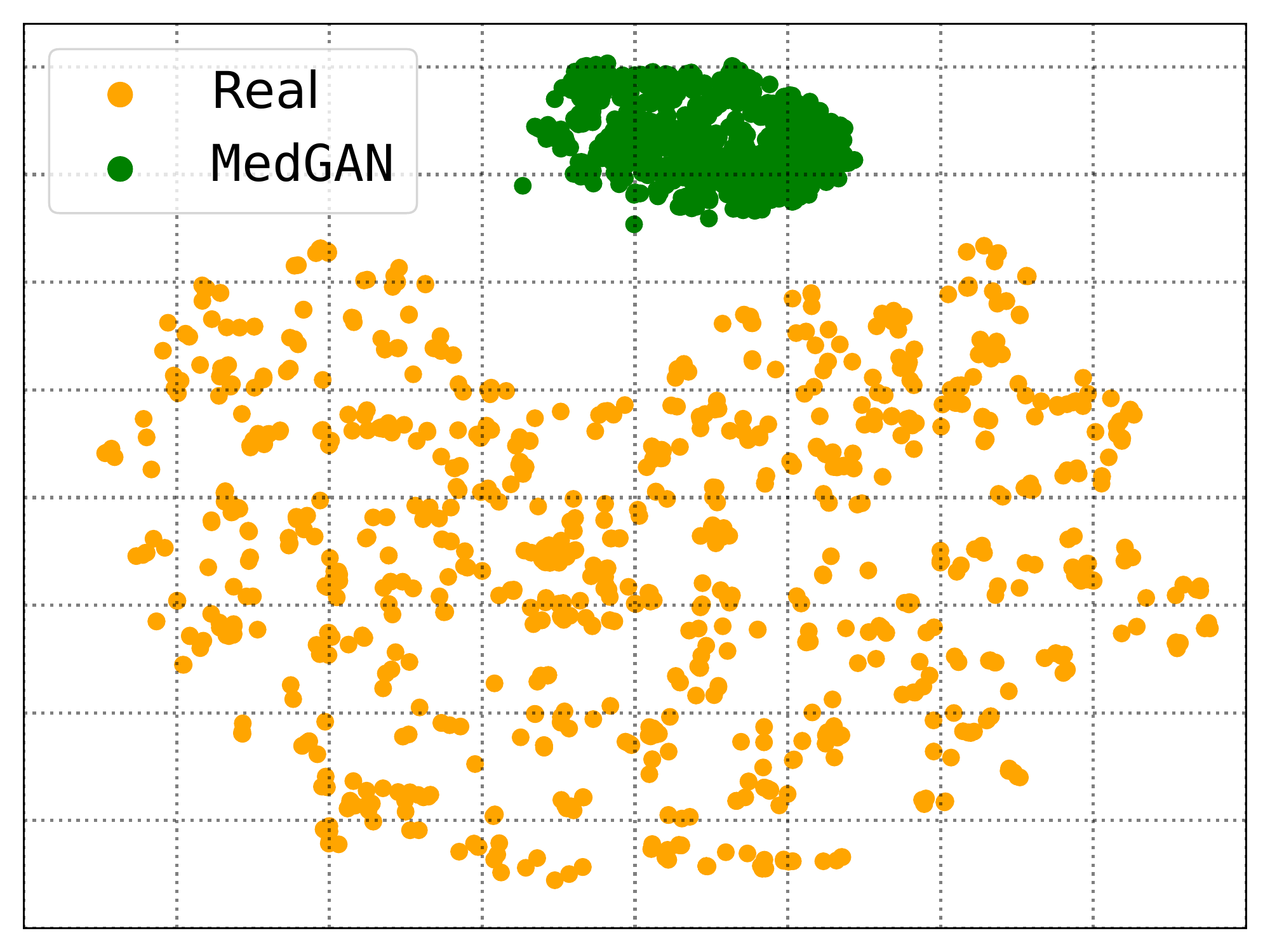}}
        \end{subfigure}
        \begin{subfigure}{\includegraphics[width=0.23\textwidth]{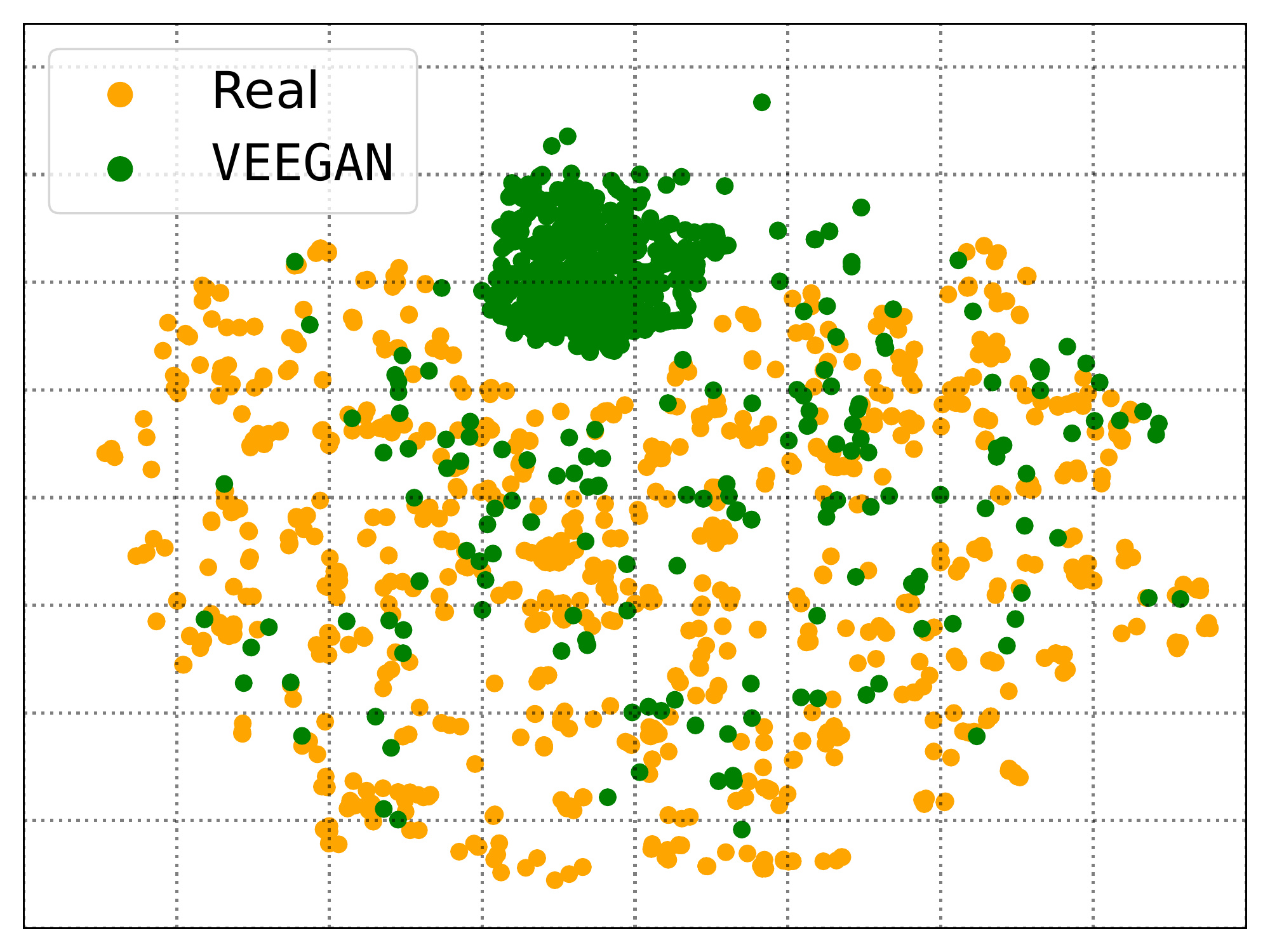}}
        \end{subfigure}
        \begin{subfigure}{\includegraphics[width=0.23\textwidth]{images/robot/CTGAN_1.pdf}}
        \end{subfigure}
        \begin{subfigure}{\includegraphics[width=0.23\textwidth]{images/robot/TVAE_1.pdf}}
        \end{subfigure}
        \begin{subfigure}{\includegraphics[width=0.23\textwidth]{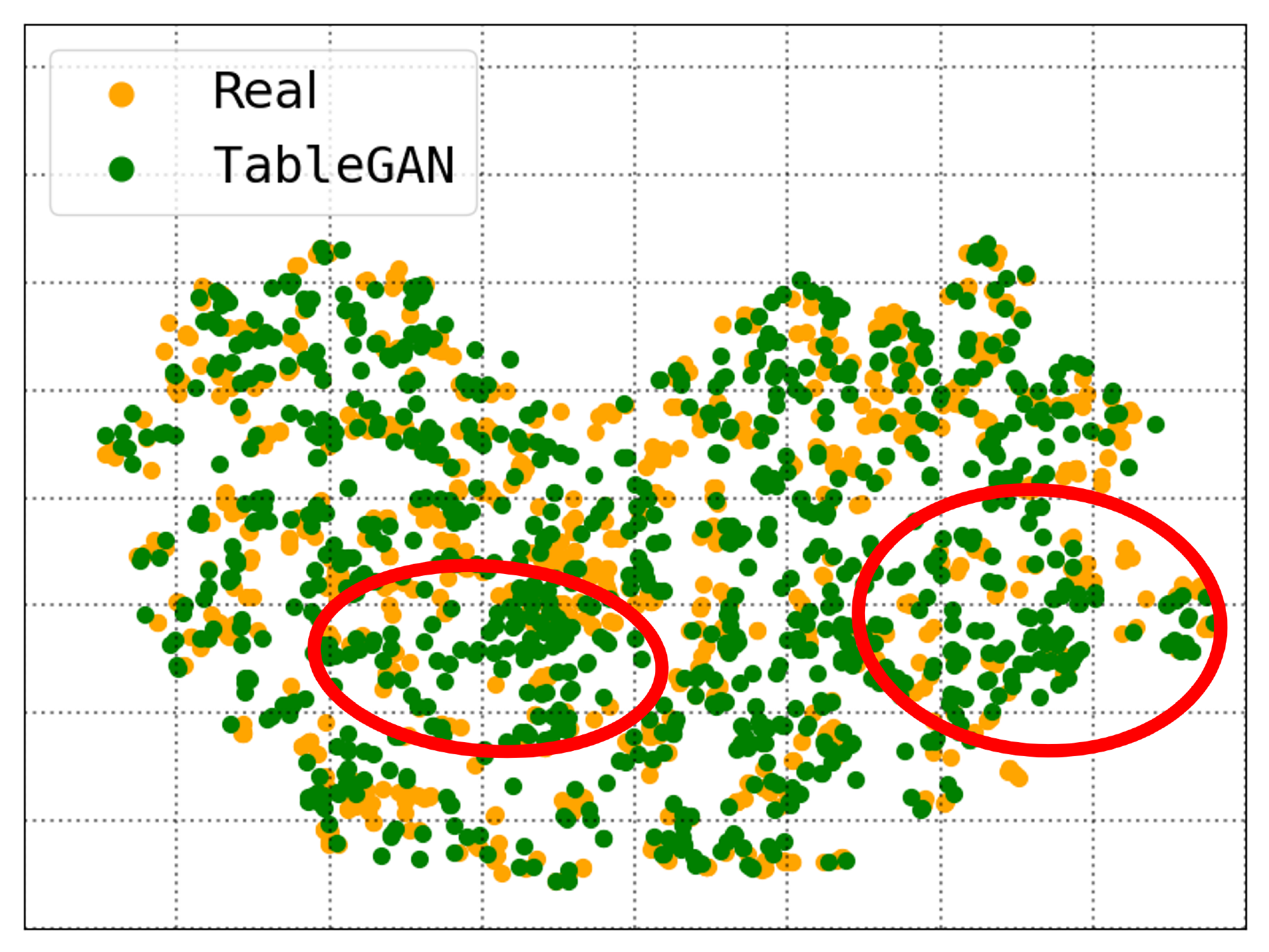}}
        \end{subfigure}
        \begin{subfigure}{\includegraphics[width=0.23\textwidth]{images/robot/OCT-GAN_1.pdf}}
        \end{subfigure}
        \begin{subfigure}{\includegraphics[width=0.23\textwidth]{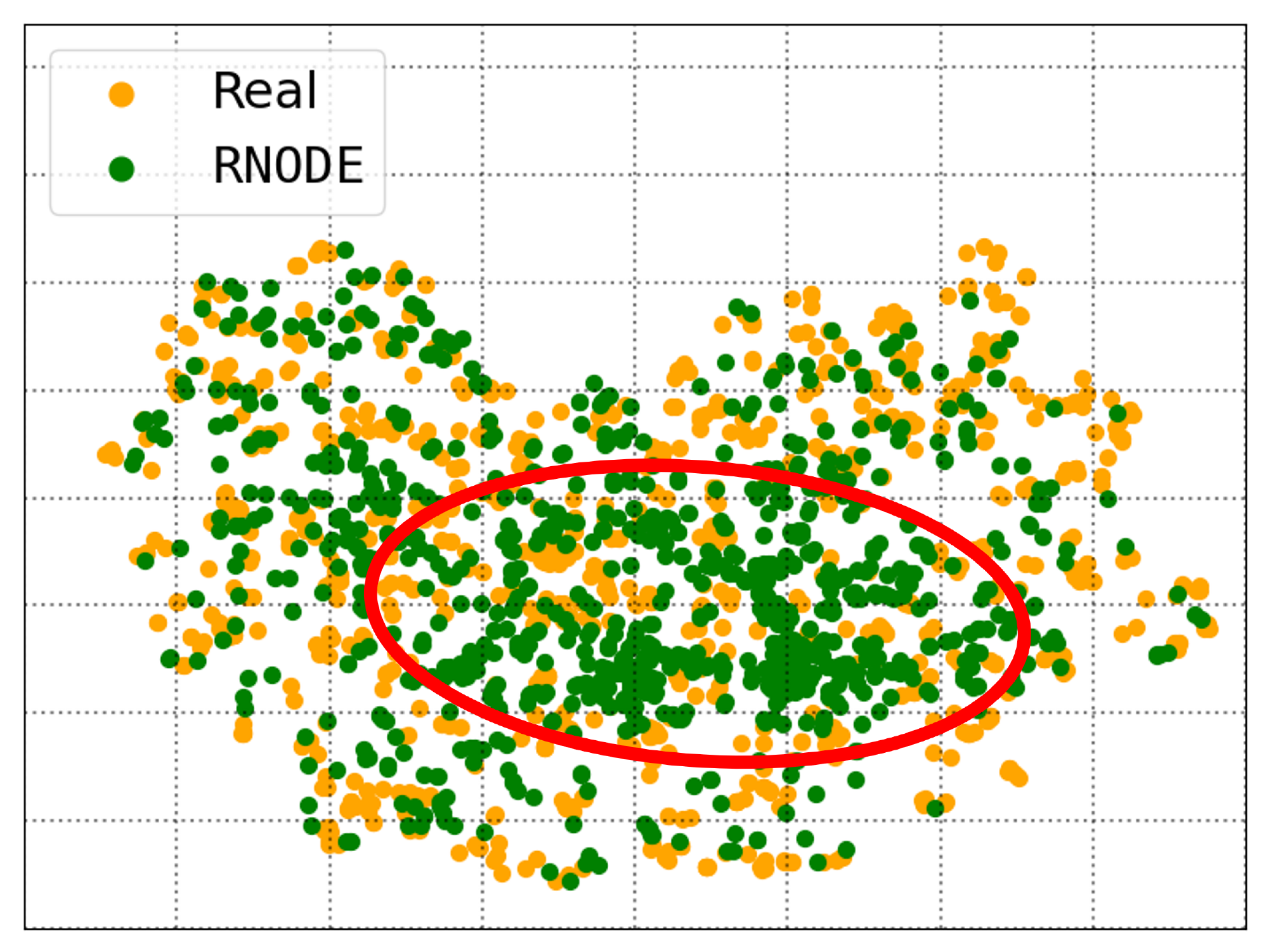}}
        \end{subfigure}
        \begin{subfigure}{\includegraphics[width=0.23\textwidth]{images/robot/STDS.pdf}}
        \end{subfigure}
        \caption{t-SNE visualizations of fake and original records in \texttt{Robot}}
        \label{fig:tsne_robot}
% \end{subfigure}
% \vspace{-3em}
\end{figure}

In \texttt{Robot}, \texttt{TVAE}, \texttt{OCT-GAN}, and \texttt{STaSy} generate relatively similar distribution to that of real data as shown in Figure~\ref{fig:histo_robot}. In Figure~\ref{fig:tsne_robot}, \texttt{CTGAN}, \texttt{TableGAN}, and \texttt{RNODE} generate out-of-distribution records as highlighted in red, and \texttt{TVAE} and \texttt{OCT-GAN} suffer from mode collapses. Our proposed methods generate the most diverse records among various methods.

% all methods except for \texttt{CTGAN} and \texttt{RNODE} well generate fake records in Fig.~\ref{fig:tsne_robot}.

\clearpage
\subsection{Additional visualizations in \texttt{News}}

\begin{figure}[h]
% \begin{subfigure}
        \centering
         \begin{subfigure}{\includegraphics[width=0.23\textwidth]{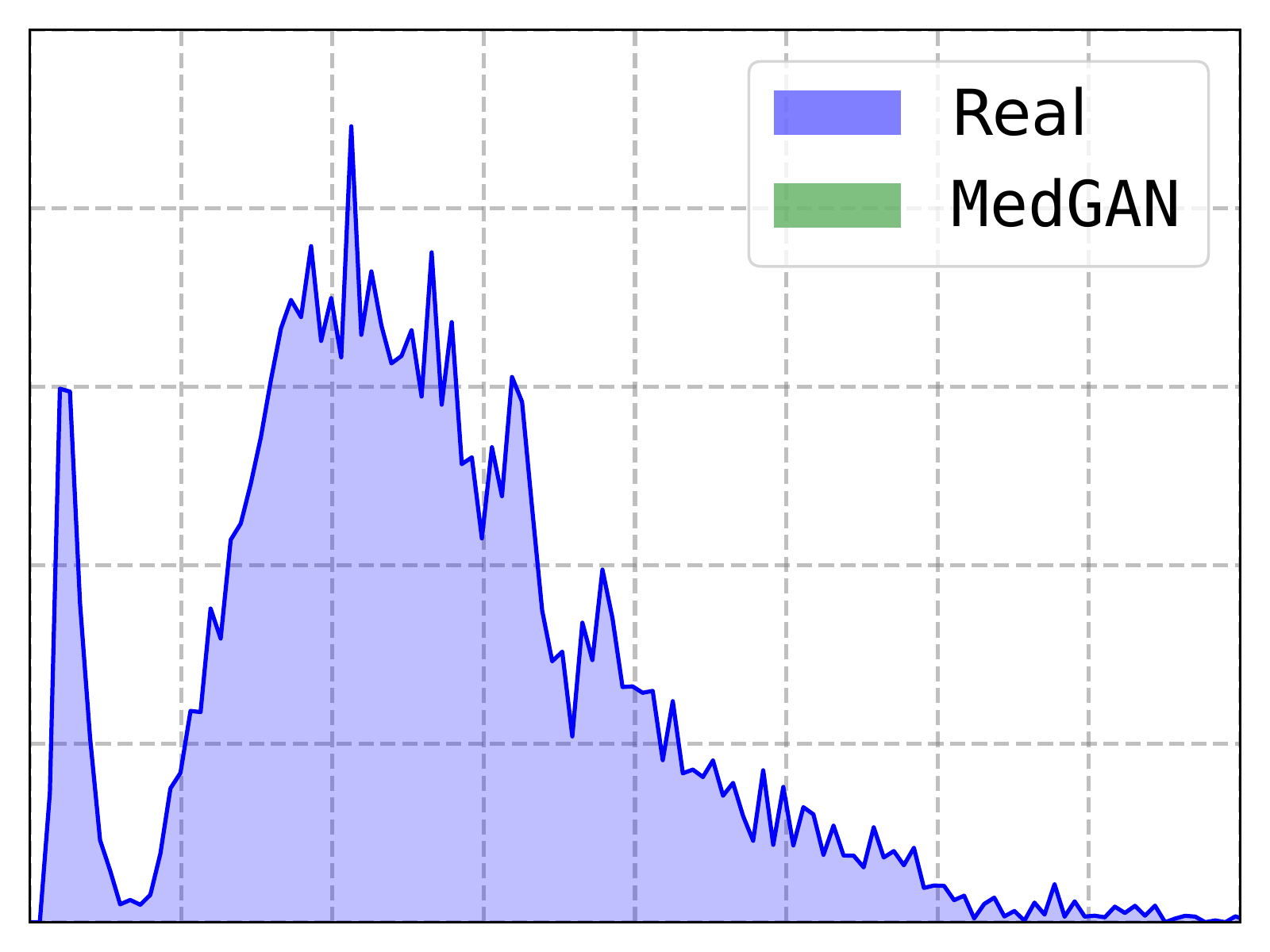}}
        \end{subfigure}
        \begin{subfigure}{\includegraphics[width=0.23\textwidth]{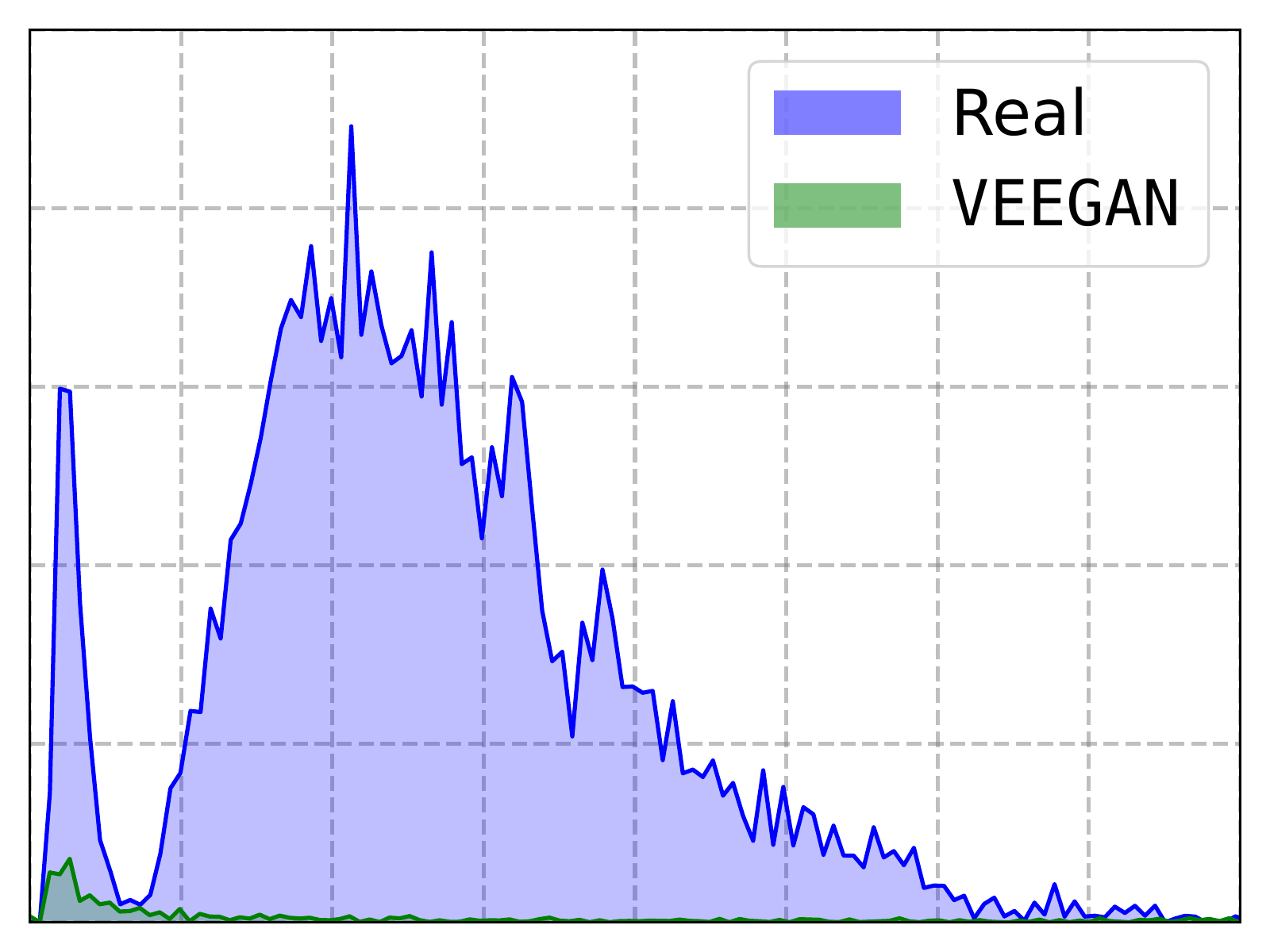}}
        \end{subfigure}
        \begin{subfigure}{\includegraphics[width=0.23\textwidth]{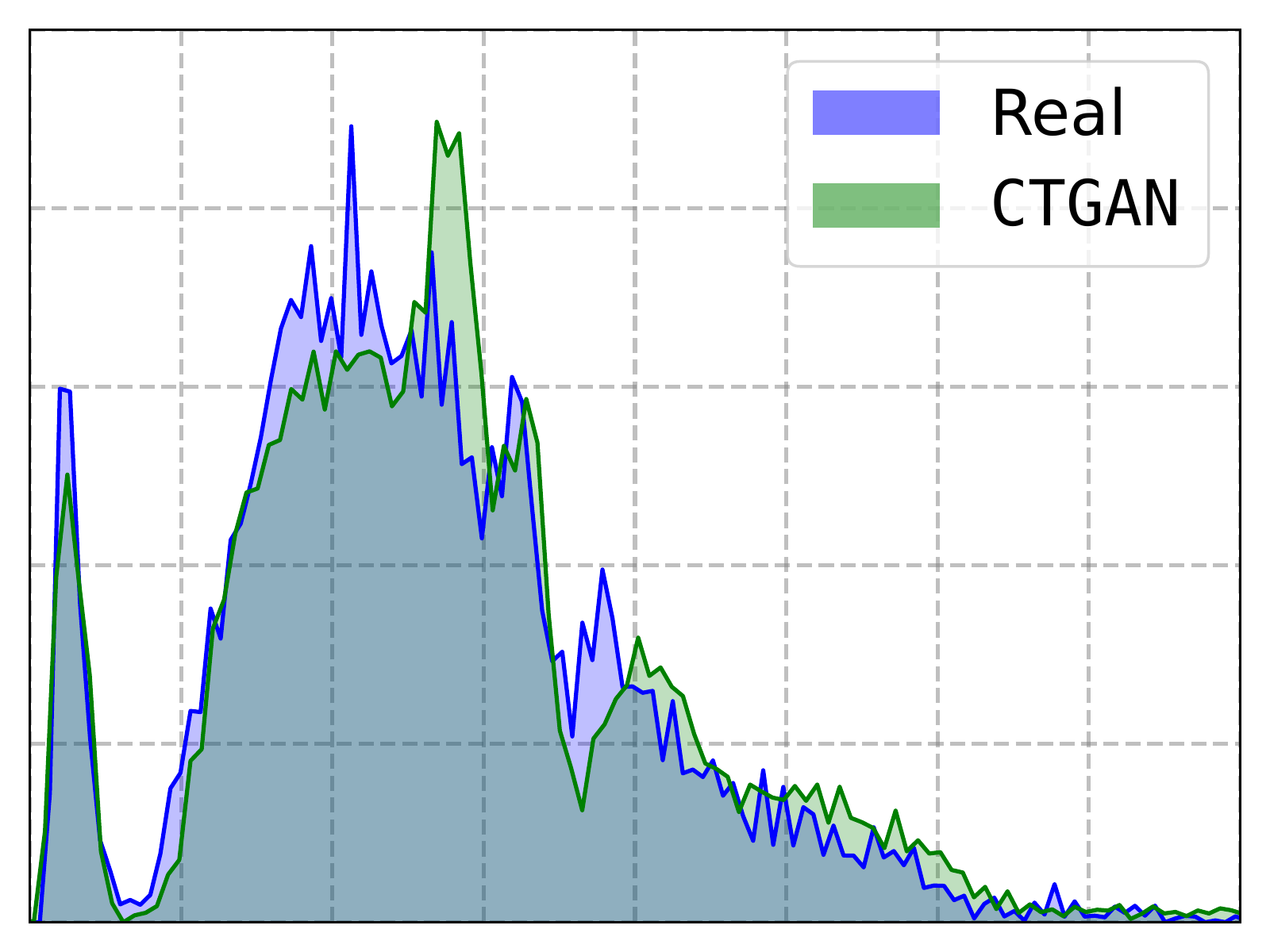}}
        \end{subfigure}
        \begin{subfigure}{\includegraphics[width=0.23\textwidth]{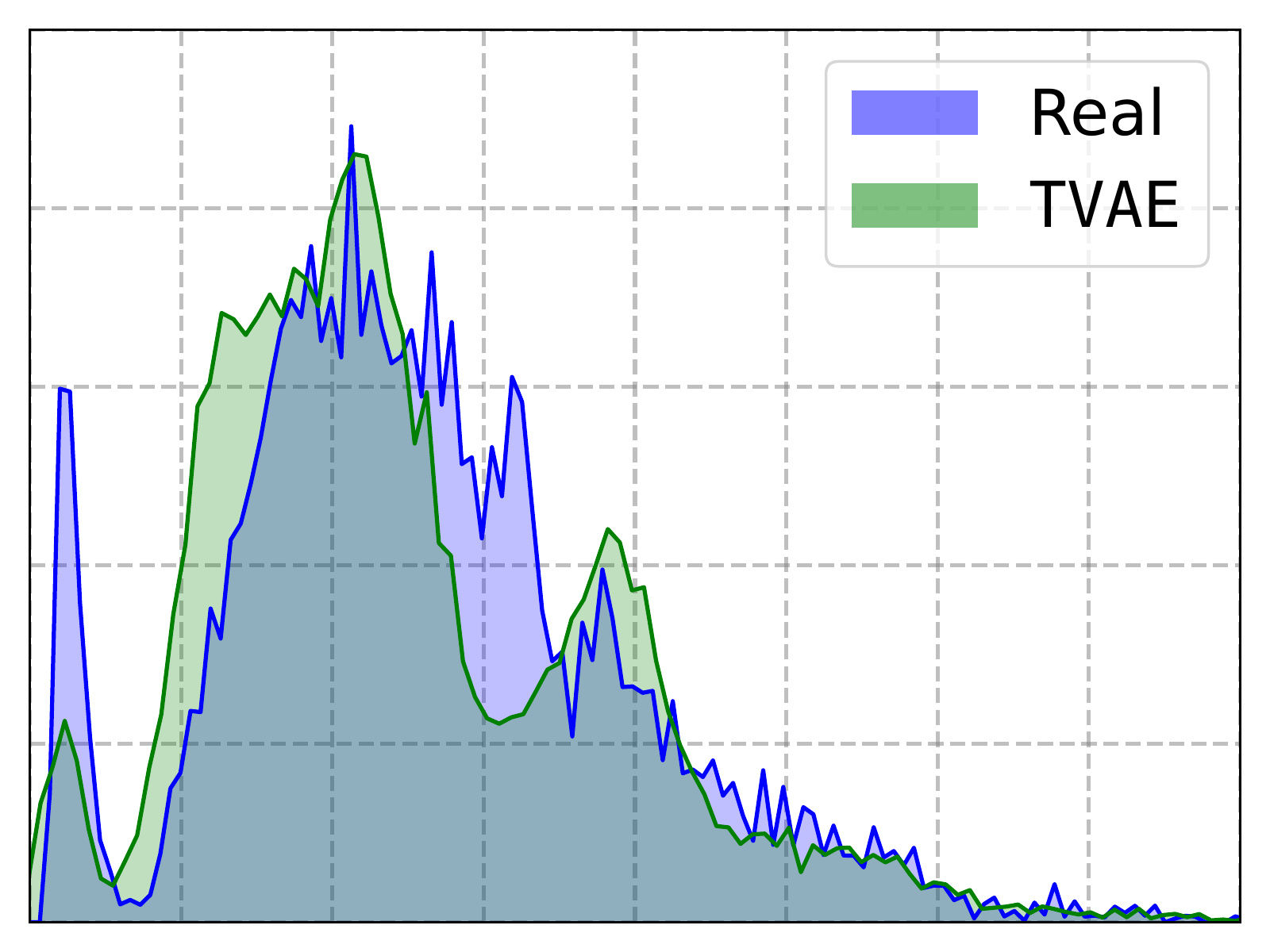}}
        \end{subfigure}
        \begin{subfigure}{\includegraphics[width=0.23\textwidth]{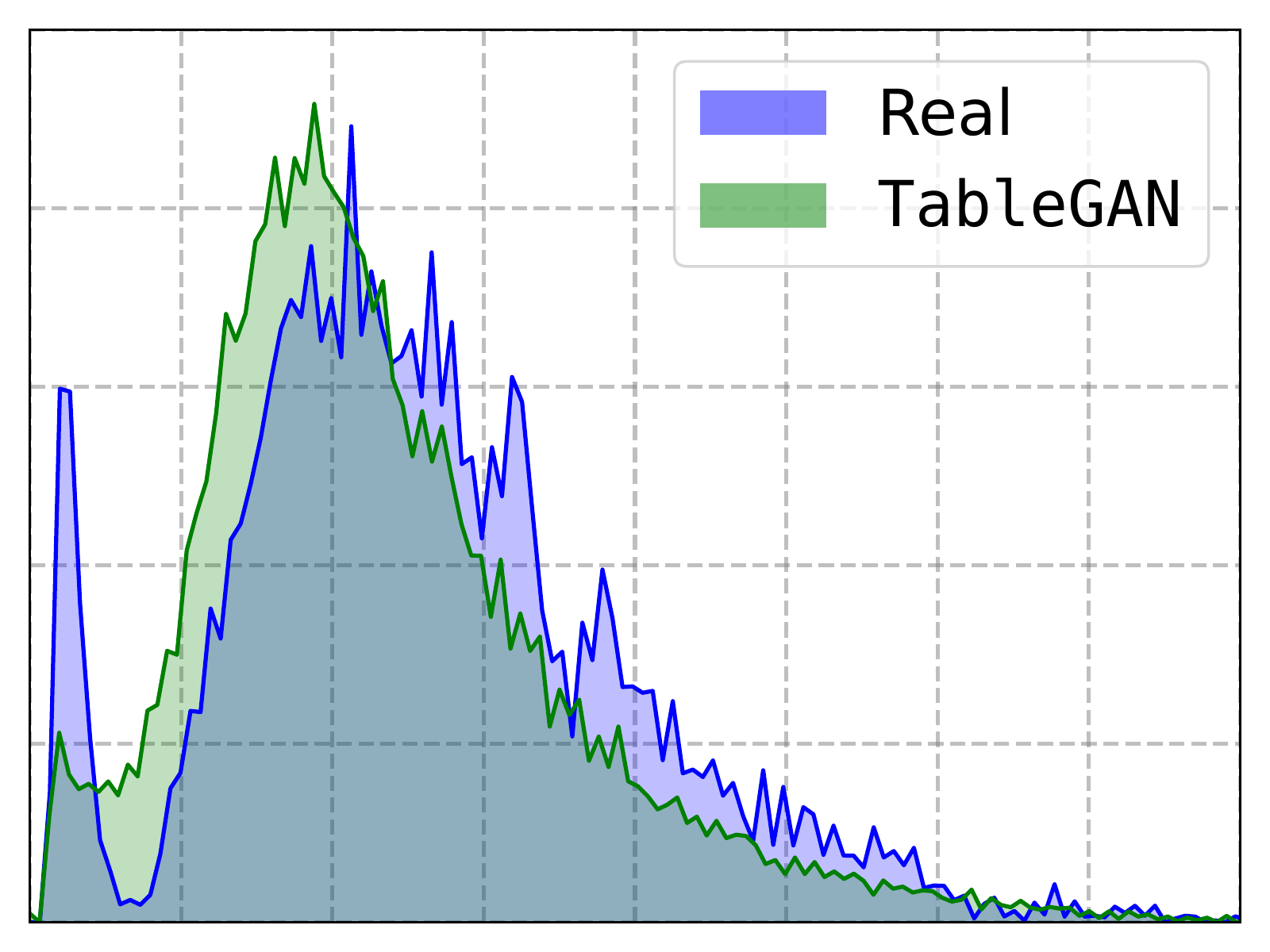}}
        \end{subfigure}
        \begin{subfigure}{\includegraphics[width=0.23\textwidth]{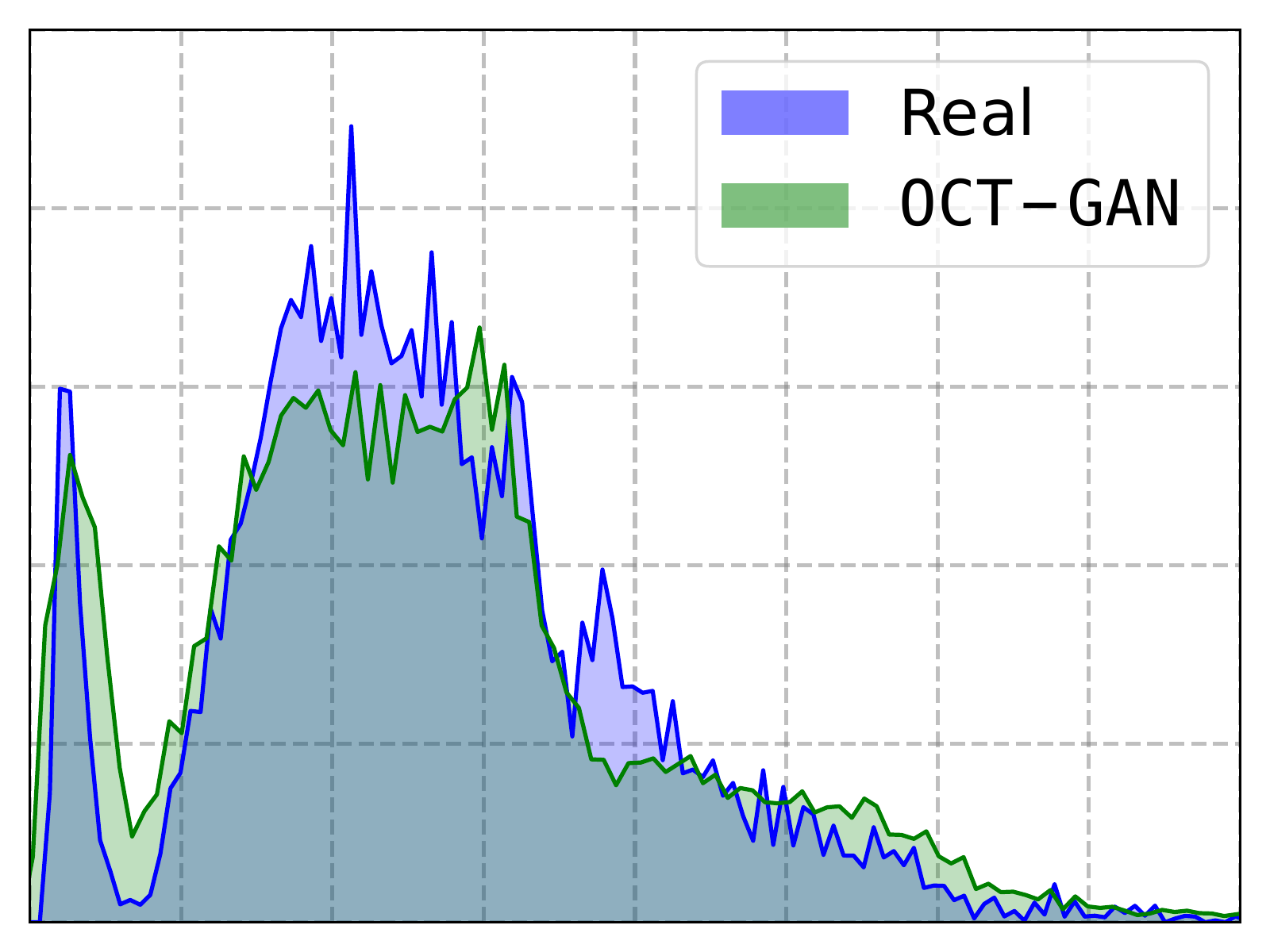}}
        \end{subfigure}
        \begin{subfigure}{\includegraphics[width=0.23\textwidth]{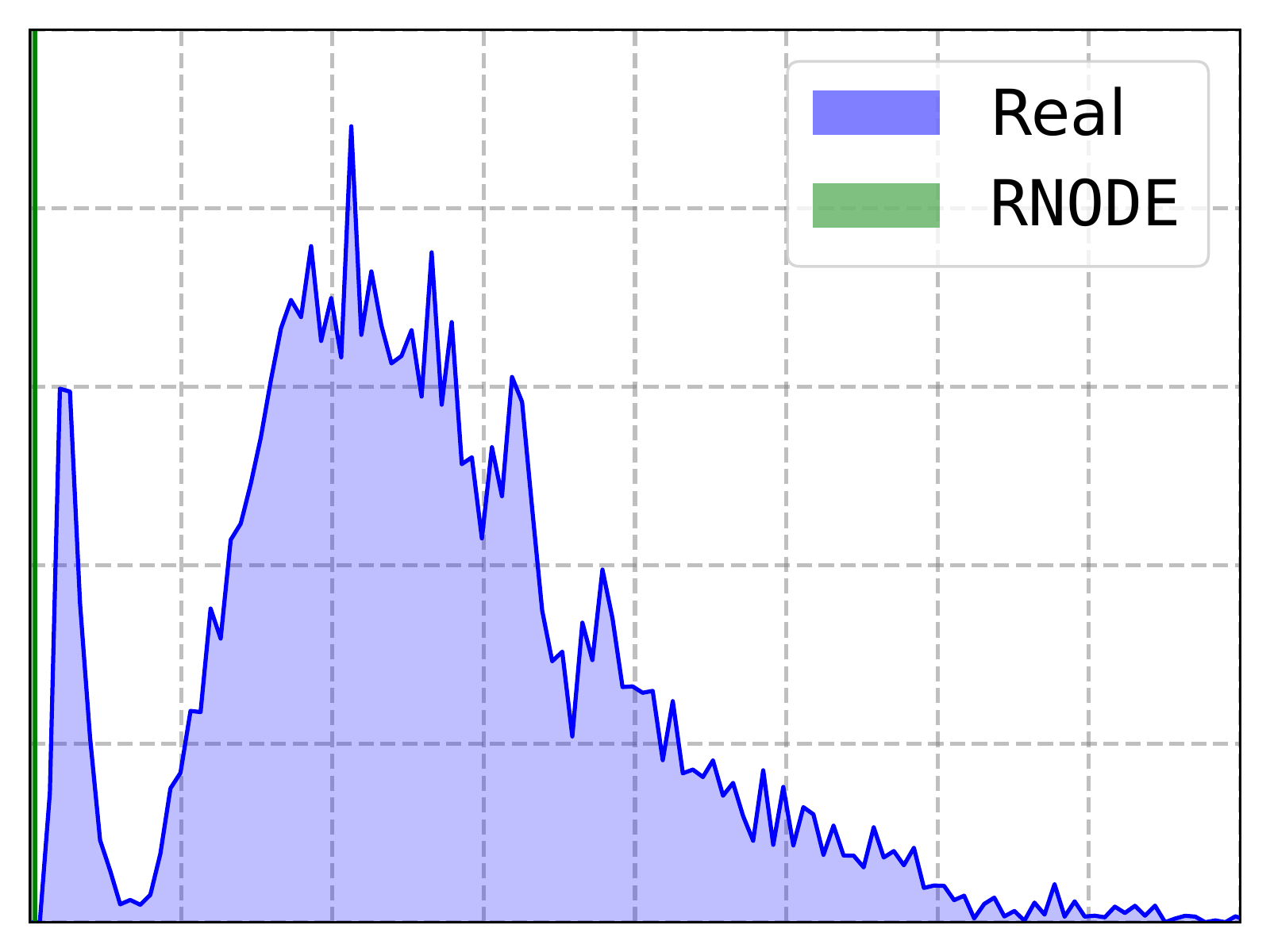}}
        \end{subfigure}
        \begin{subfigure}{\includegraphics[width=0.23\textwidth]{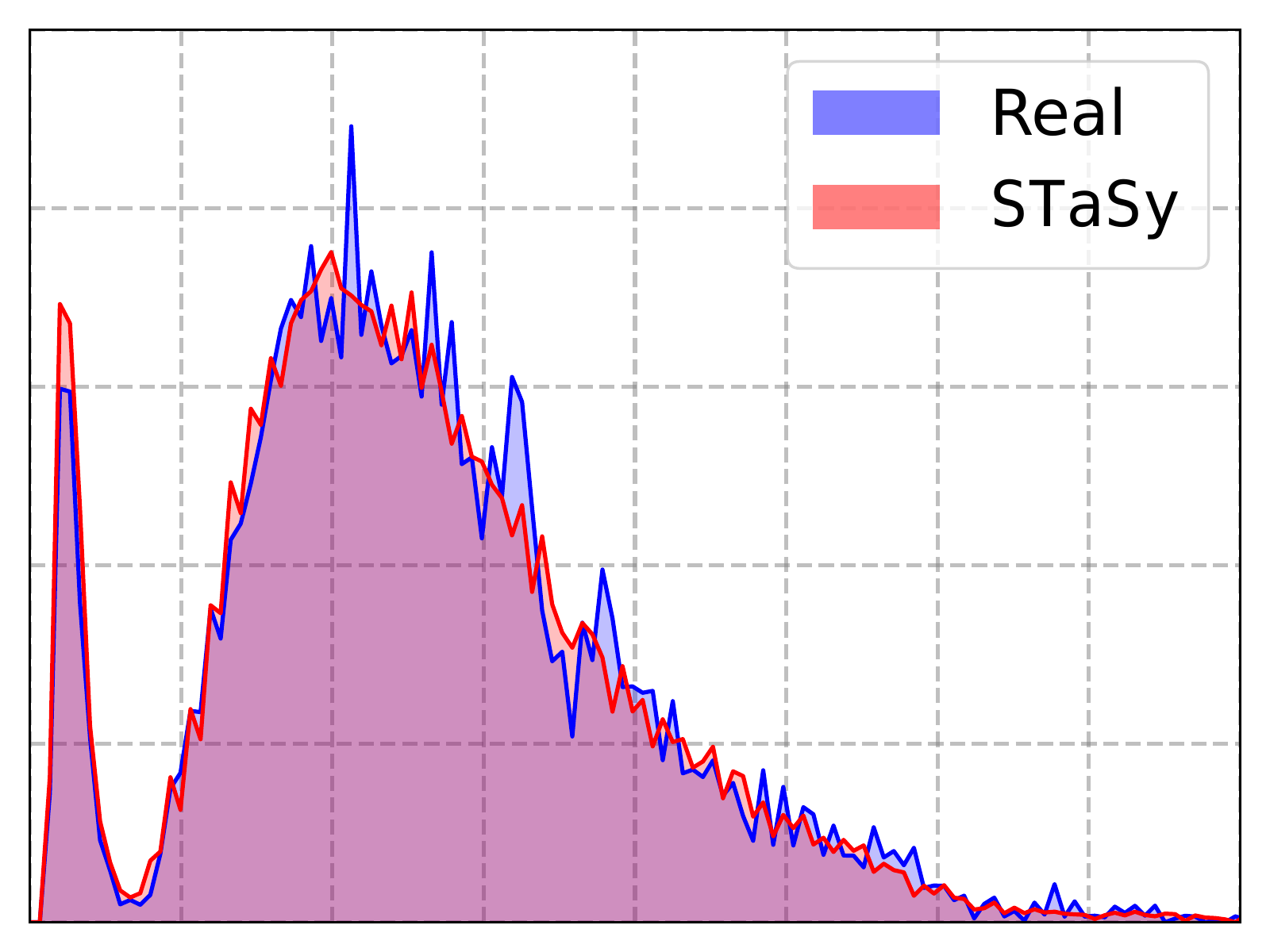}}
        \end{subfigure}
        \caption{Histograms of values in the \textit{min of best keyword} column of \texttt{News}}
        \label{fig:histo_news}
% \end{subfigure}
\end{figure}

\begin{figure}[h]
% \vspace{-1em}
% \begin{subfigure}
        \centering
         \begin{subfigure}{\includegraphics[width=0.23\textwidth]{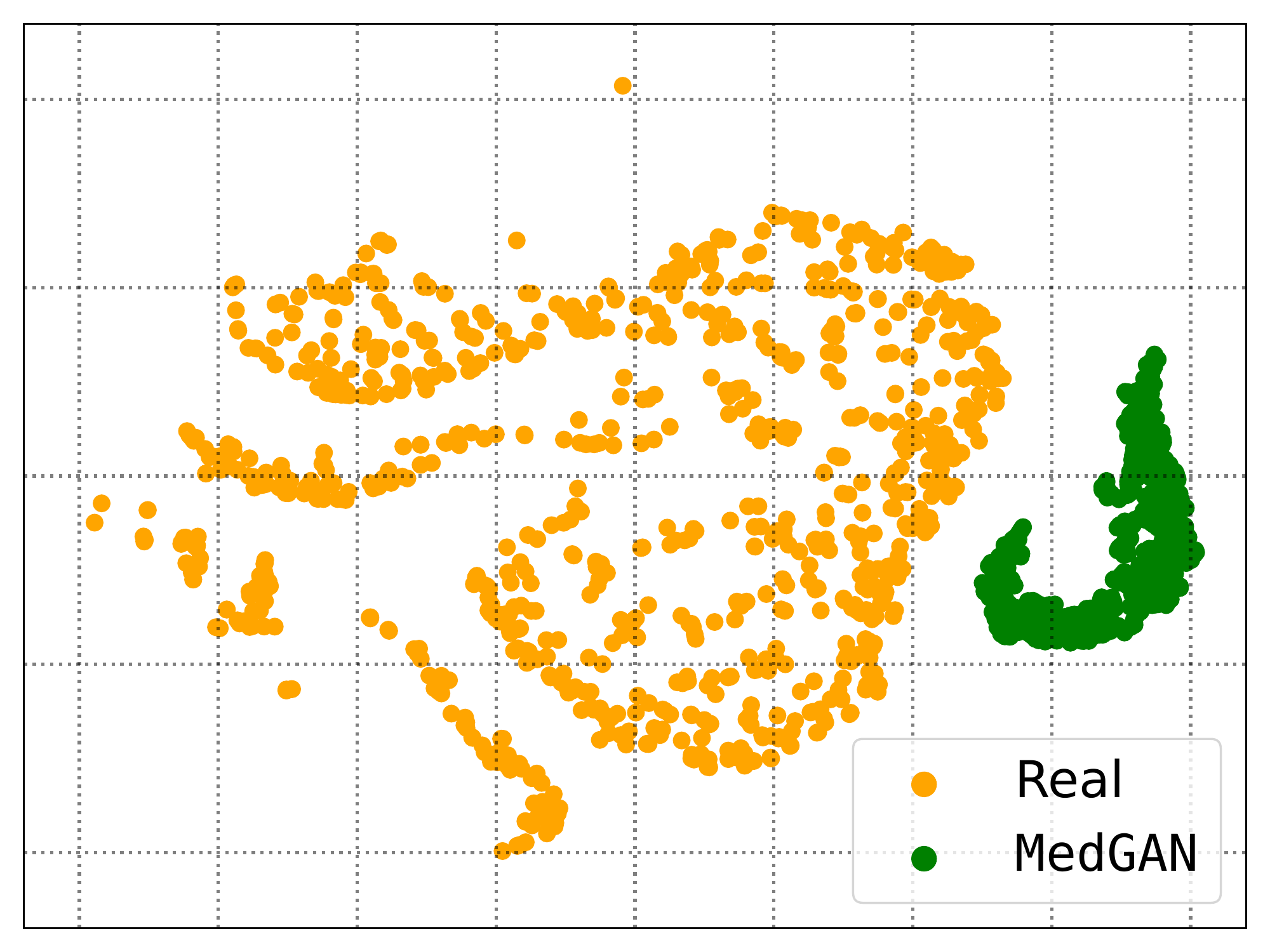}}
        \end{subfigure}
         \begin{subfigure}{\includegraphics[width=0.23\textwidth]{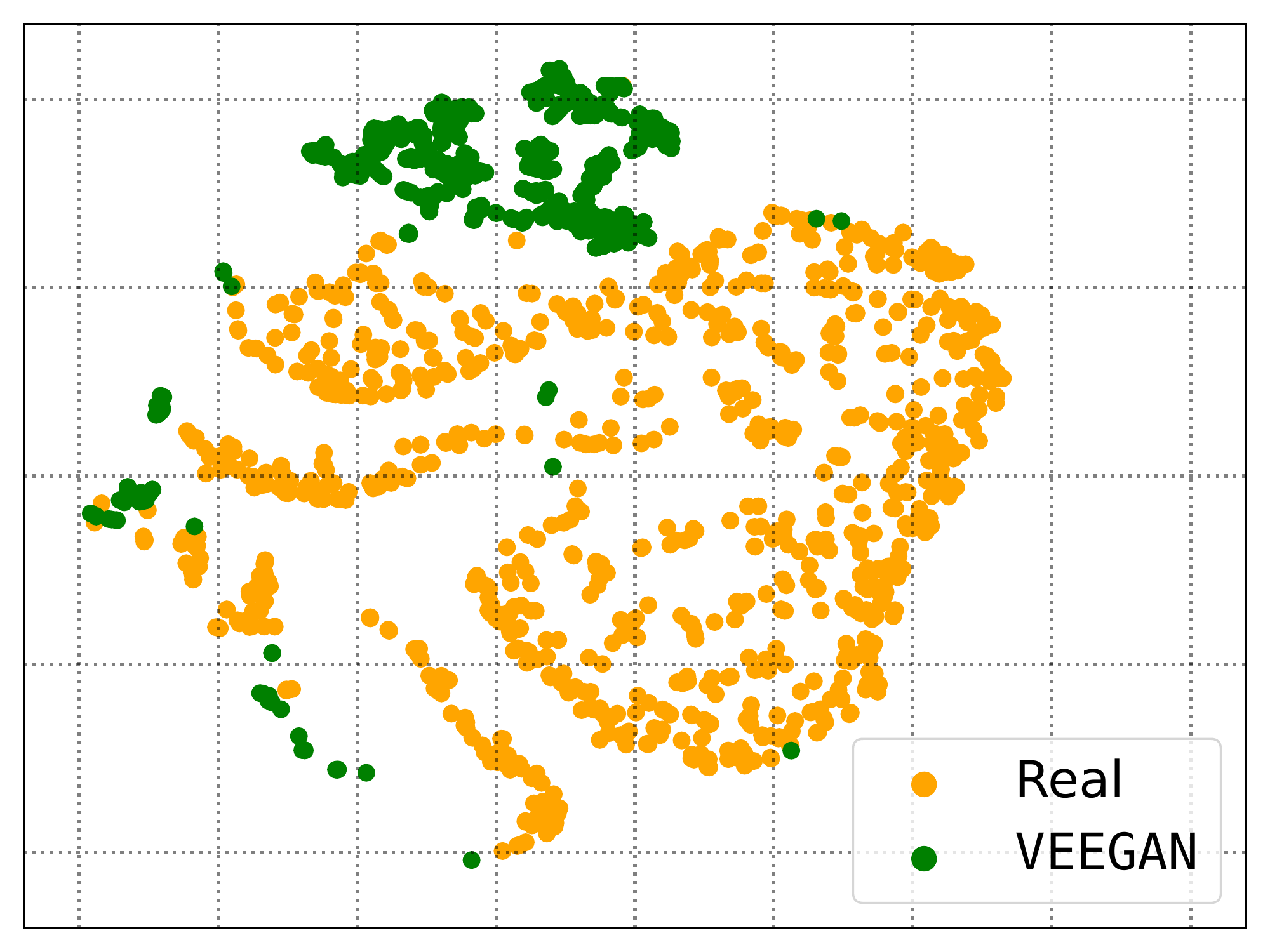}}
        \end{subfigure}
        \begin{subfigure}{\includegraphics[width=0.23\textwidth]{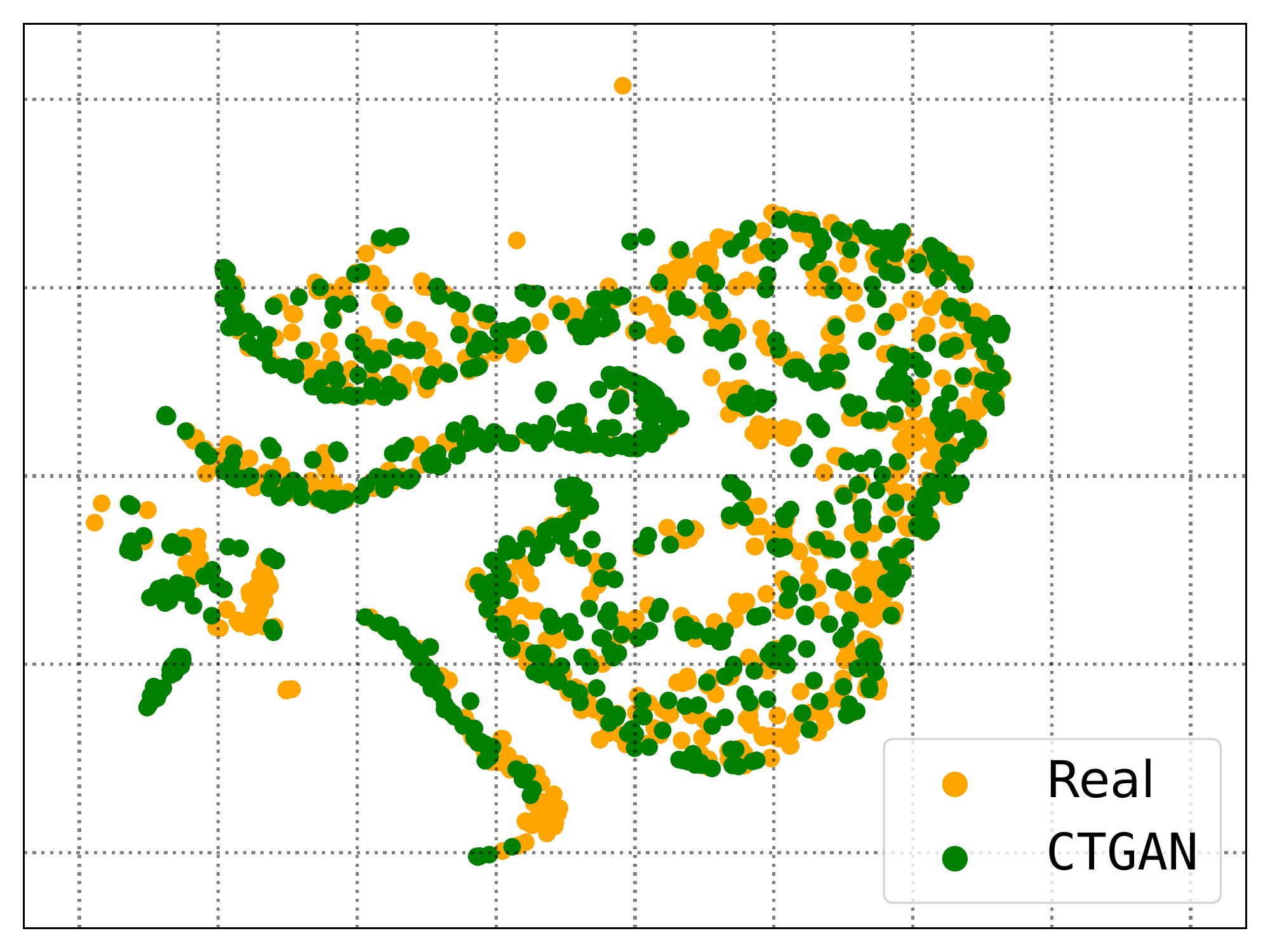}}
        \end{subfigure}
        \begin{subfigure}{\includegraphics[width=0.23\textwidth]{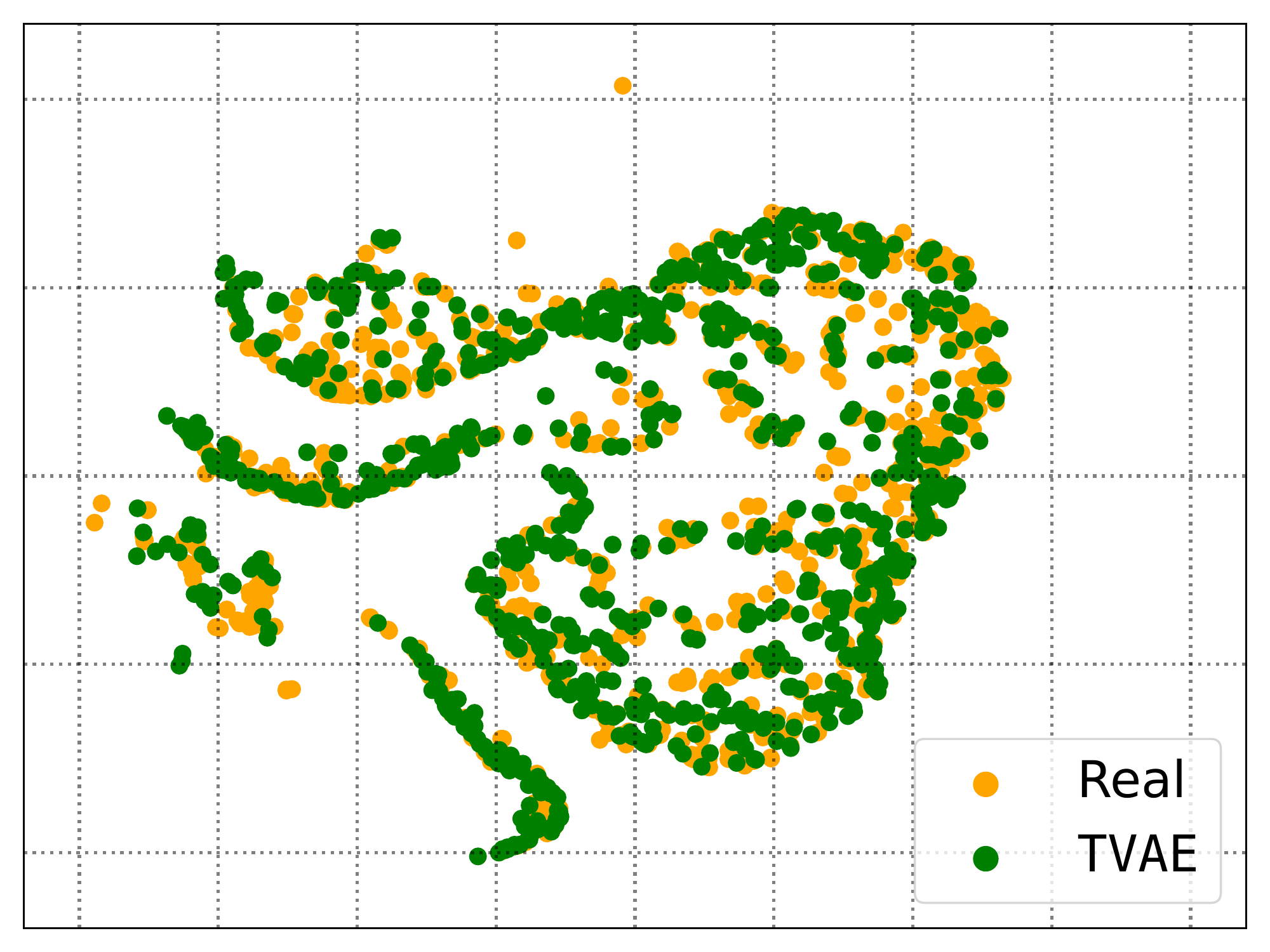}}
        \end{subfigure}
        \begin{subfigure}{\includegraphics[width=0.23\textwidth]{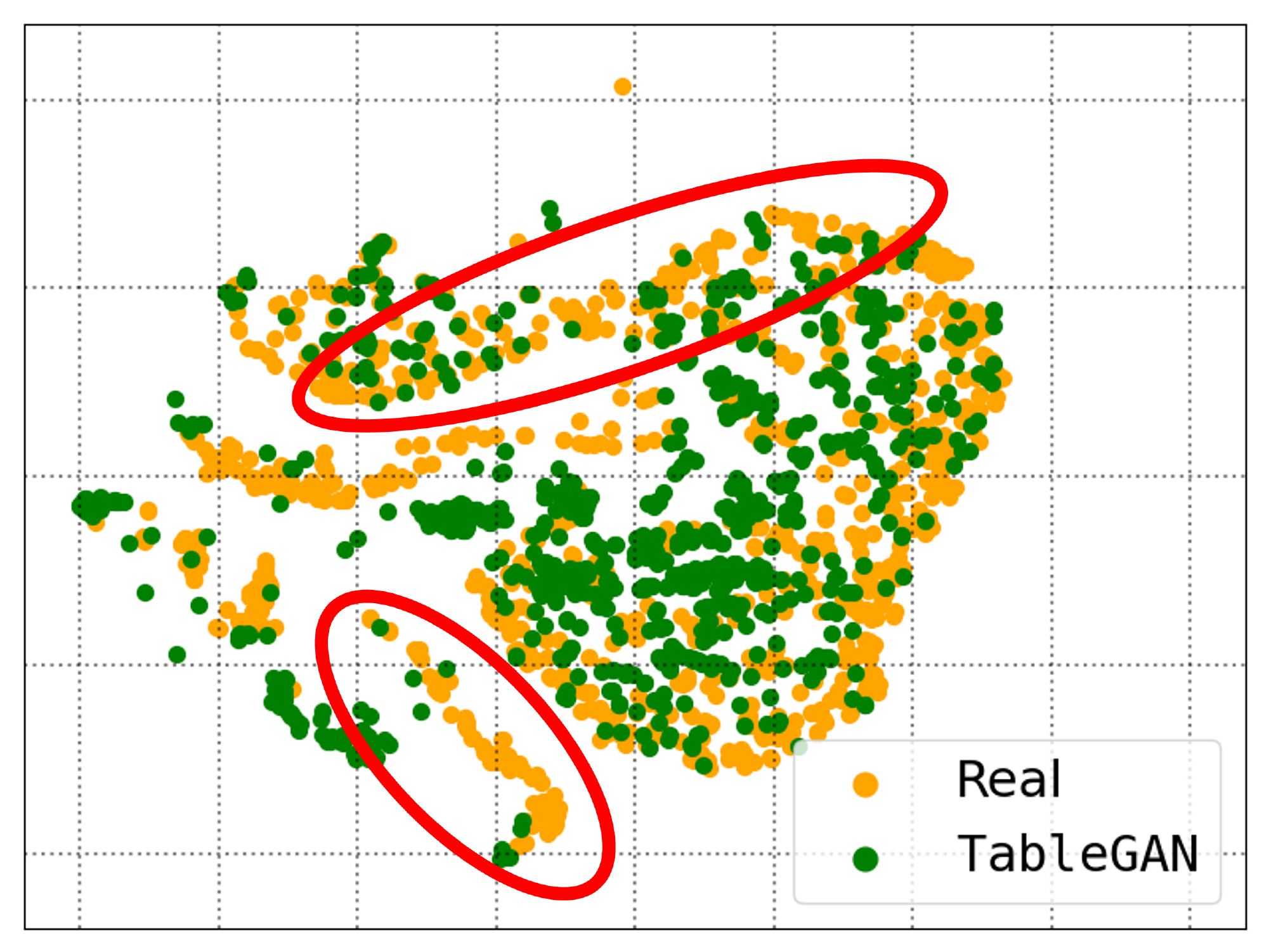}}
        \end{subfigure}
        \begin{subfigure}{\includegraphics[width=0.23\textwidth]{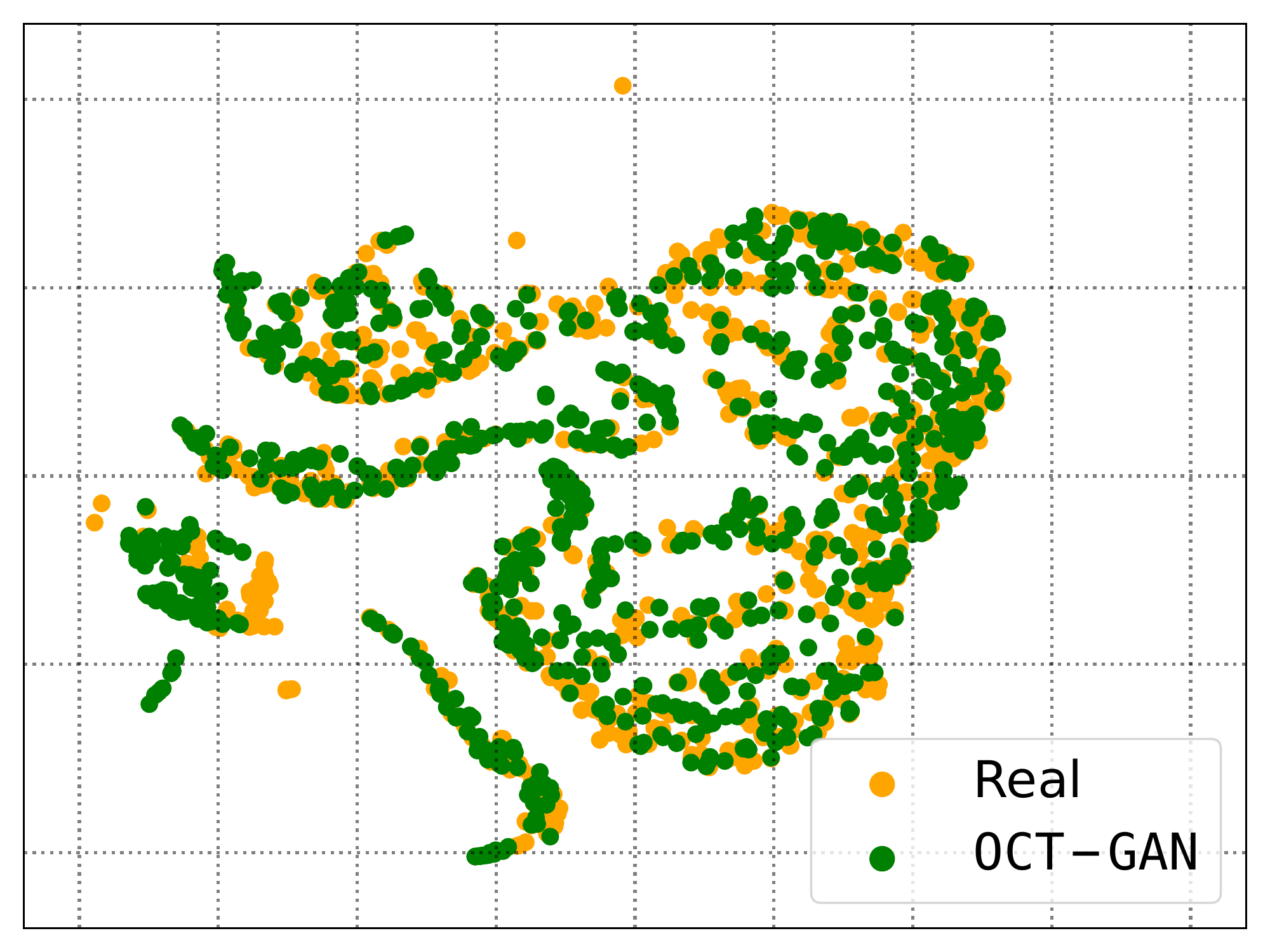}}
        \end{subfigure}
        \begin{subfigure}{\includegraphics[width=0.23\textwidth]{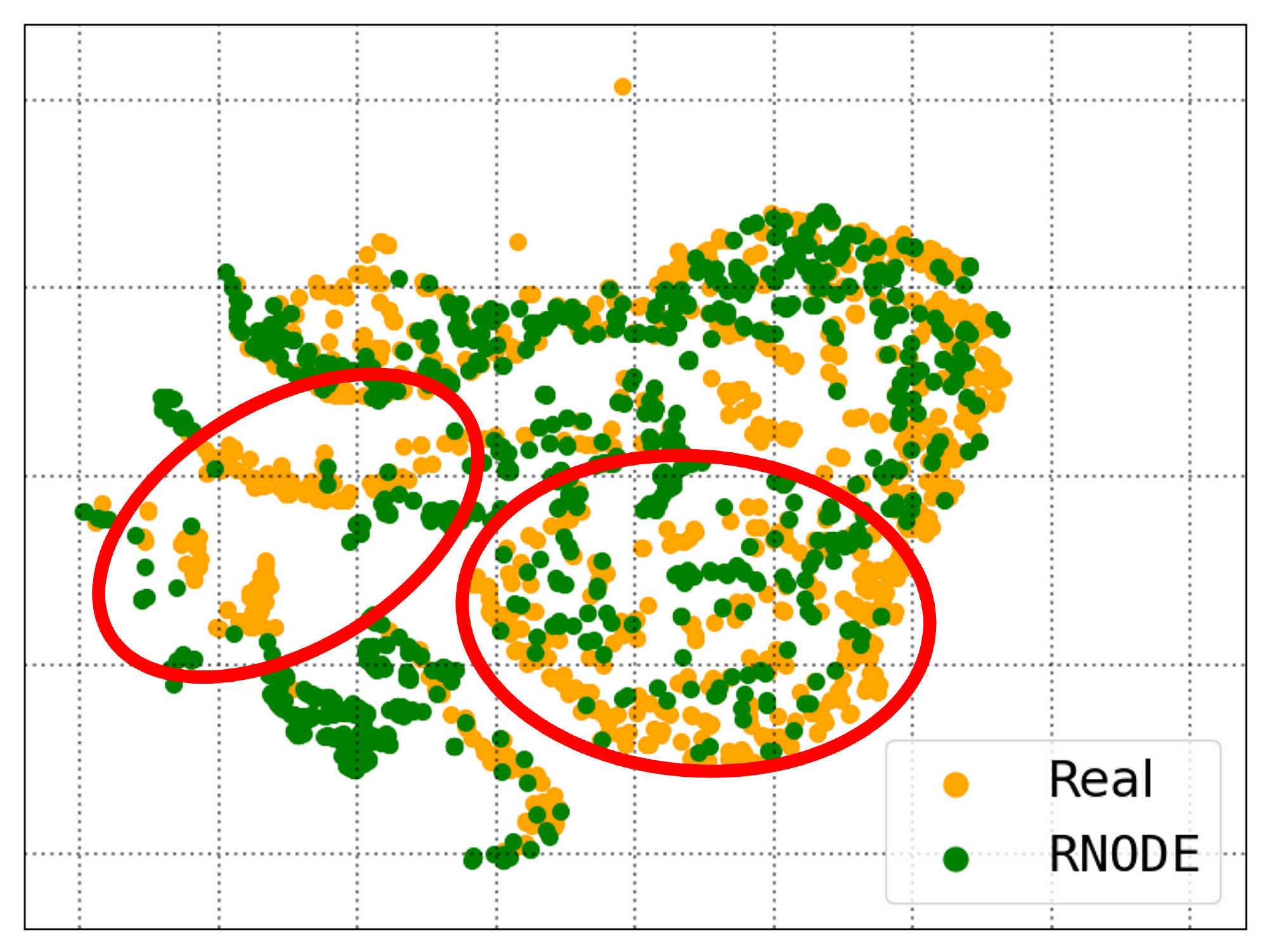}}
        \end{subfigure}
        \begin{subfigure}{\includegraphics[width=0.23\textwidth]{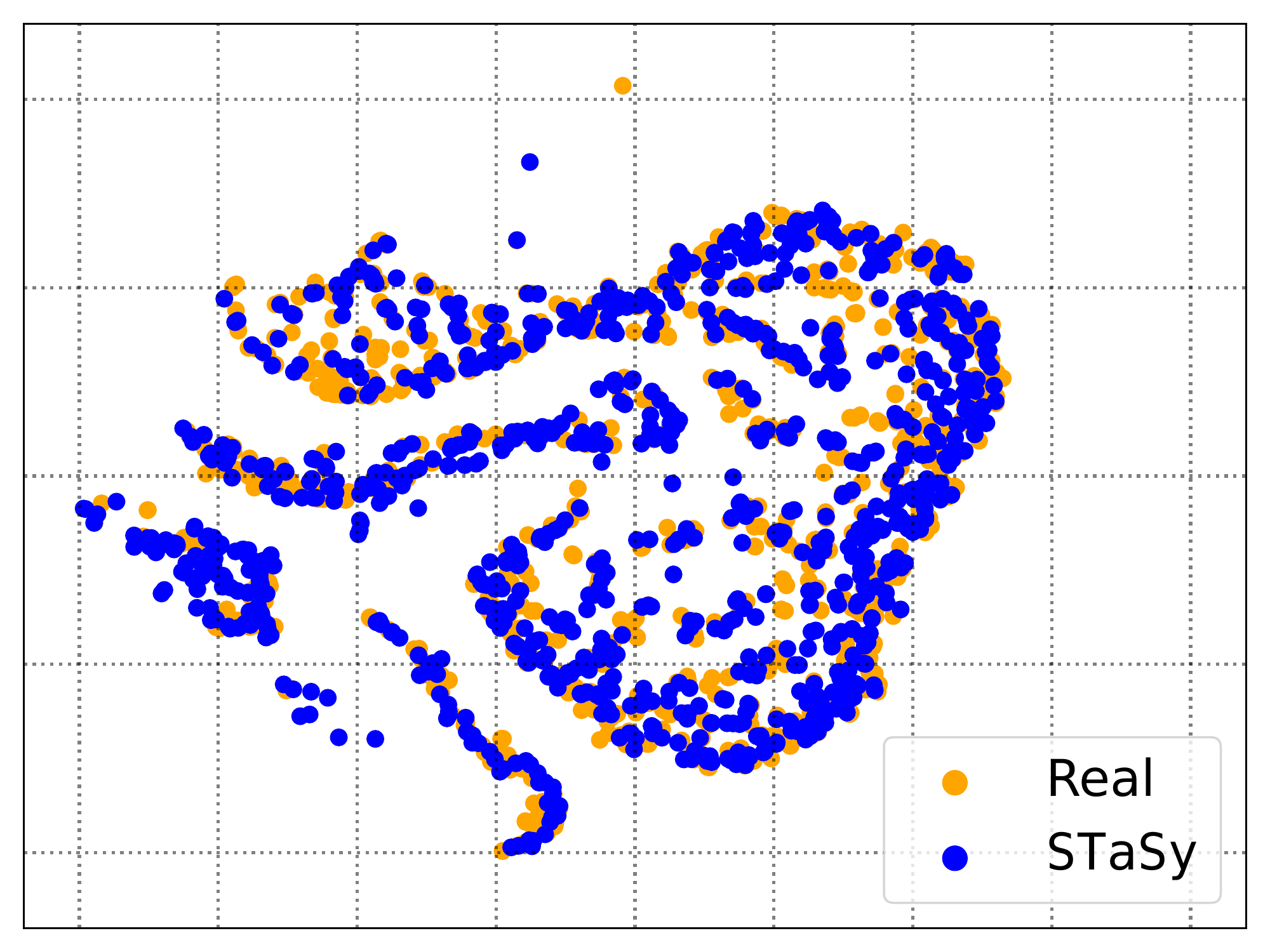}}
        \end{subfigure}
        \caption{t-SNE visualizations of fake and original records in \texttt{News}}
        \label{fig:tsne_news}
% \end{subfigure}
% \vspace{-1em}
\end{figure}

In \texttt{News}, the fake data by \texttt{STaSy} shows more reliable column-wise histogram than others in Figure~\ref{fig:histo_news}. As shown in Figure~\ref{fig:tsne_news}, all methods except for \texttt{TableGAN} and \texttt{RNODE} well generate fake records.

% \subsection{A comparison between \texttt{Na\"ive-STaSy} and \texttt{STaSy} w/o fine-tuning}

\end{document}